\documentclass{article}

\PassOptionsToPackage{numbers, compress}{natbib}
\usepackage[final]{neurips_data_2021}

\usepackage[utf8]{inputenc} 
\usepackage[T1]{fontenc}    
\usepackage{hyperref}       
\usepackage{url}            
\usepackage{booktabs}       
\usepackage{amsfonts}       
\usepackage{nicefrac}       
\usepackage{microtype}      
\usepackage{xcolor}         

\usepackage{graphicx}
\usepackage[linesnumbered,ruled,noend]{algorithm2e}
\usepackage{algpseudocode}%
\usepackage{balance} 
\usepackage{todonotes}
\usepackage{color}
\usepackage{colortbl}
\usepackage{comment}
\usepackage{xurl}
\usepackage{soul}

\title{An Information Retrieval Approach to Building Datasets for Hate Speech Detection}

\author{%
Md Mustafizur Rahman \\
School of Information \\
The University of Texas at Austin \\ 
\texttt{nahid@utexas.edu}
\And
Dinesh Balakrishnan \\
Department of Computer Science \\ 
The University of Texas at Austin \\ 
\texttt{dinesh.k.balakrishnan@utexas.edu}
\And
Dhiraj Murthy \\
School of Journalism and Media \\ 
The University of Texas at Austin \\ 
\texttt{Dhiraj.Murthy@austin.utexas.edu}
\And 
Mucahid Kutlu \\
Department of Computer Engineering  \\ 
TOBB University of Economics and Technology \\ 
\texttt{m.kutlu@etu.edu.tr} 
\And
Matthew Lease \\
School of Information \\
The University of Texas at Austin \\ 
\texttt{ml@utexas.edu}

}

\begin{document}

\maketitle

\begin{abstract}
Building a benchmark dataset for hate speech detection presents various challenges. Firstly, because hate speech is relatively rare, random sampling of tweets to annotate is very inefficient in finding hate speech. To address this, prior datasets often include only tweets matching known ``hate words''. However, restricting data to a pre-defined vocabulary may exclude portions of the real-world phenomenon we seek to model. A second challenge is that definitions of hate speech tend to be highly varying and subjective. Annotators having diverse prior notions of hate speech may not only disagree with one another but also struggle to conform to specified labeling guidelines. Our key insight is that the rarity and subjectivity of hate speech are akin to that of relevance in information retrieval (IR). This connection suggests that well-established methodologies for creating IR test collections can be usefully applied to create better benchmark datasets for hate speech. To intelligently and efficiently select which tweets to annotate, we apply standard IR techniques of {\em pooling} and {\em active learning}. To improve both consistency and value of annotations, we apply {\em task decomposition} and {\em annotator rationale} techniques. We share a new benchmark dataset for hate speech detection on Twitter that provides broader coverage of hate than prior datasets. We also show a dramatic drop in accuracy of existing detection models when tested on these broader forms of hate. Annotator rationales we collect not only justify labeling decisions but also enable future work opportunities for dual-supervision and/or explanation generation in modeling. Further details of our approach can be found in the supplementary materials \cite{rahman21-neurips21-supplementary}.
\end{abstract}

{\bf Content Warning}: We discuss hate speech and provide examples that might be disturbing to read. 

\section{Introduction}
\label{section:introduction}

Online hate speech constitutes a vast and growing problem in social media \cite{halevy2020preserving, macavaney2019hate, jurgens2019just, fortuna2018survey, schmidt2017survey, davidson2017automated}. For example, \citet{halevy2020preserving} note that the wide variety of content violations and problem scale on Facebook defies manual detection, including the rate of spread and harm such content may cause in the world. Automated detection methods can be used to block content, select and prioritize content for human review, and/or restrict circulation until human review occurs. This need for automated detection has naturally given rise to the creation of labeled datasets for hate speech \cite{poletto2021resources, hatespeechdata}.

Datasets play a pivotal role in machine learning, translating real-world phenomena into a surrogate research environments within which we formulate computational tasks and perform modeling. Training data defines the totality of model supervision, while testing data defines the yardstick by which we measure empirical success and field progress. Benchmark datasets thus serve to catalyze research and define the world within which our models operate. However, research to improve models is often prioritized over research to improve the data environments in which models operate, even though dataset flaws and limitations can lead to significant practical problems or harm \cite{vidgen2020directions, laaksonen2020datafication, madukwe2020data,  sambasivan2021everyone, northcutt2021pervasive}. \citet{grondahl2018all} argue that for hate speech, the nature and composition of the datasets are more important than the models used due to extreme variation in annotating hate speech, including definition, categories, annotation guidelines, types of annotators, and aggregation of annotations.

\begin{figure}[h]
\centerline{\includegraphics[width=0.49\textwidth]{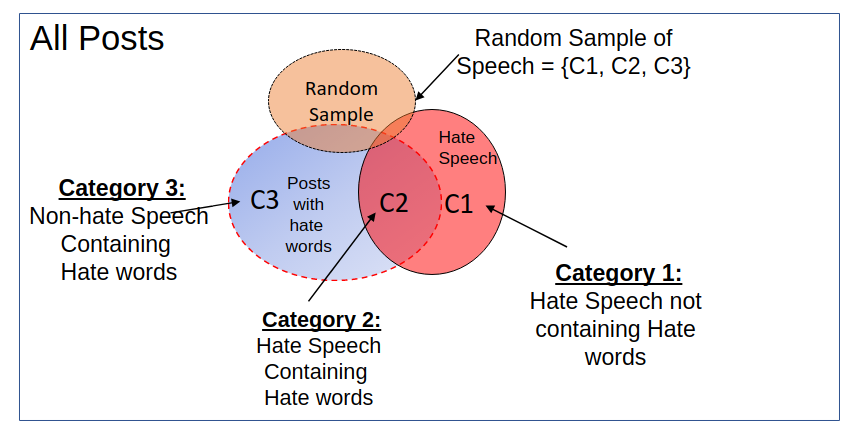}} 
\caption{Hate speech coverage resulting from different choices of which social media posts to annotate. Given some list of ``hate words'' by which to filter posts, some matching posts are indeed hateful (region C2, {\em true positives}) while other matching posts are benign (region C3, {\em false positives}). Region C1 indicates {\em false negatives}: hate speech missed by the filter and mistakenly excluded. Random sampling correctly overlaps C1+C2 but is highly inefficient in coverage.}
\label{Figure:landscape_of_hate_speech}
\vspace{-1em}
\end{figure}
\begin{table*}[h]
\begin{center}
\scriptsize
\caption{Example tweets from across the Venn diagram regions shown in Figure \ref{Figure:landscape_of_hate_speech}.} 
\begin{tabular}{c  p{0.82\linewidth}}
 \hline
 {\bf Figure \ref{Figure:landscape_of_hate_speech} Region} & {\bf Documents} \\
 \hline
 
 C1 & ``When you hold all 3 branches and still can't get anything done...you must be a republican." \\
 \hline
 C1 & ``You can yell at these Libs all day. They don't listen. But if you ridicule them nationally they can't take it" \\
 \hline

 C2 & ``Real Rednecks don't use the internet you {\color{red}pussy}'' \\
 \hline
 C2 & ``guarantee slavery wasnt even tht bad...you know {\color{red}niggas}  over exxagerate dramatize everything''\\
 \hline
 C3 & ``Don't call me {\color{red}black} no more. That word is just a color, it ain't facts no more" \\
 \hline
C3 & ``AGENT Are you crazy? You'll never make it in the white {\color{red}trash} rap/rock genre with a name like that SKIDMORE ROCKEFELLER" \\
 \hline
\label{table:example_of_hate_categories}
\end{tabular}
\end{center}
\vspace{-1em}
\end{table*}

Fortunately, many valuable datasets already exist for detecting hate speech \cite{hatespeechdata, poletto2021resources, waseem-hovy-2016-hateful, waseem2016you, davidson2017automated, founta2018large, de-gibert-etal-2018-hate, golbeck2017large}. However, each dataset can be seen to embody an underlying design tradeoff (often implicit) in how to balance cost vs.\ coverage of the phenomenon of hate speech, in all of the many forms of expression in which it manifests. At one extreme, random sampling ensures representative coverage but is highly inefficient (e.g., less than  3\% of Twitter posts are hateful \citep{founta2018large}). At the other extreme, one can annotate only those tweets matching a pre-defined vocabulary of ``hate words'' \cite{hatebase,noswearing} whose presence is strongly correlated with hateful utterances \citep{waseem-hovy-2016-hateful, waseem2016automatic, davidson2017automated, golbeck2017large, grimminger2021hate}. By restricting a dataset to only those tweets matching a pre-defined vocabulary, a higher percentage of hateful content can be found. However, this sacrifices representative coverage for cost-savings, yielding a biased dataset whose distribution diverges from the real world we seek to model and to apply these models to in practice \cite{kennedy2018gab}. If we only look for expressions of hate matching known word lists, the resulting dataset will completely miss any expressions of hate beyond this prescribed vocabulary. This is akin to traditional expert systems that relied entirely on hand-crafted, deterministic rules for classification and failed to generalize beyond their narrow rule sets. Such a resulting benchmark would provide only partial representation for the real-world phenomenon of interest.

{\bf Figure \ref{Figure:landscape_of_hate_speech}}
presents a Venn diagram illustrating this. {\bf Table \ref{table:example_of_hate_categories}} examples further highlight the weakness of only annotating posts matching known ``hate words''  \citep{davidson2017automated, waseem-hovy-2016-hateful, grimminger2021hate, golbeck2017large}, covering only regions C2-C3. The prevalence of hate found in such datasets is also limited by the word lists used \citep{sood2012profanity}. 

We can view the traditional practice above as following a two-stage pipeline. Firstly, a simple Boolean search is performed to retrieve all tweets matching a manually-curated vocabulary of known hate words. Secondly, because the above retrieval set is often quite large, random down-sampling is applied to reduce the set to a smaller, more affordable scale for annotation. In information retrieval (IR), a similar pipeline of human annotation following Boolean search has been traditionally used in legal {\em e-discovery} and medical {\em systematic review} tasks \cite{Lease16-medir}, though the Boolean filter in those domains casts a very wide net to maximize recall, whereas hate word lists tend to emphasize precision to ensure a high percentage (and efficient annotation) of hateful content. However, just as probabilistic models have largely superseded traditional deterministic, rule-based expert systems, we expect that a probabilistic retrieval model can more intelligently and efficiently select which content to annotate. 

A second key challenge in constructing hate speech datasets is that what constitutes hate speech is quite subjective, with many competing definitions across legal, regional, platform, and personal contexts \citep{davidson2017automated, waseem-hovy-2016-hateful, fortuna2018survey}. Consequently, annotators having diverse prior notions of hate speech often disagree with one another (especially when labeling guidelines allow more subjective freedom), and may also struggle to conform to stricter labeling guidelines. The use of inexpert crowd annotators can further exacerbate concerns regarding label reliability \cite{sen2015turkers,northcutt2021pervasive}. 

A key insight of our study is that the rarity and subjectivity of hate speech are akin to that of relevance in information retrieval (IR) \citep{sanderson2010test}. This suggests that established methods for creating IR {\em test collections} might also be applied to create better hate speech benchmark datasets. To intelligently and efficiently select which content to annotate for hate speech, we apply two known IR techniques for building test collections: {\em pooling} \citep{sanderson2010test} and {\em active learning} (AL) \citep{cormack_autonomy_2015, rahman_ICTIR_2020}. Both approaches begin with a very large random sample of social media posts to search (i.e., the {\em document collection}). With pooling, we use existing hate speech datasets and models to train a diverse ensemble of predictive models. We then prioritize posts for annotation by their predicted probability of being hateful, restricting annotation to the resulting {\em pool} of highly-ranked posts. For nearly 30 years, NIST TREC has applied such pooling techniques with a diverse set of ranking models in order to optimize the coverage vs.\ cost tradeoff in building test collections for IR, yielding benchmark datasets for fair and robust evaluation of IR systems. AL, on the other hand, requires only an initial set of {\em seed} posts from which a classifier is progressively trained to intelligently select which posts should be labeled next by human annotators. The tight annotation-training feedback loop provides greater efficiency in annotation, and unlike pooling, it does not require (nor is biased by) existing hate speech datasets.\footnote{Strictly speaking, initial seed documents used with AL may also include bias that influences training.} 
We re-iterate that pooling and AL are established methods in IR; our translational contribution is showing that these techniques can also be usefully applied to efficiently build hate speech datasets without vocabulary filters. 

To address subjectivity in annotation, we apply two other techniques from the IR and crowdsourcing literatures. Firstly, applying {\em task decomposition} \cite{Zhang-sigir14,noronha2011platemate}, we decompose the subjective and complex judging task into a structured set of smaller tasks. Secondly, we design the annotation task to let us measure annotator {\em self-consistency} \cite{Zhang-sigir14}. This provides a means of establishing label validity on the basis of an annotator being internally consistent, which is important for subjective labeling tasks in which we expect {\em inter-annotator} disagreement across annotators. Finally, we require that annotators provide constrained {\em rationales} justifying their labeling decisions \cite{mcdonnell16-hcomp,Kutlu18-sigir}, which prior work has found to improve label accuracy, verifiability, and utility. 

Using pooling, we create a new benchmark dataset\footnote{\url{https://github.com/mdmustafizurrahman/An-Information-Retrieval-Approach-to-Building-Datasets-for-Hate-Speech-Detection}} for hate speech detection on Twitter, consisting of 9,667 tweets in total: 4,668 labeled as part of iterative pilot experiments refining quality, and our final high quality set of 4,999 used in our experiments. Each tweet is annotated by three workers on Amazon Mechanical Turk, at a  total dataset annotation cost of approximately \$5K USD. 
We assess this dataset with respect to the {\em prevalence} (i.e., annotation efficiency) and {\em relative coverage} (breadth) of hate found, as well as evaluation of detection models on these broader forms of hate.  

Firstly, we show in Section \ref{sec:props} that pooling yields 14.6\% {\em relative coverage}, far better than the best prior work \cite{founta2018large}'s combination of random sampling and keyword filtering (10.4\%). Secondly, regarding efficiency in selecting what content to annotate, 14.1\% of our annotated content is found to be hateful, a {\em prevalence} that exceeds a number of prior datasets  \citep{davidson2017automated, fortuna2018survey, grimminger2021hate} while simultaneously also providing the aforementioned greater coverage. Note that prevalence can be inflated by only annotating a highly-restricted vocabulary or set of user posts; e.g., while \citet{waseem-hovy-2016-hateful} achieve 31\% prevalence, their dataset is highly skewed, with a single user generating 96\% of all racist tweets, while another user produces 44\% of all sexist tweets \cite{arango2019hate}. Just as precision and recall are balanced in classification, we wish to balance prevalence (efficiency) and coverage (fidelity) in creating a benchmark dataset that is faithful to the phenomenon while being affordable to create. Finally, we benchmark several recent hate speech detection models  \citep{devlin2018bert, agrawal2018deep, badjatiya2017deep} and find that the performance of these models drops drastically when tested on these broader forms of hate in our dataset (Table \ref{Table:Benchamark_results}). 

To further improve annotation efficiency, and to further reduce dependence on prior datasets (and potential bias from them), we also report retrospective experiments that assess AL on our dataset created via pooling. We compare several known AL strategies and feature representations, showing that AL can effectively find around 80\% of the hateful content at 50\% of the cost of pooling.  


As a word of caution, while machine learning models are usually intended to serve the social good (e.g., to reduce online hate speech), it is now well known that machine learning models can also aggravate inequality, perpetuate discrimination, and inflict harm \cite{noble2018algorithms, eubanksautomating}. As noted earlier, one source of such harms are dataset flaws and limitations that are unknown, ignored, or unmitigated 
\cite{vidgen2020directions, laaksonen2020datafication, madukwe2020data,  sambasivan2021everyone, northcutt2021pervasive, boyd2021datasheets}. For this reason, we emphasize that hate speech detection researchers and practitioners should attend carefully to these issues when deploying any automated hate speech detection system, including models trained on our dataset. We also want to draw attention to an important issue largely unremarked in prior work on hate speech dataset research: the potential health risks to human annotators from sustained exposure to such disturbing content \cite{steiger2021psychological}.

Please see our supplementary material \cite{rahman21-neurips21-supplementary} for additional discussion of related work, our approach, resulting dataset properties, risks to annotators, and other limitations and risks of our study.

\section{Methods}

To intelligently select which tweets to annotate in order to optimize the prevalence vs.\ coverage tradeoff in our annotated hate speech dataset, we apply two approaches from constructing test collections in information retrieval (IR): pooling and active learning. Unlike most prior work, we do not use any keywords to filter the set of tweets. Instead, we first collect a large, random (and unbiased) sample of tweets via Twitter's API in which to search for hateful content. 
Following IR terminology, we refer to each tweet as a {\em document} and the corpus as a {\em document collection}.

\subsection{Selecting Tweets via Pooling}
\label{section:pooling_algorithm}
Pooling \citep{sanderson2010test} applies a set of trained machine learning models to predict the hatefulness of documents; the documents that are predicted most likely to be hateful are then selected for human annotation. However, we found that annotating only the top-ranked documents found less diversity in forms of hate than found more broadly online (another example of the {\em prevalence} vs.\ {\em coverage} tradeoff). To further promote diversity, we instead perform a simplified form of stratified sampling. Specifically, we define a minimum threshold $t$, select all tweets with whose predicted hatefulness exceeds $t$ for any model in our ensemble, and then randomly downsample this set to $b$ tweets, given annotation budget $b$.  Intuitively, lower values of $t$ inject greater randomness into selection, promiting diversity. 

Our pooling algorithm is detailed in {\bf Algorithm \ref{algorithm:Pooling_hate_speech}}. Given document collection $X$, we select a subset of $b$ documents to annotate for our hate speech dataset $R$.  Our algorithm requires three inputs: i) a set of prior hate speech datasets $D$; ii) a set of classifiers $C$; and iii) the aforementioned threshold $t$. Given a prior hate speech dataset $D_i$ and a classifier  $C_j$, we induce a machine learning model $\hat{C_j^i}$, by training $C_j$ on $D_i$ (Line 4). Then we employ this $\hat{C_j^i}$ to predict a hatefulness score of every document $x$ in the  collection $X$ (Line 5). If the predicted score of document $x$ is greater than the provided threshold $t$, we add it to the set $S$ (Lines 6 - 8). This process iterates for all  datasets and  classifiers in the set $D$ and $C$ (Lines 2 - 8), respectively. Finally, we randomly select $b$ unique documents from $S$ (Line 9), as discussed earlier. 
to construct the hate speech dataset $R$ (Line 10).

\begin{algorithm}[t]
    \SetKwInOut{Input}{Input}
    \SetKwInOut{Output}{Output}
    \Input{~~Document collection $X$ $\bullet$ Set of prior hate speech datasets D \newline $\bullet$ Set of classifiers C $\bullet$ total budget $b$ $\bullet$ selection threshold $t$}
    \Output{~~Set of documents to be annotated hate speech dataset $R$}
        $S \leftarrow  \emptyset$ \\ 
        \For{$i \gets 1$ \KwTo $|D|$}{ 
            \For{$j \gets 1$ \KwTo $|C|$}{ 
                $\hat{C_j^i} \leftarrow$ \text{train\_model$(C_j$, $D_i$)} 
                \Comment train classifier $C_j$ using data $D_i$ \\
                $ \forall x\in X$ predict hatefulness of document $x$ using $\hat{C_j^i}$\\
                \For{$k \gets 1$ \KwTo $|X|$}{ 
                    \algorithmicif\ the predicted hatefulness score of $x^k \ge t$ \\ \algorithmicthen\  $S \leftarrow S \cup x^k$ \\ 
                }
            }
        }
        $R \leftarrow $ Randomly sample $b$ unique documents from $S$ \Comment Set of selected documents \\
\caption{Pooling Approach for Selecting Documents to Annotate}
\label{algorithm:Pooling_hate_speech}
\end{algorithm}



\subsection{Selecting Tweets via Active Learning (AL)}
\label{section:method_active_learning}
The underlying assumption of the pooling-based approach is that there exist prior hate speech datasets on which text classification models can be trained to kickstart the pooling technique. However, this assumption does not always hold \cite{oard2004building}, especially with less-studied languages (e.g., Amharic, Armenian, etc.). Instead, AL \citep{settles2012active,cormack_autonomy_2015,rahman_ICTIR_2020}  
can be applied using the following steps. 

\noindent\textbf{I. Training a Machine Learning Model}. To initiate the active learning process, we need a machine learning model. For AL, we adopt logistic regression as a simple model and two alternative feature representations: i) simple TF-IDF \citep{salton1988term}; and ii) BERT embedding from  Twitter-roBERTa-base \citep{TwitterroBERTa}. 

\noindent\textbf{II. Defining Document Selection Criteria}. 
Given a trained model, we can then utilize the predicted posterior probability $p(y^i|x^i)$ of document $x^i$ being hateful to decide whether or not to annotate that document.
While uncertainty sampling is most commonly used with AL to select instances to annotate that are close to the decision hyperplane, prior work in AL for IR \cite{cormack_autonomy_2015} has shown empirically the benefit of prioritizing selection of examples predicted to come from the rare class when the data is highly skewed. This is because it is important to expose the learner to as many examples of the rare class as possible. In addition,  because the model will often be wrong, misses (and near-misses) will still provide ample exposure to ambiguous examples across classes. Because hate speech is quite rare, \citep{founta2018large}, 
we thus select those documents for annotation which are most likely to be hateful. 
Thus we employ {\em Continuous Active Learning} (CAL) \citep{cormack_autonomy_2015,rahman_ICTIR_2020}, which utilizes $p(y^i|x^i)$ via {\bf Eqn. \ref{equation: CAL}}.

\vspace{-0.5em}
\begin{equation}
\label{equation: CAL}
x^\star = \arg\!\max\limits_{i}\  p(\text{hateful}|x^i)
\end{equation} 
\vspace{-0.5em}

We also implement traditional uncertainty sampling \citep{lewis1994sequential}, referred to in prior IR work as {\em Simple Active Learning} (SAL) \citep{cormack_autonomy_2015},  which selects a document for human annotation when the classifier is most uncertain about whether or not a document is hateful. Typically an entropy-based uncertainty function \citep{settles2012active} is employed for SAL: 

\begin{equation}
Uncertainty(x)  = - \sum_{y \in Y} P(y|x) \log P(y|x) 
\end{equation}
where $y$ is either hateful or non-hateful. With this binary classification, SAL selects: 
\begin{equation}
\label{equation: SAL}
x^\star = \arg\!\min\limits_{i}\ |p(\text{hateful}|x^i)-0.5|
\end{equation}

\noindent\textbf{III. Seed Documents}. Some {\em seed documents} are necessary to induce the initial machine learning model. Prior work \citep{rahman_ICTIR_2020} has shown that a few seed documents can suffice to kickstart the AL process. Seed documents can be collected from  annotators in various ways: they might write hateful documents (e.g., tweets, posts) or search social media platforms for hateful documents. 

\begin{algorithm}
    
    \SetAlgoLined
    \SetKwInOut{Input}{Input}
    \SetKwInOut{Output}{Output}

    \Input{~~Document collection $X$ $\bullet$ batch size $u$ $\bullet$ total budget $b$}
    \Output{~~Hate Speech Dataset $R$}

        Select seed documents $S$     	\\
        $R \leftarrow \{\langle x^i, y^i \rangle\ |\ x^i \in S\}$ \Comment Collect initial judgments \\
    	Learn the initial machine learning model $c$ using $R$\\
    	$b \leftarrow b - |S|$ \Comment Update remaining budget \\
    \While {True}{
    	\algorithmicif\ $b < u$ \algorithmicthen\ \Return \Comment Budget exhausted \\
    	$\forall x\in X$ predict hatefulness of document $x$ using $c$\\
        Select $u$ documents $S \in X$ to judge next 
        \\
       	
        $R\leftarrow R\cup \{\langle x^i, y^i \rangle\ |\ x^i \in S\}$ \Comment Collect judgments\\
       	Re-train the machine learning model $c$ using expanded $R$\\
       	$b \leftarrow b - u$ \Comment Update remaining budget
    }
    
\caption{Active Learning Approach to Selecting Documents for Annotation}
\label{algorithm:AL_hate_speech}
\end{algorithm}

The active learning-based process for developing a hate speech dataset is described in {\bf Algorithm \ref{algorithm:AL_hate_speech}}. Given seed documents, an initial machine learning model is trained using those seed documents (Lines 1 - 4). The iterative process of AL is conducted in the loop (Lines 5 - 11), where a trained model selects documents for human annotation using the document selection criteria as discussed earlier (Lines 7 - 9).  The model is re-trained using both existing documents and those newly annotated documents (Line 10). This continues until the budget is exhausted. 

\section{The Dataset}
\label{section:Constructing Hate Speech Dataset}
We create our hate speech dataset via pooling (Section \ref{section:pooling_algorithm}). We use five models to detect hate speech: logistic regression, naive bayes, an LSTM-based model \cite{badjatiya2017deep}, a BiLSTM-based model \cite{agrawal2018deep}, and a model based on BERT \cite{devlin2018bert}. For both LSTM and BiLSTM models, we use the improved versions of these models reported by \citet{arango2019hate}. Models are trained across five prior hate speech datasets:  \citet{waseem-hovy-2016-hateful, davidson2017automated, grimminger2021hate, golbeck2017large, basile-etal-2019-semeval}. 
Models are trained for binary classification, with datasets binarized accordingly. 

{\bf User statistics.} \citet{arango2019hate} note that 65\% of hate speech annotated in \citet{waseem-hovy-2016-hateful}'s dataset comes from two Twitter users, suggesting coverage is not representative of online hate. In contrast, our dataset is constructed based on a random sample of tweets from the Twitter API, then selected for annotation via pooling. Out of 9,534 unique users included in our dataset, only 7 produce more than 2 tweets, with at most 15 tweets by one user and 28 total tweets across the other 6.

{\bf Annotation Process}. We collect the annotations in two phases: a set of iterative pilot experiments (4,668 labeled tweets), followed by consistent use of our final annotation design (4,999 labeled tweets). The pilot data revealed two key issues: i) many pornography-related tweets were annotated as hateful, and ii) sometimes annotators did not highlight both the language and targets of hate speech. We thus updated the annotation guidelines accordingly. Although we release the annotated labels collected from both of these phases, we only analyze and report results using the 4,999 annotations collected from the final annotation design. 

Since hate speech is highly subjective and complex, prior work \citep{sanguinetti2018italian, assimakopoulos2020annotating} has argued that simply asking the annotators to perform binary classification of hate vs. non-hate of an online post is unlikely to be reliable. Thus, a more complex, hierarchical annotation scheme may be needed. For example,  \citet{sanguinetti2018italian} break down the task of annotating hate speech into two sub-tasks. Based on this, we design an annotation scheme that is hierarchical in nature and decomposes the hate annotation task into a set of simpler sub-tasks corresponding to the definitional criteria that must be met for a post to constitute hate speech. Our annotation process is highly structured, including use of term definitions, annotation sub-tasks, types of hate (derogatory language or inciting violence), demographic categories, required rationales for labeling decisions), and a self-consistency test. 

\textbf{Dataset Properties}. 
\label{sec:props}
If an annotator identifies (implicit or explicit) language that is derogatory toward or incites violence against a demographic group, and identifies an (implicit or explicit) target group, we can infer from these annotations that a tweet is hateful.  In addition, we also ask the annotator (step 6) directly to judge whether the tweet is hateful.  This allows us to perform a valuable {\em self-consistency} check \cite{Zhang-sigir14} on the annotator, which is especially valuable for subjective tasks in which inter-annotator agreement is naturally lower.  We observe that $94.5$\% of the time, annotators provide final judgments that are self-consistent with their sub-task annotations of 1) hateful language and 2) demographic targets. For the other $5.5$\% of tweets (274), half of these are labeled as hateful but the annotator does not select either corresponding targets or actions of hate or both, and some annotators still mark pornographic content as hateful. We discard these 274 tweets, leaving 4,725 in our final dataset. 

Comparable to prior work  \citep{ousidhoum2019multilingual, ross2017measuring, kwok2013locate, del2017hate}, we observe
inter-annotator agreement of Fleiss $\kappa=0.16$ (raw agreement of 72\%) using hate or non-hate binary label of this dataset. Given task subjectivity, this is why self-consistency checks are so important for ensuring data quality. It is also noteworthy that the $\kappa$ score for inter-annotator agreement is not comparable across studies because $\kappa$ depends on the underlying class distribution of the annotators \citep{feinstein1990high}. In fact, $\kappa$ can be very low even though there is a high level of observed agreement \citep{uebersax1988validity}. There is in fact debate in the statistical community about how to calculate the expected agreement score \citep{feinstein1990high} while calculating $\kappa$.  \citet{gwet2002kappa} advocate for a more robust inter-annotator agreement score and propose $AC_1$, with interpretation similar to the $\kappa$ statistic. \citet{Jikeli2021Anti} report both Cohen $\kappa$ and Gwet's $AC_1$ in their hate speech dataset and find that the skewed distribution of prevalence of hate speech severely affects $\kappa$ but not Gwet’s $AC_1$. We observe $AC_1 = 0.58$ agreement in binary hate vs.\ non-hate labels. Regarding inter-annotator agreement statistics for rationales supporting labeling decisions, please see our supplementary material \cite{rahman21-neurips21-supplementary}.

{\bf Relative Coverage}. Since we do not use any ``hate words'' to filter tweets, our hate speech dataset includes hateful posts from both C1 and C2 categories (Figure \ref{Figure:landscape_of_hate_speech}). To quantify the coverage of hate in a dataset, we quantify the percentage of annotated hateful posts that do not contain known hate words as {\em Relative Coverage}: $100 \times (N_T - N_{TH})/N_{TH}$, 
where $N_T$ is the total number of hateful posts in the dataset, and $N_{TH}$ is the total number of hateful posts containing known hate words. For the purpose of analysis only, we use \citet{founta2018large}'s hate word list\footnote{The union of~ \url{https://www.hatebase.org} ~and~ \url{https://www.noswearing.com/dictionary}.}. Any hate speech dataset restricted to tweets having these hate words would have $N_T = N_{TH}$, yielding a relative coverage of 0\%, as shown in {\bf Table \ref{Table:data_relative_coverage}}. Because \citet{founta2018large}\footnote{\citet{founta2018large} report 80K tweets but their online dataset contains 100K. They confirm (personal communication) collecting another 20K after publication. The 92K we report reflects removal of 8K duplicates.} use tweets from both the keyword-based search and random sampling, the relative coverage of their dataset is 10.40\%, vs.\ our 14.60\%.    

\begin{table}
\caption{Comparison of Datasets. Pooling is seen to achieve the best balance of prevalence vs.\ relative coverage compared to the keyword-based search (KS) and random sampling (RS) methods.}
\vspace{-1em}
\begin{center}\scalebox{0.9}{
\begin{tabular}{ c r c r r}
\hline
 {\bf Dataset} & {\bf Size} & {\bf Method} & {\bf Prevalence} & {\bf Rel.\ Cov.} \\
 \hline
 G\&K \cite{grimminger2021hate} & 2,999 &  KS & 11.7\% & 0.0\%\\
    \hline
 W\&H \cite{waseem-hovy-2016-hateful} & 16,914 & KS & {\bf 31\%} & 0.0\%\\
 \hline

  G.\ et al.\ \cite{golbeck2017large} & 19,838 &  KS & 15\% & 0.0\%\\
    
\hline
 D.\ et al.\ \cite{davidson2017automated} & 24,783 & KS &  5.77\% & 0.0\% \\
 
 \hline
 F.\ et al.\ \cite{founta2018large} & 91,951 & KS and RS & 4.96\% & 10.40\%\\
 \hline

 {\bf Our approach} & 4,999 &  Pooling &  14.12\% &  {\bf 14.60}\%\\
   
 \hline
 \end{tabular}}
\end{center}
\label{Table:data_relative_coverage}
\vspace{-1em}
\end{table}

{\bf Prevalence}. 
Table \ref{Table:data_relative_coverage} 
shows that the prevalence of our hate speech dataset is 14.12\%, higher than the datasets created by \citet{founta2018large} where the authors cover a broad range of hate speech by combining random sampling with the keyword-based search. In addition to that, the prevalence of our hate speech dataset exceeds that for \citet{davidson2017automated} and is comparable to the datasets created by \citet{grimminger2021hate} and \citet{golbeck2017large}, where the authors employ keyword-based search. Again, prevalence can be inflated by only labeling tweets with known strong hate words. 

\section{Model Benchmarking}
We benchmark three recent hate speech detection models: LSTM-based \citep{badjatiya2017deep}, BiLSTM-based,  \citep{agrawal2018deep}, and a model based on BERT \citep{devlin2018bert}. For both LSTM and BiLSTM models, we do not use the original versions of these models, but rather the the corrected versions reported by \citet{arango2019hate}. 


{\bf Train and Test Sets}. To compare the models, we perform an 80/20 train/test split of our dataset. To maintain class ratios in this split, we apply stratified sampling. For analysis, we also partition test results by presence/absence of known ``hate words'', using \citet{founta2018large}'s hate word list. Train and test set splits contain 3,779 and  946 tweets, respectively. In the test set, the number of normal tweets without hate words and with hate words is 294 and 518 respectively, whereas the number of hateful tweets without hate words and with hate words is 15 and 119, respectively. 

{\bf Results and Discussion}. Performance of the models on the test sets in terms of precision (P), recall (R) and $F_1$ is shown in {\bf Table \ref{Table:Benchamark_results}}. Following \citet{arango2019hate}, we also report these three performance metrics for both hateful and non-hateful class (Table \ref{Table:Benchamark_results}, Column -  Class). We discuss the performance of the models considering the two types of test sets as described above. 

\begin{table}[h]
\caption{Hate classification accuracy of models (bottom 3 rows) provides further support for prior work \cite{arango2019hate,arango2020hate}'s assertion that ``hate speech detection is not as easy as we think.''}
\vspace{-1em}
\begin{center}\scalebox{0.93}{
\begin{tabular}{ l | c | c c c | c c c} 
 \hline
        &       & \multicolumn{3}{c}{ \textbf{Without Hate Words}} & \multicolumn{3}{c}{\textbf{With Hate Words}} \\ \hline  
 {\bf Method} & \!\!{\bf Class}\!\! & {\bf P} & {\bf R} & {\bf F1} & {\bf P} & {\bf R} & {\bf F1} \\
 
\hline
BiLSTM \citep{agrawal2018deep}\!\! & Non-Hate & \cellcolor{blue!45}95.34 & \cellcolor{blue!45}97.61 & \cellcolor{blue!45}96.47 & \cellcolor{blue!35}84.98 & \cellcolor{blue!35}86.29 & \cellcolor{blue!35}85.63 \\
LSTM \citep{badjatiya2017deep} & Non-Hate & \cellcolor{blue!45}95.60 & \cellcolor{blue!45}96.25 & \cellcolor{blue!45}95.93 & \cellcolor{blue!35}84.95 & \cellcolor{blue!35}89.38 & \cellcolor{blue!35}87.11 \\
BERT \citep{devlin2018bert} & Non-Hate & \cellcolor{blue!45}96.25 & \cellcolor{blue!45}96.25 & \cellcolor{blue!45}96.25 & \cellcolor{blue!35}88.84 & \cellcolor{blue!35}83.01 & \cellcolor{blue!35}85.82 \\
\hline
\hline
BiLSTM \citep{agrawal2018deep}\!\! & Hate & \cellcolor{blue!15}12.50 & \cellcolor{blue!15}6.666 & \cellcolor{blue!15}8.695 & \cellcolor{blue!25}36.03 & \cellcolor{blue!25}33.61 & \cellcolor{blue!25}34.78 \\
LSTM \citep{badjatiya2017deep} & Hate & \cellcolor{blue!15}15.38 & \cellcolor{blue!15}13.33 & \cellcolor{blue!15}14.28 & \cellcolor{blue!25}40.21 & \cellcolor{blue!25}31.09 & \cellcolor{blue!25}35.07 \\
BERT \citep{devlin2018bert} & Hate & \cellcolor{blue!15}26.66 & \cellcolor{blue!15}26.66 & \cellcolor{blue!15}26.66 & \cellcolor{blue!25}42.48 & \cellcolor{blue!25}54.62 & \cellcolor{blue!25}47.79 \\
\hline
\end{tabular}}
\label{Table:Benchamark_results}
\end{center}
\end{table}

{\bf Case  I. Without Hate Words}. Intuition is that correctly classifying a document as hateful when canonical hate words are absent is comparatively difficult for the models. This is because models typically learn training weights on the predictive features, and for hate classification, those hate words are the most vital predictive features which are absent in this setting. By observing $F_1$, we can find that for the hateful class, all these models, including BERT, provide below-average performance ($F_1$  $\leq$ 26.66\% ). In other words, the number of false negatives is very high in this category. On the other hand, when there are no hate words, these same models provide very high performance on the non-hateful class, with $F_1$ $\approx$ 96\%. 

{\bf Case II. With Hate Words}. In this case, hate words exist in documents, and models are more effective than the previous case in their predictive performance (Table \ref{Table:Benchamark_results}, Columns 6, 7, and 8) on the hateful class. For example, $F_1$ of BERT on the hateful class has improved from 26.66\% (Column 3) to 47.79\% (Column 8). This further shows the relative importance of hate words as predictive features for these models. On the other hand, for classifying documents as non-hateful when there are hate words in documents, the performance of models are comparatively low (average $F_1$ across models is $\approx$ 86\%) vs.\ when hate words are absent (average $F_1$ across models is $\approx$ 96\%). 

\textbf{Table \ref{Table:Benchamark_results}} shows clearly that models for hate speech detection struggle to correctly classify documents as hateful when ``hate words'' are absent. While prior work by \citet{arango2019hate} already confirms that there is an issue of overestimation of the performance of current state-of-the-art approaches for hate speech detection, our experimental analysis suggests that models need further help to cover both categories of hate speech (C1 and C2 from Figure \ref{Figure:landscape_of_hate_speech}).

\section{Active Learning vs.\ Pooling}
\label{section:active_learning}

While pooling alone is used to select tweets inclusion and annotation in our dataset, we also report a retrospective evaluation of pooling vs.\ active learning (AL). In this retrospective evaluation, AL is restricted to the set of documents selected by pooling, rather than the original full, random sample of Twitter from which the pooling dataset is derived. Because of this constraint, the retrospective AL results we report here are likely lower than what might be achieved when AL run instead on the full dataset, given the larger set of documents that would be available to choose from for annotation.


\noindent{\bf Experimental Setup}. Each iteration of AL selects one document to be judged next (i.e., batch size $u$ = $1$). The total allotted budget $b$ is set to the size of the hate speech dataset constructed by pooling ($b=4,725$). For the document selection, we report SAL and CAL. As a baseline, we also report a random document selection strategy akin to how traditional supervised learning is performed on a presumably random sample of annotated data. Following prior work's nomenclature, we refer to this as {\em simple passive learning} (SPL) \cite{cormack_autonomy_2015}. As seed documents, we randomly select 5 hateful and 5 non-hateful documents from our hate speech dataset constructed using pooling. 
 
\begin{figure*}[t]
\centerline{\includegraphics[width=1.0\textwidth]{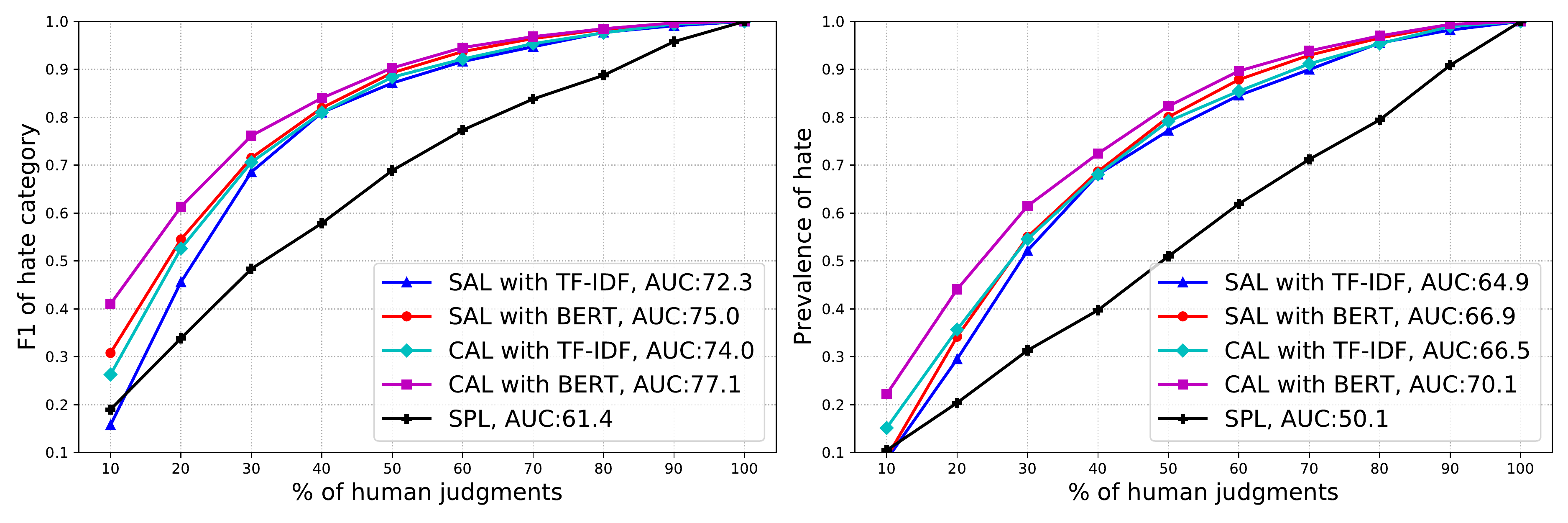}} 
\caption{Left plot shows human judging cost (x-axis) vs. F1 classification accuracy (y-axis) for hybrid human-machine judging. Right plot shows human judging cost vs. prevalence of hate speech for human-only judging of documents. The \% of human judgments on x-axis is wrt.\ the number documents in the hate speech data constructed by the pooling-based approach.}
\vspace{-1em}
\label{Figure:AL_performance}
\end{figure*}

\noindent{\bf Experimental Analysis}. We present the results ({\bf Figure \ref{Figure:AL_performance}}) as plots showing the cost vs.\ effectiveness of each method being evaluated at different cost points  (corresponding to varying evaluation budget sizes). We also report Area Under Curve (AUC) across all cost points, approximated via the Trapezoid rule. To report the effectiveness, Figure \ref{Figure:AL_performance} presents $F_1$ performance (left plot), and prevalence (right plot) results of the three document selection approaches: SPL, SAL, and CAL, along with two feature representations: TF-IDF and contextual embedding from BERT. 

While reporting $F_1$ of our AL classifier (Figure \ref{Figure:AL_performance}, left plot), following prior work \citep{nguyen2015combining}, we consider both human judgments and machine predictions to compute $F_1$. For example, when we collect 20\% human judgments from annotators, the remaining 80\% is predicted by the AL classifier (i.e., machine prediction). Then those two sets (human judgments and machine prediction) are combined to report the final $F_1$ score. However, to report the prevalence of hate (right plot), only human judgments are considered because we need the actual labels of documents, not the machine predictions here. 

{\bf Active vs.\ Passive Learning}. By comparing active learning (SAL and CAL) methods against passive learning (SPL) method in terms of $F_1$, we find that SAL and CAL significantly outperform SPL in terms of AUC. The observation holds for both types of feature representations of documents. The only exception we can see that at 10\% human judgments, $F_1$ of SPL is slightly better than SAL with TF-IDF. Nevertheless, after that point of 10\% human judgments, SAL with TF-IDF consistently outperforms  SPL in terms of $F_1$. Similarly, if we consider the prevalence of hate (right plot), active learning exceeds passive learning by a large margin in AUC.     

{\bf CAL vs. SAL}. From Figure \ref{Figure:AL_performance}, we can see that CAL consistently provides better performance than SAL both in terms of $F_1$ and prevalence. In fact, when the allotted budget is very low (e.g., budget $\le$ 20\% of human judgments), the performance difference in terms of $F_1$ and prevalence between CAL and SAL considering the underlying feature representation is very high. 

{\bf TF-IDF vs. BERT embedding}. It is evident from Figure \ref{Figure:AL_performance} that the  contextual embedding from BERT for document representation provides a significant performance boost over the TF-IDF-based representation. SAL and CAL both achieve better performance when the document is represented using BERT. Furthermore, we can find that for a low-budget situation (e.g., budget $\leq$  20\% of human judgments), CAL with BERT provides the best performance.  

{\bf Judging cost vs. Performance} We also investigate whether the AL-based approach can provide better prevalence and classification accuracy at a lower cost than the pooling-based approach. From Figure \ref{Figure:AL_performance}, we can find that the AL-based approach achieves $F_1 \ge 0.9$ (left plot) and finds 80\% of hateful documents (right plot) by only annotating 50\% of the original documents judged via pooling.

\section{Conclusion \& Future Work}
The success of an automated hate speech detection model largely depends on the quality of the underlying dataset upon which that model is trained on. However, keyword-based filtering approaches \citep{davidson2017automated, waseem-hovy-2016-hateful, grimminger2021hate} do not provide broad coverage of hate, and random sampling-based approaches \citep{kennedy2018gab, de-gibert-etal-2018-hate} suffer from low prevalence of hate. We propose an approach that adapts {\it pooling} from IR \citep{sanderson2010test} by  training multiple classifiers on prior hate speech datasets and combining that with random sampling to improve both the prevalence and the coverage of hate speech in the final constructed dataset. 

Using the pooling technique, we share a new benchmark dataset for hate speech detection on Twitter. Results show that the hate speech dataset developed by applying the proposed pooling-based method achieves better relative coverage (14.60\%) than the hate speech dataset constructed by \citet{founta2018large}  that combines random sampling with the keyword-based search (Table \ref{Table:data_relative_coverage}). Furthermore, the prevalence of hate-related content in our hate speech dataset is comparable to many prior keyword-based approaches \citep{davidson2017automated, grimminger2021hate, golbeck2017large}. We also show a dramatic drop in accuracy of existing detection models \citep{devlin2018bert, badjatiya2017deep, agrawal2018deep} when tested on these broader forms of hate. 

An important limitation of the pooling approach is that it relies on prior hate speech datasets. That said, though the datasets used to train our pooling prediction models may lack diverse tweets, the trained models still learn correlations between hate labels and all vocabulary in the dataset. This enables pooling to identify some hateful tweets (for inclusion in our dataset) that lack known hate words, though this prediction task is clearly challenging (as the bottom left of Table \ref{Table:Benchamark_results} indicates). In general, the purpose of pooling in IR is that diverse models better identify potentially relevant content for human annotation. In the translational use of pooling we present for hate speech, training a cross product of different models across different prior datasets similarly promotes diversity and ``wisdom of crowds'' in identifying potentially hateful content for inclusion and annotation.

To sidestep this reliance on prior hate speech datasets for pooling, we also present an alternative approach that utilizes active learning \citep{settles2012active} to develop the hate speech dataset. Empirical analysis on the hate speech dataset constructed via the pooling-based approach suggests that by only judging 50\% of the originally annotated documents, it is possible to  find 80\% of hateful documents with an $F_1$ accuracy of $\ge 0.9$ (Figure \ref{Figure:AL_performance}) via our AL-based approach. 

Like HateXplain \cite{mathew2020hatexplain}, our collection of rationales as well as labels creates the potential for explainable modeling \cite{strout2019human} and dual-supervision \cite{zhang2016rationale}. However, while HateXplain collects rationales only for the overall labeling decision, our collection of rationales for different annotation sub-tasks creates intriguing possibilities for dual-supervision \citep{xia2017dual} and explanations across different types of evidence contributing to the overall labeling decision \citep{lehman2019inferring}. In particular, annotator rationales identify i) derogatory language, ii) language inciting violence, and iii) the targeted demographic group. As future work, we plan to design dual-supervised \citep{xia2017dual} and/or explainable \citep{deyoung2019eraser} machine learning models that can incorporate the annotators' rationales collected in our hate speech dataset.   

Our definition of hate speech and annotation process assumed that hate speech is composed of two parts: 1) language that is derogatory or inciting violence against 2) a target demographic group. However, an interesting question is whether perceptions of hate differ based on the demographic group in question, e.g., a given derogatory expression toward a political group might be deemed acceptable while the same expression to a racial or ethnic group might be construed as hate speech. HateCheck \cite{rottger-etal-2021-hatecheck}'s differentiation of general templates of hateful language vs. template instantiations for specific demographic targets could provide a nice framework to further investigate this.

Although our annotation  process was highly structured, we ultimately still produced binary labels for hate speech, lacking nuance of finer-grained ordinal scales or categories. It would be interesting to further explore our structured annotation process with such finer-grained scales or categories.

Recent years have brought greater awareness that machine learning datasets (as well as models) can cause harm as well as good \citep{shankar2017no, buolamwini2018gender, noble2018algorithms, eubanksautomating}. For example, harm could come from deploying a machine learning model without considering, mitigating, and/or documenting \cite{gebru2018datasheets, bender2018data, boyd2021datasheets} limitations of its underlying training data, 
such as the risk of racial bias in hate speech annotations \cite{sap2019risk,wich2020impact,davidson2019racial}. 
Both researchers and practitioners of hate speech detection should be well-informed about such potential limitations and risks of any constructed hate speech dataset (including ours) and exercise care and good judgment while deploying a hate speech detection system trained on such a dataset. 

We know of no prior work studying the effect of frequent exposure to hate speech on the well-being of human annotators. However, prior evidence suggests that exposure to online abuse has serious consequences on the mental health of  workers \citep{ybarra2006examining}. Studies by 
\citet{boeckmann2002hate} and \citet{ leets2002experiencing} to understand how people experience hate speech found that low self-esteem, symptoms of trauma exposure, etc., are associated with repeated exposure. The Linguistic Data Consortium (LDC) reported its annotators experiencing nightmares and other overwhelming feelings from labeling news articles \cite{strassel2000quality}. We suggest more attention be directed toward the well-being of the annotators \cite{steiger2021psychological}.

\textbf{Acknowledgements}. We thank the many talented Amazon Mechanical Turk workers who contributed to our study and made it possible. This research was supported in part by Wipro (HELIOS), the Knight Foundation, the Micron Foundation, and Good Systems (\url{https://goodsystems.utexas.edu}), 
a UT Austin Grand Challenge to develop responsible AI technologies. Our opinions are our own.


\bibliographystyle{ACM-Reference-Format}
\bibliography{sample-base}


\begin{thebibliography}{124}


\ifx \showCODEN    \undefined \def \showCODEN     #1{\unskip}     \fi
\ifx \showDOI      \undefined \def \showDOI       #1{#1}\fi
\ifx \showISBNx    \undefined \def \showISBNx     #1{\unskip}     \fi
\ifx \showISBNxiii \undefined \def \showISBNxiii  #1{\unskip}     \fi
\ifx \showISSN     \undefined \def \showISSN      #1{\unskip}     \fi
\ifx \showLCCN     \undefined \def \showLCCN      #1{\unskip}     \fi
\ifx \shownote     \undefined \def \shownote      #1{#1}          \fi
\ifx \showarticletitle \undefined \def \showarticletitle #1{#1}   \fi
\ifx \showURL      \undefined \def \showURL       {\relax}        \fi
\providecommand\bibfield[2]{#2}
\providecommand\bibinfo[2]{#2}
\providecommand\natexlab[1]{#1}
\providecommand\showeprint[2][]{arXiv:#2}

\bibitem[\protect\citeauthoryear{Agrawal and Awekar}{Agrawal and
  Awekar}{2018}]%
        {agrawal2018deep}
\bibfield{author}{\bibinfo{person}{Sweta Agrawal} {and} \bibinfo{person}{Amit
  Awekar}.} \bibinfo{year}{2018}\natexlab{}.
\newblock \showarticletitle{Deep learning for detecting cyberbullying across
  multiple social media platforms}. In \bibinfo{booktitle}{\emph{European
  Conference on Information Retrieval}}. Springer, \bibinfo{pages}{141--153}.
\newblock


\bibitem[\protect\citeauthoryear{Al~Kuwatly, Wich, and Groh}{Al~Kuwatly
  et~al\mbox{.}}{2020}]%
        {al2020identifying}
\bibfield{author}{\bibinfo{person}{Hala Al~Kuwatly},
  \bibinfo{person}{Maximilian Wich}, {and} \bibinfo{person}{Georg Groh}.}
  \bibinfo{year}{2020}\natexlab{}.
\newblock \showarticletitle{Identifying and measuring annotator bias based on
  annotators’ demographic characteristics}. In
  \bibinfo{booktitle}{\emph{Proceedings of the Fourth Workshop on Online Abuse
  and Harms}}. \bibinfo{pages}{184--190}.
\newblock


\bibitem[\protect\citeauthoryear{Arango, P{\'e}rez, and Poblete}{Arango
  et~al\mbox{.}}{2019}]%
        {arango2019hate}
\bibfield{author}{\bibinfo{person}{Aym{\'e} Arango}, \bibinfo{person}{Jorge
  P{\'e}rez}, {and} \bibinfo{person}{Barbara Poblete}.}
  \bibinfo{year}{2019}\natexlab{}.
\newblock \showarticletitle{Hate speech detection is not as easy as you may
  think: A closer look at model validation}. In
  \bibinfo{booktitle}{\emph{Proceedings of the 42nd international acm sigir
  conference on research and development in information retrieval}}.
  \bibinfo{pages}{45--54}.
\newblock


\bibitem[\protect\citeauthoryear{Arango, P{\'e}rez, and Poblete}{Arango
  et~al\mbox{.}}{2020}]%
        {arango2020hate}
\bibfield{author}{\bibinfo{person}{Aym{\'e} Arango}, \bibinfo{person}{Jorge
  P{\'e}rez}, {and} \bibinfo{person}{Barbara Poblete}.}
  \bibinfo{year}{2020}\natexlab{}.
\newblock \showarticletitle{Hate speech detection is not as easy as you may
  think: A closer look at model validation (extended version)}.
\newblock \bibinfo{journal}{\emph{Information Systems}} (\bibinfo{year}{2020}),
  \bibinfo{pages}{101584}.
\newblock


\bibitem[\protect\citeauthoryear{Aroyo and Welty}{Aroyo and Welty}{2015}]%
        {aroyo2015truth}
\bibfield{author}{\bibinfo{person}{Lora Aroyo} {and} \bibinfo{person}{Chris
  Welty}.} \bibinfo{year}{2015}\natexlab{}.
\newblock \showarticletitle{Truth is a lie: Crowd truth and the seven myths of
  human annotation}.
\newblock \bibinfo{journal}{\emph{AI Magazine}} \bibinfo{volume}{36},
  \bibinfo{number}{1} (\bibinfo{year}{2015}), \bibinfo{pages}{15--24}.
\newblock


\bibitem[\protect\citeauthoryear{Artstein and Poesio}{Artstein and
  Poesio}{2008}]%
        {artstein2008inter}
\bibfield{author}{\bibinfo{person}{Ron Artstein} {and} \bibinfo{person}{Massimo
  Poesio}.} \bibinfo{year}{2008}\natexlab{}.
\newblock \showarticletitle{Inter-coder agreement for computational
  linguistics}.
\newblock \bibinfo{journal}{\emph{Computational Linguistics}}
  \bibinfo{volume}{34}, \bibinfo{number}{4} (\bibinfo{year}{2008}),
  \bibinfo{pages}{555--596}.
\newblock


\bibitem[\protect\citeauthoryear{Assimakopoulos, Muskat, van~der Plas, and
  Gatt}{Assimakopoulos et~al\mbox{.}}{2020}]%
        {assimakopoulos2020annotating}
\bibfield{author}{\bibinfo{person}{Stavros Assimakopoulos},
  \bibinfo{person}{Rebecca~Vella Muskat}, \bibinfo{person}{Lonneke van~der
  Plas}, {and} \bibinfo{person}{Albert Gatt}.} \bibinfo{year}{2020}\natexlab{}.
\newblock \showarticletitle{Annotating for hate speech: The MaNeCo corpus and
  some input from critical discourse analysis}.
\newblock \bibinfo{journal}{\emph{arXiv preprint arXiv:2008.06222}}
  (\bibinfo{year}{2020}).
\newblock


\bibitem[\protect\citeauthoryear{Association et~al\mbox{.}}{Association
  et~al\mbox{.}}{2013}]%
        {american2013diagnostic}
\bibfield{author}{\bibinfo{person}{American~Psychiatric Association}
  {et~al\mbox{.}}} \bibinfo{year}{2013}\natexlab{}.
\newblock \bibinfo{booktitle}{\emph{Diagnostic and statistical manual of mental
  disorders (DSM-5{\textregistered})}}.
\newblock \bibinfo{publisher}{American Psychiatric Pub}.
\newblock


\bibitem[\protect\citeauthoryear{Badjatiya, Gupta, and Varma}{Badjatiya
  et~al\mbox{.}}{2019}]%
        {badjatiya2019stereotypical}
\bibfield{author}{\bibinfo{person}{Pinkesh Badjatiya}, \bibinfo{person}{Manish
  Gupta}, {and} \bibinfo{person}{Vasudeva Varma}.}
  \bibinfo{year}{2019}\natexlab{}.
\newblock \showarticletitle{Stereotypical bias removal for hate speech
  detection task using knowledge-based generalizations}. In
  \bibinfo{booktitle}{\emph{The World Wide Web Conference}}.
  \bibinfo{pages}{49--59}.
\newblock


\bibitem[\protect\citeauthoryear{Badjatiya, Gupta, Gupta, and Varma}{Badjatiya
  et~al\mbox{.}}{2017}]%
        {badjatiya2017deep}
\bibfield{author}{\bibinfo{person}{Pinkesh Badjatiya},
  \bibinfo{person}{Shashank Gupta}, \bibinfo{person}{Manish Gupta}, {and}
  \bibinfo{person}{Vasudeva Varma}.} \bibinfo{year}{2017}\natexlab{}.
\newblock \showarticletitle{Deep learning for hate speech detection in tweets}.
  In \bibinfo{booktitle}{\emph{Proceedings of the 26th International Conference
  on World Wide Web Companion}}. \bibinfo{pages}{759--760}.
\newblock


\bibitem[\protect\citeauthoryear{Basile, Bosco, Fersini, Nozza, Patti,
  Rangel~Pardo, Rosso, and Sanguinetti}{Basile et~al\mbox{.}}{2019}]%
        {basile-etal-2019-semeval}
\bibfield{author}{\bibinfo{person}{Valerio Basile}, \bibinfo{person}{Cristina
  Bosco}, \bibinfo{person}{Elisabetta Fersini}, \bibinfo{person}{Debora Nozza},
  \bibinfo{person}{Viviana Patti}, \bibinfo{person}{Francisco~Manuel
  Rangel~Pardo}, \bibinfo{person}{Paolo Rosso}, {and} \bibinfo{person}{Manuela
  Sanguinetti}.} \bibinfo{year}{2019}\natexlab{}.
\newblock \showarticletitle{{S}em{E}val-2019 Task 5: Multilingual Detection of
  Hate Speech Against Immigrants and Women in {T}witter}. In
  \bibinfo{booktitle}{\emph{Proceedings of the 13th International Workshop on
  Semantic Evaluation}}. \bibinfo{publisher}{Association for Computational
  Linguistics}, \bibinfo{address}{Minneapolis, Minnesota, USA},
  \bibinfo{pages}{54--63}.
\newblock
\urldef\tempurl%
\url{https://doi.org/10.18653/v1/S19-2007}
\showDOI{\tempurl}


\bibitem[\protect\citeauthoryear{Bender and Friedman}{Bender and
  Friedman}{2018}]%
        {bender2018data}
\bibfield{author}{\bibinfo{person}{Emily~M Bender} {and} \bibinfo{person}{Batya
  Friedman}.} \bibinfo{year}{2018}\natexlab{}.
\newblock \showarticletitle{Data statements for natural language processing:
  Toward mitigating system bias and enabling better science}.
\newblock \bibinfo{journal}{\emph{Transactions of the Association for
  Computational Linguistics}}  \bibinfo{volume}{6} (\bibinfo{year}{2018}),
  \bibinfo{pages}{587--604}.
\newblock


\bibitem[\protect\citeauthoryear{Boeckmann and Liew}{Boeckmann and
  Liew}{2002}]%
        {boeckmann2002hate}
\bibfield{author}{\bibinfo{person}{Robert~J Boeckmann} {and}
  \bibinfo{person}{Jeffrey Liew}.} \bibinfo{year}{2002}\natexlab{}.
\newblock \showarticletitle{Hate speech: Asian American students’ justice
  judgments and psychological responses}.
\newblock \bibinfo{journal}{\emph{Journal of Social Issues}}
  \bibinfo{volume}{58}, \bibinfo{number}{2} (\bibinfo{year}{2002}),
  \bibinfo{pages}{363--381}.
\newblock


\bibitem[\protect\citeauthoryear{Bohra, Vijay, Singh, Akhtar, and
  Shrivastava}{Bohra et~al\mbox{.}}{2018}]%
        {bohra-etal-2018-dataset}
\bibfield{author}{\bibinfo{person}{Aditya Bohra}, \bibinfo{person}{Deepanshu
  Vijay}, \bibinfo{person}{Vinay Singh}, \bibinfo{person}{Syed~Sarfaraz
  Akhtar}, {and} \bibinfo{person}{Manish Shrivastava}.}
  \bibinfo{year}{2018}\natexlab{}.
\newblock \showarticletitle{A Dataset of {H}indi-{E}nglish Code-Mixed Social
  Media Text for Hate Speech Detection}. In
  \bibinfo{booktitle}{\emph{Proceedings of the Second Workshop on Computational
  Modeling of People{'}s Opinions, Personality, and Emotions in Social Media}}.
  \bibinfo{publisher}{Association for Computational Linguistics},
  \bibinfo{address}{New Orleans, Louisiana, USA}, \bibinfo{pages}{36--41}.
\newblock
\urldef\tempurl%
\url{https://doi.org/10.18653/v1/W18-1105}
\showDOI{\tempurl}


\bibitem[\protect\citeauthoryear{Boyd}{Boyd}{2021}]%
        {boyd2021datasheets}
\bibfield{author}{\bibinfo{person}{Karen~L Boyd}.}
  \bibinfo{year}{2021}\natexlab{}.
\newblock \showarticletitle{Datasheets for Datasets help ML Engineers Notice
  and Understand Ethical Issues in Training Data}.
\newblock \bibinfo{journal}{\emph{Proceedings of the ACM on Human-Computer
  Interaction}} \bibinfo{volume}{5}, \bibinfo{number}{CSCW2}
  (\bibinfo{year}{2021}), \bibinfo{pages}{1--27}.
\newblock


\bibitem[\protect\citeauthoryear{Buckley, Dimmick, Soboroff, and
  Voorhees}{Buckley et~al\mbox{.}}{2006}]%
        {buckley_bias_2006}
\bibfield{author}{\bibinfo{person}{Chris Buckley}, \bibinfo{person}{Darrin
  Dimmick}, \bibinfo{person}{Ian Soboroff}, {and} \bibinfo{person}{Voorhees}.}
  \bibinfo{year}{2006}\natexlab{}.
\newblock \showarticletitle{Bias and the limits of pooling}. In
  \bibinfo{booktitle}{\emph{Proceedings of the 29th annual international {ACM}
  {SIGIR} conference on {Research} and development in information retrieval -
  {SIGIR} '06}}. \bibinfo{publisher}{ACM Press}, \bibinfo{address}{Seattle,
  Washington, USA}, \bibinfo{pages}{619}.
\newblock
\showISBNx{978-1-59593-369-0}
\urldef\tempurl%
\url{https://doi.org/10.1145/1148170.1148284}
\showDOI{\tempurl}


\bibitem[\protect\citeauthoryear{Buolamwini and Gebru}{Buolamwini and
  Gebru}{2018}]%
        {buolamwini2018gender}
\bibfield{author}{\bibinfo{person}{Joy Buolamwini} {and}
  \bibinfo{person}{Timnit Gebru}.} \bibinfo{year}{2018}\natexlab{}.
\newblock \showarticletitle{Gender shades: Intersectional accuracy disparities
  in commercial gender classification}. In \bibinfo{booktitle}{\emph{Conference
  on fairness, accountability and transparency}}. PMLR,
  \bibinfo{pages}{77--91}.
\newblock


\bibitem[\protect\citeauthoryear{Chaudhry and Lease}{Chaudhry and
  Lease}{2020}]%
        {chaudhry20-arxiv}
\bibfield{author}{\bibinfo{person}{Prateek Chaudhry} {and}
  \bibinfo{person}{Matthew Lease}.} \bibinfo{year}{2020}\natexlab{}.
\newblock \bibinfo{booktitle}{\emph{You Are What You Tweet: Profiling Users by
  Past Tweets to Improve Hate Speech Detection}}.
\newblock \bibinfo{type}{{T}echnical {R}eport}.
  \bibinfo{institution}{University of Texas at Austin}.
\newblock
\urldef\tempurl%
\url{http://arxiv.org/abs/arXiv:2012.09090}
\showURL{%
\tempurl}
\newblock
\shownote{arXiv:2012.09090.}


\bibitem[\protect\citeauthoryear{Chen}{Chen}{2011}]%
        {chen2011detecting}
\bibfield{author}{\bibinfo{person}{Ying Chen}.}
  \bibinfo{year}{2011}\natexlab{}.
\newblock \showarticletitle{Detecting offensive language in social medias for
  protection of adolescent online safety}.
\newblock  (\bibinfo{year}{2011}).
\newblock


\bibitem[\protect\citeauthoryear{Coalition}{Coalition}{2013}]%
        {resilenceGuild2015}
\bibfield{author}{\bibinfo{person}{The~Technology Coalition}.}
  \bibinfo{year}{2013}\natexlab{}.
\newblock \showarticletitle{Employee Resilience Guidebook for Handling Child
  Sex Abuse Images}.
\newblock  (\bibinfo{year}{2013}).
\newblock
\urldef\tempurl%
\url{https://www.thorn.org/wp-content/uploads/2015/02/EmployeeResilienceGuidebookFinal7-13-1.pdf}
\showURL{%
\tempurl}


\bibitem[\protect\citeauthoryear{Cormack and Grossman}{Cormack and
  Grossman}{2015}]%
        {cormack_autonomy_2015}
\bibfield{author}{\bibinfo{person}{Gordon~V. Cormack} {and}
  \bibinfo{person}{Maura~R. Grossman}.} \bibinfo{year}{2015}\natexlab{}.
\newblock \showarticletitle{Autonomy and {Reliability} of {Continuous} {Active}
  {Learning} for {Technology}-{Assisted} {Review}}.
\newblock \bibinfo{journal}{\emph{arXiv:1504.06868 [cs]}}
  (\bibinfo{date}{April} \bibinfo{year}{2015}).
\newblock
\newblock
\shownote{arXiv: 1504.06868.}


\bibitem[\protect\citeauthoryear{Davidson, Bhattacharya, and Weber}{Davidson
  et~al\mbox{.}}{2019}]%
        {davidson2019racial}
\bibfield{author}{\bibinfo{person}{Thomas Davidson}, \bibinfo{person}{Debasmita
  Bhattacharya}, {and} \bibinfo{person}{Ingmar Weber}.}
  \bibinfo{year}{2019}\natexlab{}.
\newblock \showarticletitle{Racial bias in hate speech and abusive language
  detection datasets}.
\newblock \bibinfo{journal}{\emph{arXiv preprint arXiv:1905.12516}}
  (\bibinfo{year}{2019}).
\newblock


\bibitem[\protect\citeauthoryear{Davidson, Warmsley, Macy, and Weber}{Davidson
  et~al\mbox{.}}{2017}]%
        {davidson2017automated}
\bibfield{author}{\bibinfo{person}{Thomas Davidson}, \bibinfo{person}{Dana
  Warmsley}, \bibinfo{person}{Michael Macy}, {and} \bibinfo{person}{Ingmar
  Weber}.} \bibinfo{year}{2017}\natexlab{}.
\newblock \showarticletitle{Automated hate speech detection and the problem of
  offensive language}. In \bibinfo{booktitle}{\emph{Proceedings of the
  International AAAI Conference on Web and Social Media}},
  Vol.~\bibinfo{volume}{11}.
\newblock


\bibitem[\protect\citeauthoryear{de~Gibert, Perez, Garc{\'\i}a-Pablos, and
  Cuadros}{de~Gibert et~al\mbox{.}}{2018}]%
        {de-gibert-etal-2018-hate}
\bibfield{author}{\bibinfo{person}{Ona de Gibert}, \bibinfo{person}{Naiara
  Perez}, \bibinfo{person}{Aitor Garc{\'\i}a-Pablos}, {and}
  \bibinfo{person}{Montse Cuadros}.} \bibinfo{year}{2018}\natexlab{}.
\newblock \showarticletitle{Hate Speech Dataset from a White Supremacy Forum}.
  In \bibinfo{booktitle}{\emph{Proceedings of the 2nd Workshop on Abusive
  Language Online ({ALW}2)}}. \bibinfo{publisher}{Association for Computational
  Linguistics}, \bibinfo{address}{Brussels, Belgium}, \bibinfo{pages}{11--20}.
\newblock
\urldef\tempurl%
\url{https://doi.org/10.18653/v1/W18-5102}
\showDOI{\tempurl}


\bibitem[\protect\citeauthoryear{Del~Vigna, Cimino, Dell’Orletta, Petrocchi,
  and Tesconi}{Del~Vigna et~al\mbox{.}}{2017}]%
        {del2017hate}
\bibfield{author}{\bibinfo{person}{Fabio Del~Vigna}, \bibinfo{person}{Andrea
  Cimino}, \bibinfo{person}{Felice Dell’Orletta}, \bibinfo{person}{Marinella
  Petrocchi}, {and} \bibinfo{person}{Maurizio Tesconi}.}
  \bibinfo{year}{2017}\natexlab{}.
\newblock \showarticletitle{Hate me, hate me not: Hate speech detection on
  facebook}. In \bibinfo{booktitle}{\emph{Proceedings of the First Italian
  Conference on Cybersecurity (ITASEC17)}}. \bibinfo{pages}{86--95}.
\newblock


\bibitem[\protect\citeauthoryear{Derczynski}{Derczynski}{2021}]%
        {hatespeechdata}
\bibfield{author}{\bibinfo{person}{Leon Derczynski}.}
  \bibinfo{year}{2021}\natexlab{}.
\newblock \bibinfo{booktitle}{\emph{Hate speech data. ~\textnormal{\small
  \url{https://hatespeechdata.com/}}}}.
\newblock


\bibitem[\protect\citeauthoryear{Devlin, Chang, Lee, and Toutanova}{Devlin
  et~al\mbox{.}}{2018}]%
        {devlin2018bert}
\bibfield{author}{\bibinfo{person}{Jacob Devlin}, \bibinfo{person}{Ming-Wei
  Chang}, \bibinfo{person}{Kenton Lee}, {and} \bibinfo{person}{Kristina
  Toutanova}.} \bibinfo{year}{2018}\natexlab{}.
\newblock \showarticletitle{Bert: Pre-training of deep bidirectional
  transformers for language understanding}.
\newblock \bibinfo{journal}{\emph{arXiv preprint arXiv:1810.04805}}
  (\bibinfo{year}{2018}).
\newblock


\bibitem[\protect\citeauthoryear{DeYoung, Jain, Rajani, Lehman, Xiong, Socher,
  and Wallace}{DeYoung et~al\mbox{.}}{2019}]%
        {deyoung2019eraser}
\bibfield{author}{\bibinfo{person}{Jay DeYoung}, \bibinfo{person}{Sarthak
  Jain}, \bibinfo{person}{Nazneen~Fatema Rajani}, \bibinfo{person}{Eric
  Lehman}, \bibinfo{person}{Caiming Xiong}, \bibinfo{person}{Richard Socher},
  {and} \bibinfo{person}{Byron~C Wallace}.} \bibinfo{year}{2019}\natexlab{}.
\newblock \showarticletitle{Eraser: A benchmark to evaluate rationalized nlp
  models}.
\newblock \bibinfo{journal}{\emph{arXiv preprint arXiv:1911.03429}}
  (\bibinfo{year}{2019}).
\newblock


\bibitem[\protect\citeauthoryear{Dixon, Li, Sorensen, Thain, and
  Vasserman}{Dixon et~al\mbox{.}}{2018}]%
        {dixon2018measuring}
\bibfield{author}{\bibinfo{person}{Lucas Dixon}, \bibinfo{person}{John Li},
  \bibinfo{person}{Jeffrey Sorensen}, \bibinfo{person}{Nithum Thain}, {and}
  \bibinfo{person}{Lucy Vasserman}.} \bibinfo{year}{2018}\natexlab{}.
\newblock \showarticletitle{Measuring and mitigating unintended bias in text
  classification}. In \bibinfo{booktitle}{\emph{Proceedings of the 2018
  AAAI/ACM Conference on AI, Ethics, and Society}}. \bibinfo{pages}{67--73}.
\newblock


\bibitem[\protect\citeauthoryear{Djuric, Zhou, Morris, Grbovic, Radosavljevic,
  and Bhamidipati}{Djuric et~al\mbox{.}}{2015}]%
        {djuric2015hate}
\bibfield{author}{\bibinfo{person}{Nemanja Djuric}, \bibinfo{person}{Jing
  Zhou}, \bibinfo{person}{Robin Morris}, \bibinfo{person}{Mihajlo Grbovic},
  \bibinfo{person}{Vladan Radosavljevic}, {and} \bibinfo{person}{Narayan
  Bhamidipati}.} \bibinfo{year}{2015}\natexlab{}.
\newblock \showarticletitle{Hate speech detection with comment embeddings}. In
  \bibinfo{booktitle}{\emph{Proceedings of the 24th international conference on
  world wide web}}. \bibinfo{pages}{29--30}.
\newblock


\bibitem[\protect\citeauthoryear{Eubanks}{Eubanks}{2018}]%
        {eubanksautomating}
\bibfield{author}{\bibinfo{person}{Virginia Eubanks}.}
  \bibinfo{year}{2018}\natexlab{}.
\newblock \bibinfo{title}{Automating Inequality: How High-Tech Tools Profile,
  Police, and Punish the Poor}.
\newblock
\newblock


\bibitem[\protect\citeauthoryear{Eugenio and Glass}{Eugenio and Glass}{2004}]%
        {eugenio2004kappa}
\bibfield{author}{\bibinfo{person}{Barbara~Di Eugenio} {and}
  \bibinfo{person}{Michael Glass}.} \bibinfo{year}{2004}\natexlab{}.
\newblock \showarticletitle{The kappa statistic: A second look}.
\newblock \bibinfo{journal}{\emph{Computational linguistics}}
  \bibinfo{volume}{30}, \bibinfo{number}{1} (\bibinfo{year}{2004}),
  \bibinfo{pages}{95--101}.
\newblock


\bibitem[\protect\citeauthoryear{Feinstein and Cicchetti}{Feinstein and
  Cicchetti}{1990}]%
        {feinstein1990high}
\bibfield{author}{\bibinfo{person}{Alvan~R Feinstein} {and}
  \bibinfo{person}{Domenic~V Cicchetti}.} \bibinfo{year}{1990}\natexlab{}.
\newblock \showarticletitle{High agreement but low kappa: I. The problems of
  two paradoxes}.
\newblock \bibinfo{journal}{\emph{Journal of clinical epidemiology}}
  \bibinfo{volume}{43}, \bibinfo{number}{6} (\bibinfo{year}{1990}),
  \bibinfo{pages}{543--549}.
\newblock


\bibitem[\protect\citeauthoryear{Fortuna and Nunes}{Fortuna and Nunes}{2018}]%
        {fortuna2018survey}
\bibfield{author}{\bibinfo{person}{Paula Fortuna} {and}
  \bibinfo{person}{S{\'e}rgio Nunes}.} \bibinfo{year}{2018}\natexlab{}.
\newblock \showarticletitle{A survey on automatic detection of hate speech in
  text}.
\newblock \bibinfo{journal}{\emph{ACM Computing Surveys (CSUR)}}
  \bibinfo{volume}{51}, \bibinfo{number}{4} (\bibinfo{year}{2018}),
  \bibinfo{pages}{1--30}.
\newblock


\bibitem[\protect\citeauthoryear{Founta, Djouvas, Chatzakou, Leontiadis,
  Blackburn, Stringhini, Vakali, Sirivianos, and Kourtellis}{Founta
  et~al\mbox{.}}{2018}]%
        {founta2018large}
\bibfield{author}{\bibinfo{person}{Antigoni Founta},
  \bibinfo{person}{Constantinos Djouvas}, \bibinfo{person}{Despoina Chatzakou},
  \bibinfo{person}{Ilias Leontiadis}, \bibinfo{person}{Jeremy Blackburn},
  \bibinfo{person}{Gianluca Stringhini}, \bibinfo{person}{Athena Vakali},
  \bibinfo{person}{Michael Sirivianos}, {and} \bibinfo{person}{Nicolas
  Kourtellis}.} \bibinfo{year}{2018}\natexlab{}.
\newblock \showarticletitle{Large scale crowdsourcing and characterization of
  twitter abusive behavior}. In \bibinfo{booktitle}{\emph{Proceedings of the
  International AAAI Conference on Web and Social Media}},
  Vol.~\bibinfo{volume}{12}.
\newblock


\bibitem[\protect\citeauthoryear{Gao and Huang}{Gao and Huang}{2017}]%
        {gao2017detecting}
\bibfield{author}{\bibinfo{person}{Lei Gao} {and} \bibinfo{person}{Ruihong
  Huang}.} \bibinfo{year}{2017}\natexlab{}.
\newblock \showarticletitle{Detecting online hate speech using context aware
  models}.
\newblock \bibinfo{journal}{\emph{arXiv preprint arXiv:1710.07395}}
  (\bibinfo{year}{2017}).
\newblock


\bibitem[\protect\citeauthoryear{Gebru, Morgenstern, Vecchione, Vaughan,
  Wallach, Daum{\'e}~III, and Crawford}{Gebru et~al\mbox{.}}{2018}]%
        {gebru2018datasheets}
\bibfield{author}{\bibinfo{person}{Timnit Gebru}, \bibinfo{person}{Jamie
  Morgenstern}, \bibinfo{person}{Briana Vecchione},
  \bibinfo{person}{Jennifer~Wortman Vaughan}, \bibinfo{person}{Hanna Wallach},
  \bibinfo{person}{Hal Daum{\'e}~III}, {and} \bibinfo{person}{Kate Crawford}.}
  \bibinfo{year}{2018}\natexlab{}.
\newblock \showarticletitle{Datasheets for datasets}.
\newblock \bibinfo{journal}{\emph{arXiv preprint arXiv:1803.09010}}
  (\bibinfo{year}{2018}).
\newblock


\bibitem[\protect\citeauthoryear{Geva, Goldberg, and Berant}{Geva
  et~al\mbox{.}}{2019}]%
        {geva-etal-2019-modeling}
\bibfield{author}{\bibinfo{person}{Mor Geva}, \bibinfo{person}{Yoav Goldberg},
  {and} \bibinfo{person}{Jonathan Berant}.} \bibinfo{year}{2019}\natexlab{}.
\newblock \showarticletitle{Are We Modeling the Task or the Annotator? An
  Investigation of Annotator Bias in Natural Language Understanding Datasets}.
  In \bibinfo{booktitle}{\emph{Proceedings of the 2019 Conference on Empirical
  Methods in Natural Language Processing and the 9th International Joint
  Conference on Natural Language Processing (EMNLP-IJCNLP)}}.
  \bibinfo{publisher}{Association for Computational Linguistics},
  \bibinfo{address}{Hong Kong, China}, \bibinfo{pages}{1161--1166}.
\newblock
\urldef\tempurl%
\url{https://doi.org/10.18653/v1/D19-1107}
\showDOI{\tempurl}


\bibitem[\protect\citeauthoryear{Golbeck, Ashktorab, Banjo, Berlinger, Bhagwan,
  Buntain, Cheakalos, Geller, Gnanasekaran, Gunasekaran, et~al\mbox{.}}{Golbeck
  et~al\mbox{.}}{2017}]%
        {golbeck2017large}
\bibfield{author}{\bibinfo{person}{Jennifer Golbeck}, \bibinfo{person}{Zahra
  Ashktorab}, \bibinfo{person}{Rashad~O Banjo}, \bibinfo{person}{Alexandra
  Berlinger}, \bibinfo{person}{Siddharth Bhagwan}, \bibinfo{person}{Cody
  Buntain}, \bibinfo{person}{Paul Cheakalos}, \bibinfo{person}{Alicia~A
  Geller}, \bibinfo{person}{Rajesh~Kumar Gnanasekaran},
  \bibinfo{person}{Raja~Rajan Gunasekaran}, {et~al\mbox{.}}}
  \bibinfo{year}{2017}\natexlab{}.
\newblock \showarticletitle{A large labeled corpus for online harassment
  research}. In \bibinfo{booktitle}{\emph{Proceedings of the 2017 ACM on web
  science conference}}. \bibinfo{pages}{229--233}.
\newblock


\bibitem[\protect\citeauthoryear{Gold and Zesch}{Gold and Zesch}{2018}]%
        {gold2018women}
\bibfield{author}{\bibinfo{person}{Michael Wojatzki Tobias Horsmann~Darina
  Gold} {and} \bibinfo{person}{Torsten Zesch}.}
  \bibinfo{year}{2018}\natexlab{}.
\newblock \showarticletitle{Do women perceive hate differently: Examining the
  relationship between hate speech, gender, and agreement judgments}.
\newblock  (\bibinfo{year}{2018}).
\newblock


\bibitem[\protect\citeauthoryear{Grimminger and Klinger}{Grimminger and
  Klinger}{2021}]%
        {grimminger2021hate}
\bibfield{author}{\bibinfo{person}{Lara Grimminger} {and}
  \bibinfo{person}{Roman Klinger}.} \bibinfo{year}{2021}\natexlab{}.
\newblock \showarticletitle{Hate Towards the Political Opponent: A Twitter
  Corpus Study of the 2020 US Elections on the Basis of Offensive Speech and
  Stance Detection}.
\newblock \bibinfo{journal}{\emph{arXiv preprint arXiv:2103.01664}}
  (\bibinfo{year}{2021}).
\newblock


\bibitem[\protect\citeauthoryear{Gr{\"o}ndahl, Pajola, Juuti, Conti, and
  Asokan}{Gr{\"o}ndahl et~al\mbox{.}}{2018}]%
        {grondahl2018all}
\bibfield{author}{\bibinfo{person}{Tommi Gr{\"o}ndahl}, \bibinfo{person}{Luca
  Pajola}, \bibinfo{person}{Mika Juuti}, \bibinfo{person}{Mauro Conti}, {and}
  \bibinfo{person}{N Asokan}.} \bibinfo{year}{2018}\natexlab{}.
\newblock \showarticletitle{All you need is" love" evading hate speech
  detection}. In \bibinfo{booktitle}{\emph{Proceedings of the 11th ACM workshop
  on artificial intelligence and security}}. \bibinfo{pages}{2--12}.
\newblock


\bibitem[\protect\citeauthoryear{Gunther, Awasthi, Axelrod, Miehling, Wagh, and
  Joeng}{Gunther et~al\mbox{.}}{2021}]%
        {Jikeli2021Anti}
\bibfield{author}{\bibinfo{person}{Jikeli Gunther}, \bibinfo{person}{Deepika
  Awasthi}, \bibinfo{person}{David Axelrod}, \bibinfo{person}{Daniel Miehling},
  \bibinfo{person}{Pauravi Wagh}, {and} \bibinfo{person}{Weejoeng Joeng}.}
  \bibinfo{year}{2021}\natexlab{}.
\newblock \showarticletitle{Detecting Anti-Jewish Messages on Social Media.
  Building an Annotated Corpus That Can Serve as A Preliminary Gold Standard}.
\newblock \bibinfo{journal}{\emph{Workshop Proceedings of the 15th
  International AAAI Conference on Web and Social Media}}
  (\bibinfo{year}{2021}).
\newblock


\bibitem[\protect\citeauthoryear{Gwet}{Gwet}{2002}]%
        {gwet2002kappa}
\bibfield{author}{\bibinfo{person}{Kilem Gwet}.}
  \bibinfo{year}{2002}\natexlab{}.
\newblock \showarticletitle{Kappa statistic is not satisfactory for assessing
  the extent of agreement between raters}.
\newblock \bibinfo{journal}{\emph{Statistical methods for inter-rater
  reliability assessment}} \bibinfo{volume}{1}, \bibinfo{number}{6}
  (\bibinfo{year}{2002}), \bibinfo{pages}{1--6}.
\newblock


\bibitem[\protect\citeauthoryear{Halevy, Ferrer, Ma, Ozertem, Pantel, Saeidi,
  Silvestri, and Stoyanov}{Halevy et~al\mbox{.}}{2020}]%
        {halevy2020preserving}
\bibfield{author}{\bibinfo{person}{Alon Halevy},
  \bibinfo{person}{Cristian~Canton Ferrer}, \bibinfo{person}{Hao Ma},
  \bibinfo{person}{Umut Ozertem}, \bibinfo{person}{Patrick Pantel},
  \bibinfo{person}{Marzieh Saeidi}, \bibinfo{person}{Fabrizio Silvestri}, {and}
  \bibinfo{person}{Ves Stoyanov}.} \bibinfo{year}{2020}\natexlab{}.
\newblock \showarticletitle{Preserving integrity in online social networks}.
\newblock \bibinfo{journal}{\emph{arXiv preprint arXiv:2009.10311}}
  (\bibinfo{year}{2020}).
\newblock


\bibitem[\protect\citeauthoryear{Hatebase}{Hatebase}{[n.d.]}]%
        {hatebase}
\bibfield{author}{\bibinfo{person}{Hatebase}.}
  \bibinfo{year}{[n.d.]}\natexlab{}.
\newblock \bibinfo{title}{The world's largest structured repository of
  regionalized, multilingual hate speech}.
\newblock \bibinfo{howpublished}{\url{https://hatebase.org/}}.
\newblock


\bibitem[\protect\citeauthoryear{Hitlin}{Hitlin}{2016}]%
        {hitlin2016research}
\bibfield{author}{\bibinfo{person}{Paul Hitlin}.}
  \bibinfo{year}{2016}\natexlab{}.
\newblock \showarticletitle{Research in the crowdsourcing age: A case study}.
\newblock  (\bibinfo{year}{2016}).
\newblock


\bibitem[\protect\citeauthoryear{Hosseinmardi, Mattson, Rafiq, Han, Lv, and
  Mishra}{Hosseinmardi et~al\mbox{.}}{2015}]%
        {hosseinmardi2015detection}
\bibfield{author}{\bibinfo{person}{Homa Hosseinmardi},
  \bibinfo{person}{Sabrina~Arredondo Mattson}, \bibinfo{person}{Rahat~Ibn
  Rafiq}, \bibinfo{person}{Richard Han}, \bibinfo{person}{Qin Lv}, {and}
  \bibinfo{person}{Shivakant Mishra}.} \bibinfo{year}{2015}\natexlab{}.
\newblock \showarticletitle{Detection of cyberbullying incidents on the
  instagram social network}.
\newblock \bibinfo{journal}{\emph{arXiv preprint arXiv:1503.03909}}
  (\bibinfo{year}{2015}).
\newblock


\bibitem[\protect\citeauthoryear{Jiang and Matsubara}{Jiang and
  Matsubara}{2014}]%
        {JiangEfficient2014}
\bibfield{author}{\bibinfo{person}{Huan Jiang} {and} \bibinfo{person}{Shigeo
  Matsubara}.} \bibinfo{year}{2014}\natexlab{}.
\newblock \showarticletitle{Efficient Task Decomposition in Crowdsourcing}. In
  \bibinfo{booktitle}{\emph{PRIMA 2014: Principles and Practice of Multi-Agent
  Systems}}, \bibfield{editor}{\bibinfo{person}{Hoa~Khanh Dam},
  \bibinfo{person}{Jeremy Pitt}, \bibinfo{person}{Yang Xu},
  \bibinfo{person}{Guido Governatori}, {and} \bibinfo{person}{Takayuki Ito}}
  (Eds.). \bibinfo{publisher}{Springer International Publishing},
  \bibinfo{address}{Cham}, \bibinfo{pages}{65--73}.
\newblock
\showISBNx{978-3-319-13191-7}


\bibitem[\protect\citeauthoryear{Jurgens, Chandrasekharan, and
  Hemphill}{Jurgens et~al\mbox{.}}{2019}]%
        {jurgens2019just}
\bibfield{author}{\bibinfo{person}{David Jurgens}, \bibinfo{person}{Eshwar
  Chandrasekharan}, {and} \bibinfo{person}{Libby Hemphill}.}
  \bibinfo{year}{2019}\natexlab{}.
\newblock \showarticletitle{A Just and Comprehensive Strategy for Using NLP to
  Address Online Abuse}.
\newblock \bibinfo{journal}{\emph{arXiv preprint arXiv:1906.01738}}
  (\bibinfo{year}{2019}).
\newblock


\bibitem[\protect\citeauthoryear{Kennedy, Atari, Davani, Yeh, Omrani, Kim,
  Coombs, Havaldar, Portillo-Wightman, Gonzalez, et~al\mbox{.}}{Kennedy
  et~al\mbox{.}}{2018}]%
        {kennedy2018gab}
\bibfield{author}{\bibinfo{person}{Brendan Kennedy}, \bibinfo{person}{Mohammad
  Atari}, \bibinfo{person}{Aida~Mostafazadeh Davani}, \bibinfo{person}{Leigh
  Yeh}, \bibinfo{person}{Ali Omrani}, \bibinfo{person}{Yehsong Kim},
  \bibinfo{person}{Kris Coombs}, \bibinfo{person}{Shreya Havaldar},
  \bibinfo{person}{Gwenyth Portillo-Wightman}, \bibinfo{person}{Elaine
  Gonzalez}, {et~al\mbox{.}}} \bibinfo{year}{2018}\natexlab{}.
\newblock \showarticletitle{The Gab Hate Corpus: A collection of 27k posts
  annotated for hate speech}.
\newblock  (\bibinfo{year}{2018}).
\newblock


\bibitem[\protect\citeauthoryear{Kleim and Westphal}{Kleim and
  Westphal}{2011}]%
        {kleim2011mental}
\bibfield{author}{\bibinfo{person}{Birgit Kleim} {and} \bibinfo{person}{Maren
  Westphal}.} \bibinfo{year}{2011}\natexlab{}.
\newblock \showarticletitle{Mental health in first responders: A review and
  recommendation for prevention and intervention strategies}.
\newblock \bibinfo{journal}{\emph{Traumatology}} \bibinfo{volume}{17},
  \bibinfo{number}{4} (\bibinfo{year}{2011}), \bibinfo{pages}{17--24}.
\newblock


\bibitem[\protect\citeauthoryear{Kumar, Ojha, Malmasi, and Zampieri}{Kumar
  et~al\mbox{.}}{2018a}]%
        {kumar-etal-2018-benchmarking}
\bibfield{author}{\bibinfo{person}{Ritesh Kumar}, \bibinfo{person}{Atul~Kr.
  Ojha}, \bibinfo{person}{Shervin Malmasi}, {and} \bibinfo{person}{Marcos
  Zampieri}.} \bibinfo{year}{2018}\natexlab{a}.
\newblock \showarticletitle{Benchmarking Aggression Identification in Social
  Media}. In \bibinfo{booktitle}{\emph{Proceedings of the First Workshop on
  Trolling, Aggression and Cyberbullying ({TRAC}-2018)}}.
  \bibinfo{publisher}{Association for Computational Linguistics},
  \bibinfo{address}{Santa Fe, New Mexico, USA}, \bibinfo{pages}{1--11}.
\newblock
\urldef\tempurl%
\url{https://www.aclweb.org/anthology/W18-4401}
\showURL{%
\tempurl}


\bibitem[\protect\citeauthoryear{Kumar, Reganti, Bhatia, and Maheshwari}{Kumar
  et~al\mbox{.}}{2018b}]%
        {kumar2018aggression}
\bibfield{author}{\bibinfo{person}{Ritesh Kumar}, \bibinfo{person}{Aishwarya~N
  Reganti}, \bibinfo{person}{Akshit Bhatia}, {and} \bibinfo{person}{Tushar
  Maheshwari}.} \bibinfo{year}{2018}\natexlab{b}.
\newblock \showarticletitle{Aggression-annotated corpus of hindi-english
  code-mixed data}.
\newblock \bibinfo{journal}{\emph{arXiv preprint arXiv:1803.09402}}
  (\bibinfo{year}{2018}).
\newblock


\bibitem[\protect\citeauthoryear{Kutlu, McDonnell, Barkallah, Elsayed, and
  Lease}{Kutlu et~al\mbox{.}}{2018}]%
        {Kutlu18-sigir}
\bibfield{author}{\bibinfo{person}{Mucahid Kutlu}, \bibinfo{person}{Tyler
  McDonnell}, \bibinfo{person}{Yassmine Barkallah}, \bibinfo{person}{Tamer
  Elsayed}, {and} \bibinfo{person}{Matthew Lease}.}
  \bibinfo{year}{2018}\natexlab{}.
\newblock \showarticletitle{{What Can Rationales behind Relevance Judgments
  Tell Us About Assessor Disagreement?}}. In
  \bibinfo{booktitle}{\emph{Proceedings of the 41st international ACM SIGIR
  conference on Research and development in Information Retrieval}}.
  \bibinfo{pages}{805--814}.
\newblock


\bibitem[\protect\citeauthoryear{Kwok and Wang}{Kwok and Wang}{2013}]%
        {kwok2013locate}
\bibfield{author}{\bibinfo{person}{Irene Kwok} {and} \bibinfo{person}{Yuzhou
  Wang}.} \bibinfo{year}{2013}\natexlab{}.
\newblock \showarticletitle{Locate the hate: Detecting tweets against blacks}.
  In \bibinfo{booktitle}{\emph{Proceedings of the AAAI Conference on Artificial
  Intelligence}}, Vol.~\bibinfo{volume}{27}.
\newblock


\bibitem[\protect\citeauthoryear{Laaksonen, Haapoja, Kinnunen, Nelimarkka, and
  P{\"o}yht{\"a}ri}{Laaksonen et~al\mbox{.}}{2020}]%
        {laaksonen2020datafication}
\bibfield{author}{\bibinfo{person}{Salla-Maaria Laaksonen},
  \bibinfo{person}{Jesse Haapoja}, \bibinfo{person}{Teemu Kinnunen},
  \bibinfo{person}{Matti Nelimarkka}, {and} \bibinfo{person}{Reeta
  P{\"o}yht{\"a}ri}.} \bibinfo{year}{2020}\natexlab{}.
\newblock \showarticletitle{The datafication of hate: expectations and
  challenges in automated hate speech monitoring}.
\newblock \bibinfo{journal}{\emph{Frontiers in big Data}}  \bibinfo{volume}{3}
  (\bibinfo{year}{2020}), \bibinfo{pages}{3}.
\newblock


\bibitem[\protect\citeauthoryear{Lease, Cormack, Nguyen, Trikalinos, and
  Wallace}{Lease et~al\mbox{.}}{2016}]%
        {Lease16-medir}
\bibfield{author}{\bibinfo{person}{Matthew Lease}, \bibinfo{person}{Gordon~V
  Cormack}, \bibinfo{person}{An~Thanh Nguyen}, \bibinfo{person}{Thomas~A
  Trikalinos}, {and} \bibinfo{person}{Byron~C Wallace}.}
  \bibinfo{year}{2016}\natexlab{}.
\newblock \showarticletitle{{Systematic Review is e-Discovery in Doctor's
  Clothing}}. In \bibinfo{booktitle}{\emph{Proceedings of the Medical
  Information Retrieval (MedIR) Workshop at the 39th International ACM SIGIR
  Conference on Research and Development in Information Retrieval}}.
\newblock


\bibitem[\protect\citeauthoryear{Leets}{Leets}{2002}]%
        {leets2002experiencing}
\bibfield{author}{\bibinfo{person}{Laura Leets}.}
  \bibinfo{year}{2002}\natexlab{}.
\newblock \showarticletitle{Experiencing hate speech: Perceptions and responses
  to anti-semitism and antigay speech}.
\newblock \bibinfo{journal}{\emph{Journal of social issues}}
  \bibinfo{volume}{58}, \bibinfo{number}{2} (\bibinfo{year}{2002}),
  \bibinfo{pages}{341--361}.
\newblock


\bibitem[\protect\citeauthoryear{Lehman, DeYoung, Barzilay, and Wallace}{Lehman
  et~al\mbox{.}}{2019}]%
        {lehman2019inferring}
\bibfield{author}{\bibinfo{person}{Eric Lehman}, \bibinfo{person}{Jay DeYoung},
  \bibinfo{person}{Regina Barzilay}, {and} \bibinfo{person}{Byron~C Wallace}.}
  \bibinfo{year}{2019}\natexlab{}.
\newblock \showarticletitle{Inferring Which Medical Treatments Work from
  Reports of Clinical Trials}. In \bibinfo{booktitle}{\emph{Proceedings of the
  2019 Conference of the North American Chapter of the Association for
  Computational Linguistics: Human Language Technologies, Volume 1 (Long and
  Short Papers)}}. \bibinfo{pages}{3705--3717}.
\newblock


\bibitem[\protect\citeauthoryear{Levin}{Levin}{2017}]%
        {microsoft2017}
\bibfield{author}{\bibinfo{person}{Sam Levin}.}
  \bibinfo{year}{2017}\natexlab{}.
\newblock \showarticletitle{Moderators who had to view child abuse content sue
  Microsoft, claiming PTSD}.
\newblock  (\bibinfo{year}{2017}).
\newblock
\urldef\tempurl%
\url{https://www.theguardian.com/technology/2017/jan/11/microsoft-employees-child-abuse-lawsuit-ptsd}
\showURL{%
\tempurl}


\bibitem[\protect\citeauthoryear{Lewis and Gale}{Lewis and Gale}{1994}]%
        {lewis1994sequential}
\bibfield{author}{\bibinfo{person}{David~D Lewis} {and}
  \bibinfo{person}{William~A Gale}.} \bibinfo{year}{1994}\natexlab{}.
\newblock \showarticletitle{A sequential algorithm for training text
  classifiers}. In \bibinfo{booktitle}{\emph{Proceedings of the 17th annual
  international ACM SIGIR conference on Research and development in information
  retrieval}}. Springer-Verlag New York, Inc., \bibinfo{pages}{3--12}.
\newblock


\bibitem[\protect\citeauthoryear{Ludick and Figley}{Ludick and Figley}{2017}]%
        {ludick2017toward}
\bibfield{author}{\bibinfo{person}{Marn{\'e} Ludick} {and}
  \bibinfo{person}{Charles~R Figley}.} \bibinfo{year}{2017}\natexlab{}.
\newblock \showarticletitle{Toward a mechanism for secondary trauma induction
  and reduction: Reimagining a theory of secondary traumatic stress.}
\newblock \bibinfo{journal}{\emph{Traumatology}} \bibinfo{volume}{23},
  \bibinfo{number}{1} (\bibinfo{year}{2017}), \bibinfo{pages}{112}.
\newblock


\bibitem[\protect\citeauthoryear{MacAvaney, Yao, Yang, Russell, Goharian, and
  Frieder}{MacAvaney et~al\mbox{.}}{2019}]%
        {macavaney2019hate}
\bibfield{author}{\bibinfo{person}{Sean MacAvaney}, \bibinfo{person}{Hao-Ren
  Yao}, \bibinfo{person}{Eugene Yang}, \bibinfo{person}{Katina Russell},
  \bibinfo{person}{Nazli Goharian}, {and} \bibinfo{person}{Ophir Frieder}.}
  \bibinfo{year}{2019}\natexlab{}.
\newblock \showarticletitle{Hate speech detection: Challenges and solutions}.
\newblock \bibinfo{journal}{\emph{PloS one}} \bibinfo{volume}{14},
  \bibinfo{number}{8} (\bibinfo{year}{2019}), \bibinfo{pages}{e0221152}.
\newblock


\bibitem[\protect\citeauthoryear{Madukwe, Gao, and Xue}{Madukwe
  et~al\mbox{.}}{2020}]%
        {madukwe2020data}
\bibfield{author}{\bibinfo{person}{Kosisochukwu Madukwe},
  \bibinfo{person}{Xiaoying Gao}, {and} \bibinfo{person}{Bing Xue}.}
  \bibinfo{year}{2020}\natexlab{}.
\newblock \showarticletitle{In data we trust: A critical analysis of hate
  speech detection datasets}. In \bibinfo{booktitle}{\emph{Proceedings of the
  Fourth Workshop on Online Abuse and Harms}}. \bibinfo{pages}{150--161}.
\newblock


\bibitem[\protect\citeauthoryear{Mathew, Saha, Yimam, Biemann, Goyal, and
  Mukherjee}{Mathew et~al\mbox{.}}{2020}]%
        {mathew2020hatexplain}
\bibfield{author}{\bibinfo{person}{Binny Mathew}, \bibinfo{person}{Punyajoy
  Saha}, \bibinfo{person}{Seid~Muhie Yimam}, \bibinfo{person}{Chris Biemann},
  \bibinfo{person}{Pawan Goyal}, {and} \bibinfo{person}{Animesh Mukherjee}.}
  \bibinfo{year}{2020}\natexlab{}.
\newblock \showarticletitle{HateXplain: A Benchmark Dataset for Explainable
  Hate Speech Detection}.
\newblock \bibinfo{journal}{\emph{arXiv preprint arXiv:2012.10289}}
  (\bibinfo{year}{2020}).
\newblock


\bibitem[\protect\citeauthoryear{Mathur, Sawhney, Ayyar, and Shah}{Mathur
  et~al\mbox{.}}{2018}]%
        {mathur2018did}
\bibfield{author}{\bibinfo{person}{Puneet Mathur}, \bibinfo{person}{Ramit
  Sawhney}, \bibinfo{person}{Meghna Ayyar}, {and} \bibinfo{person}{Rajiv
  Shah}.} \bibinfo{year}{2018}\natexlab{}.
\newblock \showarticletitle{Did you offend me? classification of offensive
  tweets in hinglish language}. In \bibinfo{booktitle}{\emph{Proceedings of the
  2nd Workshop on Abusive Language Online (ALW2)}}. \bibinfo{pages}{138--148}.
\newblock


\bibitem[\protect\citeauthoryear{May and Wisco}{May and Wisco}{2016}]%
        {may2016defining}
\bibfield{author}{\bibinfo{person}{Casey~L May} {and} \bibinfo{person}{Blair~E
  Wisco}.} \bibinfo{year}{2016}\natexlab{}.
\newblock \showarticletitle{Defining trauma: How level of exposure and
  proximity affect risk for posttraumatic stress disorder.}
\newblock \bibinfo{journal}{\emph{Psychological trauma: theory, research,
  practice, and policy}} \bibinfo{volume}{8}, \bibinfo{number}{2}
  (\bibinfo{year}{2016}), \bibinfo{pages}{233}.
\newblock


\bibitem[\protect\citeauthoryear{McDonnell, Kutlu, Elsayed, and
  Lease}{McDonnell et~al\mbox{.}}{2017}]%
        {mcdonnell16-hcomp}
\bibfield{author}{\bibinfo{person}{Tyler McDonnell}, \bibinfo{person}{Mucahid
  Kutlu}, \bibinfo{person}{Tamer Elsayed}, {and} \bibinfo{person}{Matthew
  Lease}.} \bibinfo{year}{2017}\natexlab{}.
\newblock \showarticletitle{The many benefits of annotator rationales for
  relevance judgments}. In \bibinfo{booktitle}{\emph{Proceedings of the 26th
  International Joint Conference on Artificial Intelligence}}.
  \bibinfo{pages}{4909--4913}.
\newblock


\bibitem[\protect\citeauthoryear{Mehrabi, Morstatter, Saxena, Lerman, and
  Galstyan}{Mehrabi et~al\mbox{.}}{2021}]%
        {mehrabi2021survey}
\bibfield{author}{\bibinfo{person}{Ninareh Mehrabi}, \bibinfo{person}{Fred
  Morstatter}, \bibinfo{person}{Nripsuta Saxena}, \bibinfo{person}{Kristina
  Lerman}, {and} \bibinfo{person}{Aram Galstyan}.}
  \bibinfo{year}{2021}\natexlab{}.
\newblock \showarticletitle{A survey on bias and fairness in machine learning}.
\newblock \bibinfo{journal}{\emph{ACM Computing Surveys (CSUR)}}
  \bibinfo{volume}{54}, \bibinfo{number}{6} (\bibinfo{year}{2021}),
  \bibinfo{pages}{1--35}.
\newblock


\bibitem[\protect\citeauthoryear{Nguyen, Halpern, Wallace, and Lease}{Nguyen
  et~al\mbox{.}}{2016a}]%
        {Nguyen16-hcomp}
\bibfield{author}{\bibinfo{person}{An~Thanh Nguyen}, \bibinfo{person}{Matthew
  Halpern}, \bibinfo{person}{Byron~C.\ Wallace}, {and} \bibinfo{person}{Matthew
  Lease}.} \bibinfo{year}{2016}\natexlab{a}.
\newblock \showarticletitle{{Probabilistic Modeling for Crowdsourcing
  Partially-Subjective Ratings}}. In \bibinfo{booktitle}{\emph{{Proceedings of
  the 4th AAAI Conference on Human Computation and Crowdsourcing (HCOMP)}}}.
  \bibinfo{pages}{149--158}.
\newblock


\bibitem[\protect\citeauthoryear{Nguyen, Wallace, and Lease}{Nguyen
  et~al\mbox{.}}{2015}]%
        {nguyen2015combining}
\bibfield{author}{\bibinfo{person}{An~Thanh Nguyen}, \bibinfo{person}{Byron~C
  Wallace}, {and} \bibinfo{person}{Matthew Lease}.}
  \bibinfo{year}{2015}\natexlab{}.
\newblock \showarticletitle{Combining crowd and expert labels using decision
  theoretic active learning}. In \bibinfo{booktitle}{\emph{Third AAAI
  Conference on Human Computation and Crowdsourcing}}.
\newblock


\bibitem[\protect\citeauthoryear{Nguyen, Rosenberg, Song, Gao, Tiwary,
  Majumder, and Deng}{Nguyen et~al\mbox{.}}{2016b}]%
        {nguyen2016ms}
\bibfield{author}{\bibinfo{person}{Tri Nguyen}, \bibinfo{person}{Mir
  Rosenberg}, \bibinfo{person}{Xia Song}, \bibinfo{person}{Jianfeng Gao},
  \bibinfo{person}{Saurabh Tiwary}, \bibinfo{person}{Rangan Majumder}, {and}
  \bibinfo{person}{Li Deng}.} \bibinfo{year}{2016}\natexlab{b}.
\newblock \showarticletitle{MS MARCO: A human generated machine reading
  comprehension dataset}. In \bibinfo{booktitle}{\emph{CoCo@ NIPS}}.
\newblock


\bibitem[\protect\citeauthoryear{Nobata, Tetreault, Thomas, Mehdad, and
  Chang}{Nobata et~al\mbox{.}}{2016}]%
        {nobata2016abusive}
\bibfield{author}{\bibinfo{person}{Chikashi Nobata}, \bibinfo{person}{Joel
  Tetreault}, \bibinfo{person}{Achint Thomas}, \bibinfo{person}{Yashar Mehdad},
  {and} \bibinfo{person}{Yi Chang}.} \bibinfo{year}{2016}\natexlab{}.
\newblock \showarticletitle{Abusive language detection in online user content}.
  In \bibinfo{booktitle}{\emph{Proceedings of the 25th international conference
  on world wide web}}. \bibinfo{pages}{145--153}.
\newblock


\bibitem[\protect\citeauthoryear{Noble}{Noble}{2018}]%
        {noble2018algorithms}
\bibfield{author}{\bibinfo{person}{Safiya~Umoja Noble}.}
  \bibinfo{year}{2018}\natexlab{}.
\newblock \bibinfo{booktitle}{\emph{Algorithms of oppression}}.
\newblock \bibinfo{publisher}{New York University Press}.
\newblock


\bibitem[\protect\citeauthoryear{Noronha, Hysen, Zhang, and Gajos}{Noronha
  et~al\mbox{.}}{2011}]%
        {noronha2011platemate}
\bibfield{author}{\bibinfo{person}{Jon Noronha}, \bibinfo{person}{Eric Hysen},
  \bibinfo{person}{Haoqi Zhang}, {and} \bibinfo{person}{Krzysztof~Z Gajos}.}
  \bibinfo{year}{2011}\natexlab{}.
\newblock \showarticletitle{Platemate: crowdsourcing nutritional analysis from
  food photographs}. In \bibinfo{booktitle}{\emph{Proceedings of the 24th
  annual ACM symposium on User interface software and technology}}.
  \bibinfo{pages}{1--12}.
\newblock


\bibitem[\protect\citeauthoryear{Northcutt, Athalye, and Mueller}{Northcutt
  et~al\mbox{.}}{2021}]%
        {northcutt2021pervasive}
\bibfield{author}{\bibinfo{person}{Curtis~G Northcutt}, \bibinfo{person}{Anish
  Athalye}, {and} \bibinfo{person}{Jonas Mueller}.}
  \bibinfo{year}{2021}\natexlab{}.
\newblock \showarticletitle{Pervasive Label Errors in Test Sets Destabilize
  Machine Learning Benchmarks}. In \bibinfo{booktitle}{\emph{Thirty-fifth
  Conference on Neural Information Processing Systems: Datasets and Benchmarks
  Track}}.
\newblock


\bibitem[\protect\citeauthoryear{NoSwearing}{NoSwearing}{[n.d.]}]%
        {noswearing}
\bibfield{author}{\bibinfo{person}{NoSwearing}.}
  \bibinfo{year}{[n.d.]}\natexlab{}.
\newblock \bibinfo{title}{List of Swear Words, Bad Words, \& Curse Words -
  Starting With A}.
\newblock \bibinfo{howpublished}{\url{https://www.noswearing.com/}}.
\newblock


\bibitem[\protect\citeauthoryear{Oard, Soergel, Doermann, Huang, Murray, Wang,
  Ramabhadran, Franz, Gustman, Mayfield, et~al\mbox{.}}{Oard
  et~al\mbox{.}}{2004}]%
        {oard2004building}
\bibfield{author}{\bibinfo{person}{Douglas~W Oard}, \bibinfo{person}{Dagobert
  Soergel}, \bibinfo{person}{David Doermann}, \bibinfo{person}{Xiaoli Huang},
  \bibinfo{person}{G~Craig Murray}, \bibinfo{person}{Jianqiang Wang},
  \bibinfo{person}{Bhuvana Ramabhadran}, \bibinfo{person}{Martin Franz},
  \bibinfo{person}{Samuel Gustman}, \bibinfo{person}{James Mayfield},
  {et~al\mbox{.}}} \bibinfo{year}{2004}\natexlab{}.
\newblock \showarticletitle{Building an information retrieval test collection
  for spontaneous conversational speech}. In
  \bibinfo{booktitle}{\emph{Proceedings of the 27th annual international ACM
  SIGIR conference on Research and development in information retrieval}}. ACM,
  \bibinfo{pages}{41--48}.
\newblock


\bibitem[\protect\citeauthoryear{Ousidhoum, Lin, Zhang, Song, and
  Yeung}{Ousidhoum et~al\mbox{.}}{2019}]%
        {ousidhoum2019multilingual}
\bibfield{author}{\bibinfo{person}{Nedjma Ousidhoum}, \bibinfo{person}{Zizheng
  Lin}, \bibinfo{person}{Hongming Zhang}, \bibinfo{person}{Yangqiu Song}, {and}
  \bibinfo{person}{Dit-Yan Yeung}.} \bibinfo{year}{2019}\natexlab{}.
\newblock \showarticletitle{Multilingual and multi-aspect hate speech
  analysis}.
\newblock \bibinfo{journal}{\emph{arXiv preprint arXiv:1908.11049}}
  (\bibinfo{year}{2019}).
\newblock


\bibitem[\protect\citeauthoryear{Pavlopoulos, Malakasiotis, and
  Androutsopoulos}{Pavlopoulos et~al\mbox{.}}{2017}]%
        {pavlopoulos2017deep}
\bibfield{author}{\bibinfo{person}{John Pavlopoulos},
  \bibinfo{person}{Prodromos Malakasiotis}, {and} \bibinfo{person}{Ion
  Androutsopoulos}.} \bibinfo{year}{2017}\natexlab{}.
\newblock \showarticletitle{Deep learning for user comment moderation}.
\newblock \bibinfo{journal}{\emph{arXiv preprint arXiv:1705.09993}}
  (\bibinfo{year}{2017}).
\newblock


\bibitem[\protect\citeauthoryear{Pennington, Socher, and Manning}{Pennington
  et~al\mbox{.}}{2014}]%
        {pennington2014glove}
\bibfield{author}{\bibinfo{person}{Jeffrey Pennington},
  \bibinfo{person}{Richard Socher}, {and} \bibinfo{person}{Christopher~D
  Manning}.} \bibinfo{year}{2014}\natexlab{}.
\newblock \showarticletitle{Glove: Global vectors for word representation}.
  \bibinfo{howpublished}{\url{http://nlp.stanford.edu/data/glove.twitter.27B.zip}}.
  In \bibinfo{booktitle}{\emph{Proceedings of the 2014 conference on empirical
  methods in natural language processing (EMNLP)}}.
  \bibinfo{pages}{1532--1543}.
\newblock


\bibitem[\protect\citeauthoryear{Perez, Jones, Englert, and Sachau}{Perez
  et~al\mbox{.}}{2010}]%
        {perez2010secondary}
\bibfield{author}{\bibinfo{person}{Lisa~M Perez}, \bibinfo{person}{Jeremy
  Jones}, \bibinfo{person}{David~R Englert}, {and} \bibinfo{person}{Daniel
  Sachau}.} \bibinfo{year}{2010}\natexlab{}.
\newblock \showarticletitle{Secondary traumatic stress and burnout among law
  enforcement investigators exposed to disturbing media images}.
\newblock \bibinfo{journal}{\emph{Journal of Police and Criminal Psychology}}
  \bibinfo{volume}{25}, \bibinfo{number}{2} (\bibinfo{year}{2010}),
  \bibinfo{pages}{113--124}.
\newblock


\bibitem[\protect\citeauthoryear{Poletto, Basile, Sanguinetti, Bosco, and
  Patti}{Poletto et~al\mbox{.}}{2021}]%
        {poletto2021resources}
\bibfield{author}{\bibinfo{person}{Fabio Poletto}, \bibinfo{person}{Valerio
  Basile}, \bibinfo{person}{Manuela Sanguinetti}, \bibinfo{person}{Cristina
  Bosco}, {and} \bibinfo{person}{Viviana Patti}.}
  \bibinfo{year}{2021}\natexlab{}.
\newblock \showarticletitle{Resources and benchmark corpora for hate speech
  detection: a systematic review}.
\newblock \bibinfo{journal}{\emph{Language Resources and Evaluation}}
  \bibinfo{volume}{55}, \bibinfo{number}{2} (\bibinfo{year}{2021}),
  \bibinfo{pages}{477--523}.
\newblock


\bibitem[\protect\citeauthoryear{Poletto, Stranisci, Sanguinetti, Patti, and
  Bosco}{Poletto et~al\mbox{.}}{2017}]%
        {poletto2017hate}
\bibfield{author}{\bibinfo{person}{Fabio Poletto}, \bibinfo{person}{Marco
  Stranisci}, \bibinfo{person}{Manuela Sanguinetti}, \bibinfo{person}{Viviana
  Patti}, {and} \bibinfo{person}{Cristina Bosco}.}
  \bibinfo{year}{2017}\natexlab{}.
\newblock \showarticletitle{Hate speech annotation: Analysis of an italian
  twitter corpus}. In \bibinfo{booktitle}{\emph{4th Italian Conference on
  Computational Linguistics, CLiC-it 2017}}, Vol.~\bibinfo{volume}{2006}.
  CEUR-WS, \bibinfo{pages}{1--6}.
\newblock


\bibitem[\protect\citeauthoryear{Prabhu, Dognin, and Singh}{Prabhu
  et~al\mbox{.}}{2019}]%
        {prabhu2019sampling}
\bibfield{author}{\bibinfo{person}{Ameya Prabhu}, \bibinfo{person}{Charles
  Dognin}, {and} \bibinfo{person}{Maneesh Singh}.}
  \bibinfo{year}{2019}\natexlab{}.
\newblock \showarticletitle{Sampling bias in deep active classification: An
  empirical study}.
\newblock \bibinfo{journal}{\emph{arXiv preprint arXiv:1909.09389}}
  (\bibinfo{year}{2019}).
\newblock


\bibitem[\protect\citeauthoryear{Rahman, Balakrishnan, Murthy, Kutlu, and
  Lease}{Rahman et~al\mbox{.}}{2021}]%
        {rahman21-neurips21-supplementary}
\bibfield{author}{\bibinfo{person}{Md~Mustafizur Rahman},
  \bibinfo{person}{Dinesh Balakrishnan}, \bibinfo{person}{Dhiraj Murthy},
  \bibinfo{person}{Mucahid Kutlu}, {and} \bibinfo{person}{Matthew Lease}.}
  \bibinfo{year}{2021}\natexlab{}.
\newblock \showarticletitle{{Supplementary Material: An Information Retrieval
  Approach to Building Datasets for Hate Speech Detection}}. In
  \bibinfo{booktitle}{\emph{Proceedings of the Thirty-fifth Conference on
  Neural Information Processing Systems (NeurIPS): Datasets and Benchmarks
  Track}}.
\newblock
\newblock
\shownote{Also available online at
  \url{https://www.ischool.utexas.edu/~ml/publications/}.}


\bibitem[\protect\citeauthoryear{Rahman, Kutlu, Elsayed, and Lease}{Rahman
  et~al\mbox{.}}{2020}]%
        {rahman_ICTIR_2020}
\bibfield{author}{\bibinfo{person}{Md~Mustafizur Rahman},
  \bibinfo{person}{Mucahid Kutlu}, \bibinfo{person}{Tamer Elsayed}, {and}
  \bibinfo{person}{Matthew Lease}.} \bibinfo{year}{2020}\natexlab{}.
\newblock \showarticletitle{Efficient Test Collection Construction via Active
  Learning}. In \bibinfo{booktitle}{\emph{Proceedings of the 2020 ACM SIGIR on
  International Conference on Theory of Information Retrieval}} (Virtual Event,
  Norway) \emph{(\bibinfo{series}{ICTIR '20})}. \bibinfo{publisher}{Association
  for Computing Machinery}, \bibinfo{address}{New York, NY, USA},
  \bibinfo{pages}{177–184}.
\newblock
\showISBNx{9781450380676}
\urldef\tempurl%
\url{https://doi.org/10.1145/3409256.3409837}
\showDOI{\tempurl}


\bibitem[\protect\citeauthoryear{Rajpurkar, Zhang, Lopyrev, and
  Liang}{Rajpurkar et~al\mbox{.}}{2016}]%
        {rajpurkar2016squad}
\bibfield{author}{\bibinfo{person}{Pranav Rajpurkar}, \bibinfo{person}{Jian
  Zhang}, \bibinfo{person}{Konstantin Lopyrev}, {and} \bibinfo{person}{Percy
  Liang}.} \bibinfo{year}{2016}\natexlab{}.
\newblock \showarticletitle{Squad: 100,000+ questions for machine comprehension
  of text}.
\newblock \bibinfo{journal}{\emph{arXiv preprint arXiv:1606.05250}}
  (\bibinfo{year}{2016}).
\newblock


\bibitem[\protect\citeauthoryear{roBERTa base}{roBERTa base}{2020}]%
        {TwitterroBERTa}
\bibfield{author}{\bibinfo{person}{Twitter roBERTa base}.}
  \bibinfo{year}{2020}\natexlab{}.
\newblock
  \bibinfo{howpublished}{\url{https://huggingface.co/cardiffnlp/twitter-roberta-base}}.
\newblock


\bibitem[\protect\citeauthoryear{Ross, Rist, Carbonell, Cabrera, Kurowsky, and
  Wojatzki}{Ross et~al\mbox{.}}{2017}]%
        {ross2017measuring}
\bibfield{author}{\bibinfo{person}{Bj{\"o}rn Ross}, \bibinfo{person}{Michael
  Rist}, \bibinfo{person}{Guillermo Carbonell}, \bibinfo{person}{Benjamin
  Cabrera}, \bibinfo{person}{Nils Kurowsky}, {and} \bibinfo{person}{Michael
  Wojatzki}.} \bibinfo{year}{2017}\natexlab{}.
\newblock \showarticletitle{Measuring the reliability of hate speech
  annotations: The case of the european refugee crisis}.
\newblock \bibinfo{journal}{\emph{arXiv preprint arXiv:1701.08118}}
  (\bibinfo{year}{2017}).
\newblock


\bibitem[\protect\citeauthoryear{R{\"o}ttger, Vidgen, Nguyen, Waseem, Margetts,
  and Pierrehumbert}{R{\"o}ttger et~al\mbox{.}}{2021}]%
        {rottger-etal-2021-hatecheck}
\bibfield{author}{\bibinfo{person}{Paul R{\"o}ttger}, \bibinfo{person}{Bertie
  Vidgen}, \bibinfo{person}{Dong Nguyen}, \bibinfo{person}{Zeerak Waseem},
  \bibinfo{person}{Helen Margetts}, {and} \bibinfo{person}{Janet
  Pierrehumbert}.} \bibinfo{year}{2021}\natexlab{}.
\newblock \showarticletitle{{H}ate{C}heck: Functional Tests for Hate Speech
  Detection Models}. In \bibinfo{booktitle}{\emph{Proceedings of the 59th
  Annual Meeting of the Association for Computational Linguistics and the 11th
  International Joint Conference on Natural Language Processing (Volume 1: Long
  Papers)}}. \bibinfo{publisher}{Association for Computational Linguistics},
  \bibinfo{address}{Online}, \bibinfo{pages}{41--58}.
\newblock
\urldef\tempurl%
\url{https://doi.org/10.18653/v1/2021.acl-long.4}
\showDOI{\tempurl}


\bibitem[\protect\citeauthoryear{Salton and Buckley}{Salton and
  Buckley}{1988}]%
        {salton1988term}
\bibfield{author}{\bibinfo{person}{Gerard Salton} {and}
  \bibinfo{person}{Christopher Buckley}.} \bibinfo{year}{1988}\natexlab{}.
\newblock \showarticletitle{Term-weighting approaches in automatic text
  retrieval}.
\newblock \bibinfo{journal}{\emph{Information processing \& management}}
  \bibinfo{volume}{24}, \bibinfo{number}{5} (\bibinfo{year}{1988}),
  \bibinfo{pages}{513--523}.
\newblock


\bibitem[\protect\citeauthoryear{Sambasivan, Kapania, Highfill, Akrong,
  Paritosh, and Aroyo}{Sambasivan et~al\mbox{.}}{2021}]%
        {sambasivan2021everyone}
\bibfield{author}{\bibinfo{person}{Nithya Sambasivan}, \bibinfo{person}{Shivani
  Kapania}, \bibinfo{person}{Hannah Highfill}, \bibinfo{person}{Diana Akrong},
  \bibinfo{person}{Praveen Paritosh}, {and} \bibinfo{person}{Lora~M Aroyo}.}
  \bibinfo{year}{2021}\natexlab{}.
\newblock \showarticletitle{“Everyone wants to do the model work, not the
  data work”: Data Cascades in High-Stakes AI}. In
  \bibinfo{booktitle}{\emph{proceedings of the 2021 CHI Conference on Human
  Factors in Computing Systems}}. \bibinfo{pages}{1--15}.
\newblock


\bibitem[\protect\citeauthoryear{Sanderson}{Sanderson}{2010}]%
        {sanderson2010test}
\bibfield{author}{\bibinfo{person}{Mark Sanderson}.}
  \bibinfo{year}{2010}\natexlab{}.
\newblock \showarticletitle{Test collection based evaluation of information
  retrieval systems}.
\newblock \bibinfo{journal}{\emph{Foundations and Trends{\textregistered} in
  Information Retrieval}} \bibinfo{volume}{4}, \bibinfo{number}{4}
  (\bibinfo{year}{2010}), \bibinfo{pages}{247--375}.
\newblock


\bibitem[\protect\citeauthoryear{Sanguinetti, Poletto, Bosco, Patti, and
  Stranisci}{Sanguinetti et~al\mbox{.}}{2018}]%
        {sanguinetti2018italian}
\bibfield{author}{\bibinfo{person}{Manuela Sanguinetti}, \bibinfo{person}{Fabio
  Poletto}, \bibinfo{person}{Cristina Bosco}, \bibinfo{person}{Viviana Patti},
  {and} \bibinfo{person}{Marco Stranisci}.} \bibinfo{year}{2018}\natexlab{}.
\newblock \showarticletitle{An italian twitter corpus of hate speech against
  immigrants}. In \bibinfo{booktitle}{\emph{Proceedings of the Eleventh
  International Conference on Language Resources and Evaluation (LREC 2018)}}.
\newblock


\bibitem[\protect\citeauthoryear{Sap, Card, Gabriel, Choi, and Smith}{Sap
  et~al\mbox{.}}{2019}]%
        {sap2019risk}
\bibfield{author}{\bibinfo{person}{Maarten Sap}, \bibinfo{person}{Dallas Card},
  \bibinfo{person}{Saadia Gabriel}, \bibinfo{person}{Yejin Choi}, {and}
  \bibinfo{person}{Noah~A Smith}.} \bibinfo{year}{2019}\natexlab{}.
\newblock \showarticletitle{The risk of racial bias in hate speech detection}.
  In \bibinfo{booktitle}{\emph{Proceedings of the 57th annual meeting of the
  association for computational linguistics}}. \bibinfo{pages}{1668--1678}.
\newblock


\bibitem[\protect\citeauthoryear{Schmidt and Wiegand}{Schmidt and
  Wiegand}{2017a}]%
        {schmidt2017survey}
\bibfield{author}{\bibinfo{person}{Anna Schmidt} {and} \bibinfo{person}{Michael
  Wiegand}.} \bibinfo{year}{2017}\natexlab{a}.
\newblock \showarticletitle{A survey on hate speech detection using natural
  language processing}. In \bibinfo{booktitle}{\emph{Proceedings of the Fifth
  International workshop on natural language processing for social media}}.
  \bibinfo{pages}{1--10}.
\newblock


\bibitem[\protect\citeauthoryear{Schmidt and Wiegand}{Schmidt and
  Wiegand}{2017b}]%
        {schmidt-wiegand-2017-survey}
\bibfield{author}{\bibinfo{person}{Anna Schmidt} {and} \bibinfo{person}{Michael
  Wiegand}.} \bibinfo{year}{2017}\natexlab{b}.
\newblock \showarticletitle{A Survey on Hate Speech Detection using Natural
  Language Processing}. In \bibinfo{booktitle}{\emph{Proceedings of the Fifth
  International Workshop on Natural Language Processing for Social Media}}.
  \bibinfo{publisher}{Association for Computational Linguistics},
  \bibinfo{address}{Valencia, Spain}, \bibinfo{pages}{1--10}.
\newblock
\urldef\tempurl%
\url{https://doi.org/10.18653/v1/W17-1101}
\showDOI{\tempurl}


\bibitem[\protect\citeauthoryear{Sen, Giesel, Gold, Hillmann, Lesicko, Naden,
  Russell, Wang, and Hecht}{Sen et~al\mbox{.}}{2015}]%
        {sen2015turkers}
\bibfield{author}{\bibinfo{person}{Shilad Sen}, \bibinfo{person}{Margaret~E
  Giesel}, \bibinfo{person}{Rebecca Gold}, \bibinfo{person}{Benjamin Hillmann},
  \bibinfo{person}{Matt Lesicko}, \bibinfo{person}{Samuel Naden},
  \bibinfo{person}{Jesse Russell}, \bibinfo{person}{Zixiao Wang}, {and}
  \bibinfo{person}{Brent Hecht}.} \bibinfo{year}{2015}\natexlab{}.
\newblock \showarticletitle{Turkers, Scholars," Arafat" and" Peace" Cultural
  Communities and Algorithmic Gold Standards}. In
  \bibinfo{booktitle}{\emph{Proceedings of the 18th acm conference on computer
  supported cooperative work \& social computing}}. \bibinfo{pages}{826--838}.
\newblock


\bibitem[\protect\citeauthoryear{Settles}{Settles}{2012}]%
        {settles2012active}
\bibfield{author}{\bibinfo{person}{Burr Settles}.}
  \bibinfo{year}{2012}\natexlab{}.
\newblock \showarticletitle{Active learning}.
\newblock \bibinfo{journal}{\emph{Synthesis Lectures on Artificial Intelligence
  and Machine Learning}} \bibinfo{volume}{6}, \bibinfo{number}{1}
  (\bibinfo{year}{2012}), \bibinfo{pages}{1--114}.
\newblock


\bibitem[\protect\citeauthoryear{Shankar, Halpern, Breck, Atwood, Wilson, and
  Sculley}{Shankar et~al\mbox{.}}{2017}]%
        {shankar2017no}
\bibfield{author}{\bibinfo{person}{Shreya Shankar}, \bibinfo{person}{Yoni
  Halpern}, \bibinfo{person}{Eric Breck}, \bibinfo{person}{James Atwood},
  \bibinfo{person}{Jimbo Wilson}, {and} \bibinfo{person}{D Sculley}.}
  \bibinfo{year}{2017}\natexlab{}.
\newblock \showarticletitle{No Classification without Representation: Assessing
  Geodiversity Issues in Open Data Sets for the Developing World}.
\newblock \bibinfo{journal}{\emph{stat}}  \bibinfo{volume}{1050}
  (\bibinfo{year}{2017}), \bibinfo{pages}{22}.
\newblock


\bibitem[\protect\citeauthoryear{Sood, Antin, and Churchill}{Sood
  et~al\mbox{.}}{2012}]%
        {sood2012profanity}
\bibfield{author}{\bibinfo{person}{Sara Sood}, \bibinfo{person}{Judd Antin},
  {and} \bibinfo{person}{Elizabeth Churchill}.}
  \bibinfo{year}{2012}\natexlab{}.
\newblock \showarticletitle{Profanity use in online communities}. In
  \bibinfo{booktitle}{\emph{Proceedings of the SIGCHI Conference on Human
  Factors in Computing Systems}}. \bibinfo{pages}{1481--1490}.
\newblock


\bibitem[\protect\citeauthoryear{Steiger, Bharucha, Venkatagiri, Riedl, and
  Lease}{Steiger et~al\mbox{.}}{2021}]%
        {steiger2021psychological}
\bibfield{author}{\bibinfo{person}{Miriah Steiger}, \bibinfo{person}{Timir~J
  Bharucha}, \bibinfo{person}{Sukrit Venkatagiri}, \bibinfo{person}{Martin~J
  Riedl}, {and} \bibinfo{person}{Matthew Lease}.}
  \bibinfo{year}{2021}\natexlab{}.
\newblock \showarticletitle{The Psychological Well-Being of Content Moderators:
  The Emotional Labor of Commercial Moderation and Avenues for Improving
  Support}. In \bibinfo{booktitle}{\emph{Proceedings of the 2021 CHI Conference
  on Human Factors in Computing Systems}}. \bibinfo{pages}{1--14}.
\newblock


\bibitem[\protect\citeauthoryear{Strassel, Graff, Martey, and Cieri}{Strassel
  et~al\mbox{.}}{2000}]%
        {strassel2000quality}
\bibfield{author}{\bibinfo{person}{Stephanie Strassel}, \bibinfo{person}{David
  Graff}, \bibinfo{person}{Nii Martey}, {and} \bibinfo{person}{Christopher
  Cieri}.} \bibinfo{year}{2000}\natexlab{}.
\newblock \showarticletitle{Quality Control in Large Annotation Projects
  Involving Multiple Judges: The Case of the TDT Corpora}. In
  \bibinfo{booktitle}{\emph{Proceedings of the Second International Conference
  on Language Resources and Evaluation (LREC’00)}}.
\newblock


\bibitem[\protect\citeauthoryear{Strout, Zhang, and Mooney}{Strout
  et~al\mbox{.}}{2019}]%
        {strout2019human}
\bibfield{author}{\bibinfo{person}{Julia Strout}, \bibinfo{person}{Ye Zhang},
  {and} \bibinfo{person}{Raymond Mooney}.} \bibinfo{year}{2019}\natexlab{}.
\newblock \showarticletitle{Do Human Rationales Improve Machine Explanations?}.
  In \bibinfo{booktitle}{\emph{Proceedings of the 2019 ACL Workshop
  BlackboxNLP: Analyzing and Interpreting Neural Networks for NLP}}.
  \bibinfo{pages}{56--62}.
\newblock


\bibitem[\protect\citeauthoryear{Thompson}{Thompson}{2016}]%
        {thompson2016anti}
\bibfield{author}{\bibinfo{person}{Neil Thompson}.}
  \bibinfo{year}{2016}\natexlab{}.
\newblock \bibinfo{booktitle}{\emph{Anti-discriminatory practice: Equality,
  diversity and social justice}}.
\newblock \bibinfo{publisher}{Macmillan International Higher Education}.
\newblock


\bibitem[\protect\citeauthoryear{Tommasi, Patricia, Caputo, and
  Tuytelaars}{Tommasi et~al\mbox{.}}{2017}]%
        {tommasi2017deeper}
\bibfield{author}{\bibinfo{person}{Tatiana Tommasi}, \bibinfo{person}{Novi
  Patricia}, \bibinfo{person}{Barbara Caputo}, {and} \bibinfo{person}{Tinne
  Tuytelaars}.} \bibinfo{year}{2017}\natexlab{}.
\newblock \showarticletitle{A deeper look at dataset bias}.
\newblock In \bibinfo{booktitle}{\emph{Domain adaptation in computer vision
  applications}}. \bibinfo{publisher}{Springer}, \bibinfo{pages}{37--55}.
\newblock


\bibitem[\protect\citeauthoryear{Uebersax}{Uebersax}{1988}]%
        {uebersax1988validity}
\bibfield{author}{\bibinfo{person}{John~S Uebersax}.}
  \bibinfo{year}{1988}\natexlab{}.
\newblock \showarticletitle{Validity inferences from interobserver agreement.}
\newblock \bibinfo{journal}{\emph{Psychological Bulletin}}
  \bibinfo{volume}{104}, \bibinfo{number}{3} (\bibinfo{year}{1988}),
  \bibinfo{pages}{405}.
\newblock


\bibitem[\protect\citeauthoryear{Vidgen and Derczynski}{Vidgen and
  Derczynski}{2020}]%
        {vidgen2020directions}
\bibfield{author}{\bibinfo{person}{Bertie Vidgen} {and} \bibinfo{person}{Leon
  Derczynski}.} \bibinfo{year}{2020}\natexlab{}.
\newblock \showarticletitle{Directions in abusive language training data, a
  systematic review: Garbage in, garbage out}.
\newblock \bibinfo{journal}{\emph{PloS one}} \bibinfo{volume}{15},
  \bibinfo{number}{12} (\bibinfo{year}{2020}), \bibinfo{pages}{e0243300}.
\newblock


\bibitem[\protect\citeauthoryear{Wagaman, Geiger, Shockley, and Segal}{Wagaman
  et~al\mbox{.}}{2015}]%
        {wagaman2015role}
\bibfield{author}{\bibinfo{person}{M~Alex Wagaman}, \bibinfo{person}{Jennifer~M
  Geiger}, \bibinfo{person}{Clara Shockley}, {and} \bibinfo{person}{Elizabeth~A
  Segal}.} \bibinfo{year}{2015}\natexlab{}.
\newblock \showarticletitle{The role of empathy in burnout, compassion
  satisfaction, and secondary traumatic stress among social workers}.
\newblock \bibinfo{journal}{\emph{Social work}} \bibinfo{volume}{60},
  \bibinfo{number}{3} (\bibinfo{year}{2015}), \bibinfo{pages}{201--209}.
\newblock


\bibitem[\protect\citeauthoryear{Warner and Hirschberg}{Warner and
  Hirschberg}{2012}]%
        {warner2012detecting}
\bibfield{author}{\bibinfo{person}{William Warner} {and} \bibinfo{person}{Julia
  Hirschberg}.} \bibinfo{year}{2012}\natexlab{}.
\newblock \showarticletitle{Detecting hate speech on the world wide web}. In
  \bibinfo{booktitle}{\emph{Proceedings of the second workshop on language in
  social media}}. \bibinfo{pages}{19--26}.
\newblock


\bibitem[\protect\citeauthoryear{Waseem}{Waseem}{2016a}]%
        {waseem2016you}
\bibfield{author}{\bibinfo{person}{Zeerak Waseem}.}
  \bibinfo{year}{2016}\natexlab{a}.
\newblock \showarticletitle{Are you a racist or am i seeing things? annotator
  influence on hate speech detection on twitter}. In
  \bibinfo{booktitle}{\emph{Proceedings of the first workshop on NLP and
  computational social science}}. \bibinfo{pages}{138--142}.
\newblock


\bibitem[\protect\citeauthoryear{Waseem}{Waseem}{2016b}]%
        {waseem2016automatic}
\bibfield{author}{\bibinfo{person}{Zeerak Waseem}.}
  \bibinfo{year}{2016}\natexlab{b}.
\newblock \emph{\bibinfo{title}{Automatic hate speech detection}}.
\newblock \bibinfo{thesistype}{Ph.D. Dissertation}. \bibinfo{school}{Master’s
  thesis, University of Copenhagen}.
\newblock


\bibitem[\protect\citeauthoryear{Waseem and Hovy}{Waseem and Hovy}{2016}]%
        {waseem-hovy-2016-hateful}
\bibfield{author}{\bibinfo{person}{Zeerak Waseem} {and} \bibinfo{person}{Dirk
  Hovy}.} \bibinfo{year}{2016}\natexlab{}.
\newblock \showarticletitle{Hateful Symbols or Hateful People? Predictive
  Features for Hate Speech Detection on {T}witter}. In
  \bibinfo{booktitle}{\emph{Proceedings of the {NAACL} Student Research
  Workshop}}. \bibinfo{publisher}{Association for Computational Linguistics},
  \bibinfo{address}{San Diego, California}, \bibinfo{pages}{88--93}.
\newblock
\urldef\tempurl%
\url{https://doi.org/10.18653/v1/N16-2013}
\showDOI{\tempurl}


\bibitem[\protect\citeauthoryear{Watanabe, Bouazizi, and Ohtsuki}{Watanabe
  et~al\mbox{.}}{2018}]%
        {Watanabe}
\bibfield{author}{\bibinfo{person}{Hajime Watanabe}, \bibinfo{person}{Mondher
  Bouazizi}, {and} \bibinfo{person}{Tomoaki Ohtsuki}.}
  \bibinfo{year}{2018}\natexlab{}.
\newblock \showarticletitle{Hate Speech on Twitter: A Pragmatic Approach to
  Collect Hateful and Offensive Expressions and Perform Hate Speech Detection}.
\newblock \bibinfo{journal}{\emph{IEEE Access}}  \bibinfo{volume}{6}
  (\bibinfo{year}{2018}), \bibinfo{pages}{13825--13835}.
\newblock
\urldef\tempurl%
\url{https://doi.org/10.1109/ACCESS.2018.2806394}
\showDOI{\tempurl}


\bibitem[\protect\citeauthoryear{Wich, Bauer, and Groh}{Wich
  et~al\mbox{.}}{2020}]%
        {wich2020impact}
\bibfield{author}{\bibinfo{person}{Maximilian Wich}, \bibinfo{person}{Jan
  Bauer}, {and} \bibinfo{person}{Georg Groh}.} \bibinfo{year}{2020}\natexlab{}.
\newblock \showarticletitle{Impact of politically biased data on hate speech
  classification}. In \bibinfo{booktitle}{\emph{Proceedings of the Fourth
  Workshop on Online Abuse and Harms}}. \bibinfo{pages}{54--64}.
\newblock


\bibitem[\protect\citeauthoryear{Wiegand, Ruppenhofer, and Kleinbauer}{Wiegand
  et~al\mbox{.}}{2019}]%
        {wiegand2019detection}
\bibfield{author}{\bibinfo{person}{Michael Wiegand}, \bibinfo{person}{Josef
  Ruppenhofer}, {and} \bibinfo{person}{Thomas Kleinbauer}.}
  \bibinfo{year}{2019}\natexlab{}.
\newblock \showarticletitle{Detection of abusive language: the problem of
  biased datasets}. In \bibinfo{booktitle}{\emph{Proceedings of the 2019
  conference of the North American Chapter of the Association for Computational
  Linguistics: human language technologies, volume 1 (long and short papers)}}.
  \bibinfo{pages}{602--608}.
\newblock


\bibitem[\protect\citeauthoryear{Willett}{Willett}{2006}]%
        {willett2006porter}
\bibfield{author}{\bibinfo{person}{Peter Willett}.}
  \bibinfo{year}{2006}\natexlab{}.
\newblock \showarticletitle{The Porter stemming algorithm: then and now}.
\newblock \bibinfo{journal}{\emph{Program}} (\bibinfo{year}{2006}).
\newblock


\bibitem[\protect\citeauthoryear{Wulczyn, Thain, and Dixon}{Wulczyn
  et~al\mbox{.}}{2017}]%
        {wulczyn2017ex}
\bibfield{author}{\bibinfo{person}{Ellery Wulczyn}, \bibinfo{person}{Nithum
  Thain}, {and} \bibinfo{person}{Lucas Dixon}.}
  \bibinfo{year}{2017}\natexlab{}.
\newblock \showarticletitle{Ex machina: Personal attacks seen at scale}. In
  \bibinfo{booktitle}{\emph{Proceedings of the 26th international conference on
  world wide web}}. \bibinfo{pages}{1391--1399}.
\newblock


\bibitem[\protect\citeauthoryear{Xia, Qin, Chen, Bian, Yu, and Liu}{Xia
  et~al\mbox{.}}{2017}]%
        {xia2017dual}
\bibfield{author}{\bibinfo{person}{Yingce Xia}, \bibinfo{person}{Tao Qin},
  \bibinfo{person}{Wei Chen}, \bibinfo{person}{Jiang Bian},
  \bibinfo{person}{Nenghai Yu}, {and} \bibinfo{person}{Tie-Yan Liu}.}
  \bibinfo{year}{2017}\natexlab{}.
\newblock \showarticletitle{Dual supervised learning}. In
  \bibinfo{booktitle}{\emph{International Conference on Machine Learning}}.
  PMLR, \bibinfo{pages}{3789--3798}.
\newblock


\bibitem[\protect\citeauthoryear{Ybarra, Mitchell, Wolak, and Finkelhor}{Ybarra
  et~al\mbox{.}}{2006}]%
        {ybarra2006examining}
\bibfield{author}{\bibinfo{person}{Michele~L Ybarra},
  \bibinfo{person}{Kimberly~J Mitchell}, \bibinfo{person}{Janis Wolak}, {and}
  \bibinfo{person}{David Finkelhor}.} \bibinfo{year}{2006}\natexlab{}.
\newblock \showarticletitle{Examining characteristics and associated distress
  related to Internet harassment: findings from the Second Youth Internet
  Safety Survey}.
\newblock \bibinfo{journal}{\emph{Pediatrics}} \bibinfo{volume}{118},
  \bibinfo{number}{4} (\bibinfo{year}{2006}), \bibinfo{pages}{e1169--e1177}.
\newblock


\bibitem[\protect\citeauthoryear{Zhang, Marshall, and Wallace}{Zhang
  et~al\mbox{.}}{2016}]%
        {zhang2016rationale}
\bibfield{author}{\bibinfo{person}{Ye Zhang}, \bibinfo{person}{Iain Marshall},
  {and} \bibinfo{person}{Byron~C Wallace}.} \bibinfo{year}{2016}\natexlab{}.
\newblock \showarticletitle{Rationale-augmented convolutional neural networks
  for text classification}. In \bibinfo{booktitle}{\emph{Proceedings of the
  Conference on Empirical Methods in Natural Language Processing. Conference on
  Empirical Methods in Natural Language Processing}},
  Vol.~\bibinfo{volume}{2016}. NIH Public Access, \bibinfo{pages}{795}.
\newblock


\bibitem[\protect\citeauthoryear{Zhang, Zhang, Lease, and Gwizdka}{Zhang
  et~al\mbox{.}}{2014}]%
        {Zhang-sigir14}
\bibfield{author}{\bibinfo{person}{Yinglong Zhang}, \bibinfo{person}{Jin
  Zhang}, \bibinfo{person}{Matthew Lease}, {and} \bibinfo{person}{Jacek
  Gwizdka}.} \bibinfo{year}{2014}\natexlab{}.
\newblock \showarticletitle{Multidimensional Relevance Modeling via
  Psychometrics and Crowdsourcing}. In \bibinfo{booktitle}{\emph{{Proceedings
  of the 37th international ACM SIGIR conference on Research and Development in
  Information Retrieval}}}. \bibinfo{pages}{435--444}.
\newblock


\end{thebibliography}

\end{document}


\maketitle

\begin{abstract}
Building a benchmark dataset for hate speech detection presents various challenges. Firstly, because hate speech is relatively rare, random sampling of tweets to annotate is very inefficient in finding hate speech. To address this, prior datasets often include only tweets matching known ``hate words''. However, restricting data to a pre-defined vocabulary may exclude portions of the real-world phenomenon we seek to model. A second challenge is that definitions of hate speech tend to be highly varying and subjective. Annotators having diverse prior notions of hate speech may not only disagree with one another but also struggle to conform to specified labeling guidelines. Our key insight is that the rarity and subjectivity of hate speech are akin to that of relevance in information retrieval (IR). This connection suggests that well-established methodologies for creating IR test collections can be usefully applied to create better benchmark datasets for hate speech. To intelligently and efficiently select which tweets to annotate, we apply standard IR techniques of {\em pooling} and {\em active learning}. To improve both consistency and value of annotations, we apply {\em task decomposition} and {\em annotator rationale} techniques. We share a new benchmark dataset for hate speech detection on Twitter that provides broader coverage of hate than prior datasets. We also show a dramatic drop in accuracy of existing detection models when tested on these broader forms of hate. Annotator rationales we collect not only justify labeling decisions but also enable future work opportunities for dual-supervision and/or explanation generation in modeling. Further details of our approach can be found in the supplementary materials.

\end{abstract}

{\bf Content Warning}: We discuss hate speech and provide examples that might be disturbing to read. 

\section{Introduction}
\label{section:introduction}

Online hate speech constitutes a vast and growing problem in social media \cite{halevy2020preserving, macavaney2019hate, jurgens2019just, fortuna2018survey, schmidt2017survey, davidson2017automated}. For example, \citet{halevy2020preserving} note that the wide variety of content violations and problem scale on Facebook defies manual detection, including the rate of spread and harm such content may cause in the world. Automated detection methods can be used to block content, select and prioritize content for human review, and/or restrict circulation until human review occurs. This need for automated detection has naturally given rise to the creation of labeled datasets for hate speech \cite{poletto2021resources, hatespeechdata}.

Datasets play a pivotal role in machine learning, translating real-world phenomena into a surrogate research environments within which we formulate computational tasks and perform modeling. Training data defines the totality of model supervision, while testing data defines the yardstick by which we measure empirical success and field progress. Benchmark datasets thus serve to catalyze research and define the world within which our models operate. However, research to improve models is often prioritized over research to improve the data environments in which models operate, even though dataset flaws and limitations can lead to significant practical problems or harm \cite{vidgen2020directions, laaksonen2020datafication, madukwe2020data,  sambasivan2021everyone, northcutt2021pervasive}. \citet{grondahl2018all} argue that for hate speech, the nature and composition of the datasets are more important than the models used due to extreme variation in annotating hate speech, including definition, categories, annotation guidelines, types of annotators, and aggregation of annotations.

\begin{figure}[h]
\centerline{\includegraphics[width=0.49\textwidth]{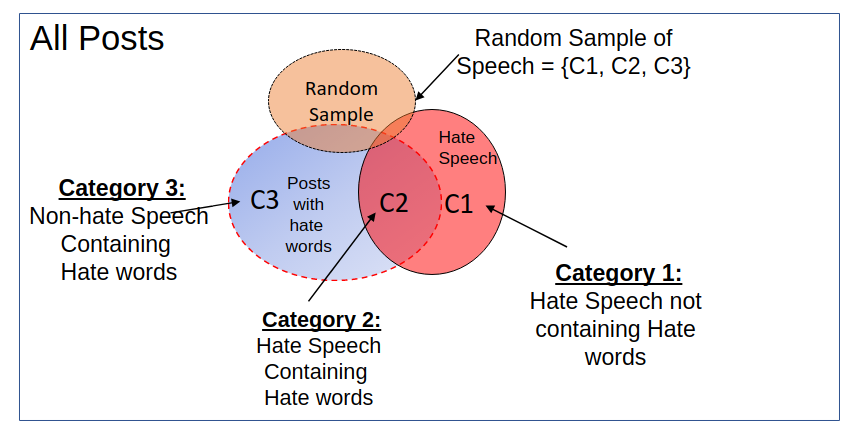}} 
\caption{Hate speech coverage resulting from different choices of which social media posts to annotate. Given some list of ``hate words'' by which to filter posts, some matching posts are indeed hateful (region C2, {\em true positives}) while other matching posts are benign (region C3, {\em false positives}). Region C1 indicates {\em false negatives}: hate speech missed by the filter and mistakenly excluded. Random sampling correctly overlaps C1+C2 but is highly inefficient in coverage.}
\label{Figure:landscape_of_hate_speech}
\vspace{-1em}
\end{figure}
\begin{table*}[h]
\begin{center}
\scriptsize
\caption{Example tweets from across the Venn diagram regions shown in Figure \ref{Figure:landscape_of_hate_speech}.} 
\begin{tabular}{c  p{0.82\linewidth}}
 \hline
 {\bf Figure \ref{Figure:landscape_of_hate_speech} Region} & {\bf Documents} \\
 \hline
 
 C1 & ``When you hold all 3 branches and still can't get anything done...you must be a republican." \\
 \hline
 C1 & ``You can yell at these Libs all day. They don't listen. But if you ridicule them nationally they can't take it" \\
 \hline

 C2 & ``Real Rednecks don't use the internet you {\color{red}pussy}'' \\
 \hline
 C2 & ``guarantee slavery wasnt even tht bad...you know {\color{red}niggas}  over exxagerate dramatize everything''\\
 \hline
 C3 & ``Don't call me {\color{red}black} no more. That word is just a color, it ain't facts no more" \\
 \hline
C3 & ``AGENT Are you crazy? You'll never make it in the white {\color{red}trash} rap/rock genre with a name like that SKIDMORE ROCKEFELLER" \\
 \hline
\label{table:example_of_hate_categories}
\end{tabular}
\end{center}
\vspace{-1em}
\end{table*}

Fortunately, many valuable datasets already exist for detecting hate speech \cite{hatespeechdata, poletto2021resources, waseem-hovy-2016-hateful, waseem2016you, davidson2017automated, founta2018large, de-gibert-etal-2018-hate, golbeck2017large}. However, each dataset can be seen to embody an underlying design tradeoff (often implicit) in how to balance cost vs.\ coverage of the phenomenon of hate speech, in all of the many forms of expression in which it manifests. At one extreme, random sampling ensures representative coverage but is highly inefficient (e.g., less than  3\% of Twitter posts are hateful \citep{founta2018large}). At the other extreme, one can annotate only those tweets matching a pre-defined vocabulary of ``hate words'' \cite{hatebase,noswearing} whose presence is strongly correlated with hateful utterances \citep{waseem-hovy-2016-hateful, waseem2016automatic, davidson2017automated, golbeck2017large, grimminger2021hate}. By restricting a dataset to only those tweets matching a pre-defined vocabulary, a higher percentage of hateful content can be found. However, this sacrifices representative coverage for cost-savings, yielding a biased dataset whose distribution diverges from the real world we seek to model and to apply these models to in practice \cite{kennedy2018gab}. If we only look for expressions of hate matching known word lists, the resulting dataset will completely miss any expressions of hate beyond this prescribed vocabulary. This is akin to traditional expert systems that relied entirely on hand-crafted, deterministic rules for classification and failed to generalize beyond their narrow rule sets. Such a resulting benchmark would provide only partial representation for the real-world phenomenon of interest.

{\bf Figure \ref{Figure:landscape_of_hate_speech}}
presents a Venn diagram illustrating this. {\bf Table \ref{table:example_of_hate_categories}} examples further highlight the weakness of only annotating posts matching known ``hate words''  \citep{davidson2017automated, waseem-hovy-2016-hateful, grimminger2021hate, golbeck2017large}, covering only regions C2-C3. The prevalence of hate found in such datasets is also limited by the word lists used \citep{sood2012profanity}. 

We can view the traditional practice above as following a two-stage pipeline. Firstly, a simple Boolean search is performed to retrieve all tweets matching a manually-curated vocabulary of known hate words. Secondly, because the above retrieval set is often quite large, random down-sampling is applied to reduce the set to a smaller, more affordable scale for annotation. In information retrieval (IR), a similar pipeline of human annotation following Boolean search has been traditionally used in legal {\em e-discovery} and medical {\em systematic review} tasks \cite{Lease16-medir}, though the Boolean filter in those domains casts a very wide net to maximize recall, whereas hate word lists tend to emphasize precision to ensure a high percentage (and efficient annotation) of hateful content. However, just as probabilistic models have largely superseded traditional deterministic, rule-based expert systems, we expect that a probabilistic retrieval model can more intelligently and efficiently select which content to annotate. 

A second key challenge in constructing hate speech datasets is that what constitutes hate speech is quite subjective, with many competing definitions across legal, regional, platform, and personal contexts \citep{davidson2017automated, waseem-hovy-2016-hateful, fortuna2018survey}. Consequently, annotators having diverse prior notions of hate speech often disagree with one another (especially when labeling guidelines allow more subjective freedom), and may also struggle to conform to stricter labeling guidelines. The use of inexpert crowd annotators can further exacerbate concerns regarding label reliability \cite{sen2015turkers,northcutt2021pervasive}. 

A key insight of our study is that the rarity and subjectivity of hate speech are akin to that of relevance in information retrieval (IR) \citep{sanderson2010test}. This suggests that established methods for creating IR {\em test collections} might also be applied to create better hate speech benchmark datasets. To intelligently and efficiently select which content to annotate for hate speech, we apply two known IR techniques for building test collections: {\em pooling} \citep{sanderson2010test} and {\em active learning} (AL) \citep{cormack_autonomy_2015, rahman_ICTIR_2020}. Both approaches begin with a very large random sample of social media posts to search (i.e., the {\em document collection}). With pooling, we use existing hate speech datasets and models to train a diverse ensemble of predictive models. We then prioritize posts for annotation by their predicted probability of being hateful, restricting annotation to the resulting {\em pool} of highly-ranked posts. For nearly 30 years, NIST TREC has applied such pooling techniques with a diverse set of ranking models in order to optimize the coverage vs.\ cost tradeoff in building test collections for IR, yielding benchmark datasets for fair and robust evaluation of IR systems. AL, on the other hand, requires only an initial set of {\em seed} posts from which a classifier is progressively trained to intelligently select which posts should be labeled next by human annotators. The tight annotation-training feedback loop provides greater efficiency in annotation, and unlike pooling, it does not require (nor is biased by) existing hate speech datasets.\footnote{Strictly speaking, initial seed documents used with AL may also include bias that influences training.} 
We re-iterate that pooling and AL are established methods in IR; our translational contribution is showing that these techniques can also be usefully applied to efficiently build hate speech datasets without vocabulary filters. 

To address subjectivity in annotation, we apply two other techniques from the IR and crowdsourcing literatures. Firstly, applying {\em task decomposition} \cite{Zhang-sigir14,noronha2011platemate}, we decompose the subjective and complex judging task into a structured set of smaller tasks. Secondly, we design the annotation task to let us measure annotator {\em self-consistency} \cite{Zhang-sigir14}. This provides a means of establishing label validity on the basis of an annotator being internally consistent, which is important for subjective labeling tasks in which we expect {\em inter-annotator} disagreement across annotators. Finally, we require that annotators provide constrained {\em rationales} justifying their labeling decisions \cite{mcdonnell16-hcomp,Kutlu18-sigir}, which prior work has found to improve label accuracy, verifiability, and utility. 

Using pooling, we create a new benchmark dataset\footnote{\url{https://github.com/mdmustafizurrahman/An-Information-Retrieval-Approach-to-Building-Datasets-for-Hate-Speech-Detection}} for hate speech detection on Twitter, consisting of 9,667 tweets in total: 4,668 labeled as part of iterative pilot experiments refining quality, and our final high quality set of 4,999 used in our experiments. Each tweet is annotated by three workers on Amazon Mechanical Turk, at a  total dataset annotation cost of approximately \$5K USD. 
We assess this dataset with respect to the {\em prevalence} (i.e., annotation efficiency) and {\em relative coverage} (breadth) of hate found, as well as evaluation of detection models on these broader forms of hate.  

Firstly, we show in Section \ref{sec:props} that pooling yields 14.6\% {\em relative coverage}, far better than the best prior work \cite{founta2018large}'s combination of random sampling and keyword filtering (10.4\%). Secondly, regarding efficiency in selecting what content to annotate, 14.1\% of our annotated content is found to be hateful, a {\em prevalence} that exceeds a number of prior datasets  \citep{davidson2017automated, fortuna2018survey, grimminger2021hate} while simultaneously also providing the aforementioned greater coverage. Note that prevalence can be inflated by only annotating a highly-restricted vocabulary or set of user posts; e.g., while \citet{waseem-hovy-2016-hateful} achieve 31\% prevalence, their dataset is highly skewed, with a single user generating 96\% of all racist tweets, while another user produces 44\% of all sexist tweets \cite{arango2019hate}. Just as precision and recall are balanced in classification, we wish to balance prevalence (efficiency) and coverage (fidelity) in creating a benchmark dataset that is faithful to the phenomenon while being affordable to create. Finally, we benchmark several recent hate speech detection models  \citep{devlin2018bert, agrawal2018deep, badjatiya2017deep} and find that the performance of these models drops drastically when tested on these broader forms of hate in our dataset (Table \ref{Table:Benchamark_results}). 

To further improve annotation efficiency, and to further reduce dependence on prior datasets (and potential bias from them), we also report retrospective experiments that assess AL on our dataset created via pooling. We compare several known AL strategies and feature representations, showing that AL can effectively find around 80\% of the hateful content at 50\% of the cost of pooling.  


As a word of caution, while machine learning models are usually intended to serve the social good (e.g., to reduce online hate speech), it is now well known that machine learning models can also aggravate inequality, perpetuate discrimination, and inflict harm \cite{noble2018algorithms, eubanksautomating}. As noted earlier, one source of such harms are dataset flaws and limitations that are unknown, ignored, or unmitigated 
\cite{vidgen2020directions, laaksonen2020datafication, madukwe2020data,  sambasivan2021everyone, northcutt2021pervasive, boyd2021datasheets}. For this reason, we emphasize that hate speech detection researchers and practitioners should attend carefully to these issues when deploying any automated hate speech detection system, including models trained on our dataset. We also want to draw attention to an important issue largely unremarked in prior work on hate speech dataset research: the potential health risks to human annotators from sustained exposure to such disturbing content \cite{steiger2021psychological}.

Please see our supplementary material \cite{rahman21-neurips21-supplementary} for additional discussion of related work, our approach, resulting dataset properties, risks to annotators, and other limitations and risks of our study.

\section{Methods}

To intelligently select which tweets to annotate in order to optimize the prevalence vs.\ coverage tradeoff in our annotated hate speech dataset, we apply two approaches from constructing test collections in information retrieval (IR): pooling and active learning. Unlike most prior work, we do not use any keywords to filter the set of tweets. Instead, we first collect a large, random (and unbiased) sample of tweets via Twitter's API in which to search for hateful content. 
Following IR terminology, we refer to each tweet as a {\em document} and the corpus as a {\em document collection}.

\subsection{Selecting Tweets via Pooling}
\label{section:pooling_algorithm}
Pooling \citep{sanderson2010test} applies a set of trained machine learning models to predict the hatefulness of documents; the documents that are predicted most likely to be hateful are then selected for human annotation. However, we found that annotating only the top-ranked documents found less diversity in forms of hate than found more broadly online (another example of the {\em prevalence} vs.\ {\em coverage} tradeoff). To further promote diversity, we instead perform a simplified form of stratified sampling. Specifically, we define a minimum threshold $t$, select all tweets with whose predicted hatefulness exceeds $t$ for any model in our ensemble, and then randomly downsample this set to $b$ tweets, given annotation budget $b$.  Intuitively, lower values of $t$ inject greater randomness into selection, promiting diversity. 

Our pooling algorithm is detailed in {\bf Algorithm \ref{algorithm:Pooling_hate_speech}}. Given document collection $X$, we select a subset of $b$ documents to annotate for our hate speech dataset $R$.  Our algorithm requires three inputs: i) a set of prior hate speech datasets $D$; ii) a set of classifiers $C$; and iii) the aforementioned threshold $t$. Given a prior hate speech dataset $D_i$ and a classifier  $C_j$, we induce a machine learning model $\hat{C_j^i}$, by training $C_j$ on $D_i$ (Line 4). Then we employ this $\hat{C_j^i}$ to predict a hatefulness score of every document $x$ in the  collection $X$ (Line 5). If the predicted score of document $x$ is greater than the provided threshold $t$, we add it to the set $S$ (Lines 6 - 8). This process iterates for all  datasets and  classifiers in the set $D$ and $C$ (Lines 2 - 8), respectively. Finally, we randomly select $b$ unique documents from $S$ (Line 9), as discussed earlier. 
to construct the hate speech dataset $R$ (Line 10).

\begin{algorithm}[t]
    \SetKwInOut{Input}{Input}
    \SetKwInOut{Output}{Output}
    \Input{~~Document collection $X$ $\bullet$ Set of prior hate speech datasets D \newline $\bullet$ Set of classifiers C $\bullet$ total budget $b$ $\bullet$ selection threshold $t$}
    \Output{~~Set of documents to be annotated hate speech dataset $R$}
        $S \leftarrow  \emptyset$ \\ 
        \For{$i \gets 1$ \KwTo $|D|$}{ 
            \For{$j \gets 1$ \KwTo $|C|$}{ 
                $\hat{C_j^i} \leftarrow $ train\_model$(C_j$, $D_i$) \Comment train classifier $C_j$ using data $D_i$ \\
                $ \forall x\in X$ predict hatefulness of document $x$ using $\hat{C_j^i}$\\
                \For{$k \gets 1$ \KwTo $|X|$}{ 
                    \algorithmicif\ the predicted hatefulness score of $x^k \ge t$ \\ \algorithmicthen\  $S \leftarrow S \cup x^k$ \\ 
                }
            }
        }
       $R \leftarrow $ Randomly sample $b$ unique documents from $S$ \Comment Set of selected documents \\     

\caption{Pooling Approach for Selecting Documents to Annotate}
\label{algorithm:Pooling_hate_speech}
\end{algorithm}



\subsection{Selecting Tweets via Active Learning (AL)}
\label{section:method_active_learning}
The underlying assumption of the pooling-based approach is that there exist prior hate speech datasets on which text classification models can be trained to kickstart the pooling technique. However, this assumption does not always hold \cite{oard2004building}, especially with less-studied languages (e.g., Amharic, Armenian, etc.). Instead, AL \citep{settles2012active,cormack_autonomy_2015,rahman_ICTIR_2020}  
can be applied using the following steps. 

\noindent\textbf{I. Training a Machine Learning Model}. To initiate the active learning process, we need a machine learning model. For AL, we adopt logistic regression as a simple model and two alternative feature representations: i) simple TF-IDF \citep{salton1988term}; and ii) BERT embedding from  Twitter-roBERTa-base \citep{TwitterroBERTa}. 

\noindent\textbf{II. Defining Document Selection Criteria}. 
Given a trained model, we can then utilize the predicted posterior probability $p(y^i|x^i)$ of document $x^i$ being hateful to decide whether or not to annotate that document.
While uncertainty sampling is most commonly used with AL to select instances to annotate that are close to the decision hyperplane, prior work in AL for IR \cite{cormack_autonomy_2015} has shown empirically the benefit of prioritizing selection of examples predicted to come from the rare class when the data is highly skewed. This is because it is important to expose the learner to as many examples of the rare class as possible. In addition,  because the model will often be wrong, misses (and near-misses) will still provide ample exposure to ambiguous examples across classes. Because hate speech is quite rare, \citep{founta2018large}, 
we thus select those documents for annotation which are most likely to be hateful. 
Thus we employ {\em Continuous Active Learning} (CAL) \citep{cormack_autonomy_2015,rahman_ICTIR_2020}, which utilizes $p(y^i|x^i)$ via {\bf Eqn. \ref{equation: CAL}}.

\vspace{-0.5em}
\begin{equation}
\label{equation: CAL}
x^\star = \arg\!\max\limits_{i}\  p(hateful|x^i)
\end{equation} 
\vspace{-0.5em}

We also implement traditional uncertainty sampling \citep{lewis1994sequential}, referred to in prior IR work as {\em Simple Active Learning} (SAL) \citep{cormack_autonomy_2015},  which selects a document for human annotation when the classifier is most uncertain about whether or not a document is hateful. Typically an entropy-based uncertainty function \citep{settles2012active} is employed for SAL: 

\begin{equation}
Uncertainty(x)  = - \sum_{y \in Y} P(y|x) \log P(y|x) 
\end{equation}
where $y$ is either hateful or non-hateful. With this binary classification, SAL selects: 
\begin{equation}
\label{equation: SAL}
x^\star = \arg \!\min \limits_{i} \ |p(hateful|x^i)-0.5|
\end{equation}

\noindent\textbf{III. Seed Documents}. Some {\em seed documents} are necessary to induce the initial machine learning model. Prior work \citep{rahman_ICTIR_2020} has shown that a few seed documents can suffice to kickstart the AL process. Seed documents can be collected from  annotators in various ways: they might write hateful documents (e.g., tweets, posts) or search social media platforms for hateful documents. 

\begin{algorithm}
    
    \SetAlgoLined
    \SetKwInOut{Input}{Input}
    \SetKwInOut{Output}{Output}

    \Input{~~Document collection $X$ $\bullet$ batch size $u$ $\bullet$ total budget $b$}
    \Output{~~Hate Speech Dataset $R$}

        Select seed documents $S$     	\\
        $R \leftarrow \{\langle x^i, y^i \rangle\ |\ x^i \in S\}$ \Comment Collect initial judgments \\
    	Learn the initial machine learning model $c$ using $R$\\
    	$b \leftarrow b - |S|$ \Comment Update remaining budget \\
    \While {True}{
    	\algorithmicif\ $b < u$ \algorithmicthen\ \Return \Comment Budget exhausted \\
    	$\forall x\in X$ predict hatefulness of document $x$ using $c$\\
        Select $u$ documents $S \in X$ to judge next 
        \\
       	
        $R\leftarrow R\cup \{\langle x^i, y^i \rangle\ |\ x^i \in S\}$ \Comment Collect judgments\\
       	Re-train the machine learning model $c$ using expanded $R$\\
       	$b \leftarrow b - u$ \Comment Update remaining budget
    }
    
\caption{Active Learning Approach to Selecting Documents for Annotation}
\label{algorithm:AL_hate_speech}
\end{algorithm}

The active learning-based process for developing a hate speech dataset is described in {\bf Algorithm \ref{algorithm:AL_hate_speech}}. Given seed documents, an initial machine learning model is trained using those seed documents (Lines 1 - 4). The iterative process of AL is conducted in the loop (Lines 5 - 11), where a trained model selects documents for human annotation using the document selection criteria as discussed earlier (Lines 7 - 9).  The model is re-trained using both existing documents and those newly annotated documents (Line 10). This continues until the budget is exhausted. 

\section{The Dataset}
\label{section:Constructing Hate Speech Dataset}
We create our hate speech dataset via pooling (Section \ref{section:pooling_algorithm}). We use five models to detect hate speech: logistic regression, naive bayes, an LSTM-based model \cite{badjatiya2017deep}, a BiLSTM-based model \cite{agrawal2018deep}, and a model based on BERT \cite{devlin2018bert}. For both LSTM and BiLSTM models, we use the improved versions of these models reported by \citet{arango2019hate}. Models are trained across five prior hate speech datasets:  \citet{waseem-hovy-2016-hateful, davidson2017automated, grimminger2021hate, golbeck2017large, basile-etal-2019-semeval}. 
Models are trained for binary classification, with datasets binarized accordingly. 

{\bf User statistics.} \citet{arango2019hate} note that 65\% of hate speech annotated in \citet{waseem-hovy-2016-hateful}'s dataset comes from two Twitter users, suggesting coverage is not representative of online hate. In contrast, our dataset is constructed based on a random sample of tweets from the Twitter API, then selected for annotation via pooling. Out of 9,534 unique users included in our dataset, only 7 produce more than 2 tweets, with at most 15 tweets by one user and 28 total tweets across the other 6.

{\bf Annotation Process}. We collect the annotations in two phases: a set of iterative pilot experiments (4,668 labeled tweets), followed by consistent use of our final annotation design (4,999 labeled tweets). The pilot data revealed two key issues: i) many pornography-related tweets were annotated as hateful, and ii) sometimes annotators did not highlight both the language and targets of hate speech. We thus updated the annotation guidelines accordingly. Although we release the annotated labels collected from both of these phases, we only analyze and report results using the 4,999 annotations collected from the final annotation design. 

Since hate speech is highly subjective and complex, prior work \citep{sanguinetti2018italian, assimakopoulos2020annotating} has argued that simply asking the annotators to perform binary classification of hate vs. non-hate of an online post is unlikely to be reliable. Thus, a more complex, hierarchical annotation scheme may be needed. For example,  \citet{sanguinetti2018italian} break down the task of annotating hate speech into two sub-tasks. Based on this, we design an annotation scheme that is hierarchical in nature and decomposes the hate annotation task into a set of simpler sub-tasks corresponding to the definitional criteria that must be met for a post to constitute hate speech. Our annotation process is highly structured, including use of term definitions, annotation sub-tasks, types of hate (derogatory language or inciting violence), demographic categories, required rationales for labeling decisions), and a self-consistency test. 

\textbf{Dataset Properties}. 
\label{sec:props}
If an annotator identifies (implicit or explicit) language that is derogatory toward or incites violence against a demographic group, and identifies an (implicit or explicit) target group, we can infer from these annotations that a tweet is hateful.  In addition, we also ask the annotator (step 6) directly to judge whether the tweet is hateful.  This allows us to perform a valuable {\em self-consistency} check \cite{Zhang-sigir14} on the annotator, which is especially valuable for subjective tasks in which inter-annotator agreement is naturally lower.  We observe that $94.5$\% of the time, annotators provide final judgments that are self-consistent with their sub-task annotations of 1) hateful language and 2) demographic targets. For the other $5.5$\% of tweets (274), half of these are labeled as hateful but the annotator does not select either corresponding targets or actions of hate or both, and some annotators still mark pornographic content as hateful. We discard these 274 tweets, leaving 4,725 in our final dataset. 

Comparable to prior work  \citep{ousidhoum2019multilingual, ross2017measuring, kwok2013locate, del2017hate}, we observe
inter-annotator agreement of Fleiss $\kappa=0.16$ (raw agreement of 72\%) using hate or non-hate binary label of this dataset. Given task subjectivity, this is why self-consistency checks are so important for ensuring data quality. It is also noteworthy that the $\kappa$ score for inter-annotator agreement is not comparable across studies because $\kappa$ depends on the underlying class distribution of the annotators \citep{feinstein1990high}. In fact, $\kappa$ can be very low even though there is a high level of observed agreement \citep{uebersax1988validity}. There is in fact debate in the statistical community about how to calculate the expected agreement score \citep{feinstein1990high} while calculating $\kappa$.  \citet{gwet2002kappa} advocate for a more robust inter-annotator agreement score and propose $AC_1$, with interpretation similar to the $\kappa$ statistic. \citet{Jikeli2021Anti} report both Cohen $\kappa$ and Gwet's $AC_1$ in their hate speech dataset and find that the skewed distribution of prevalence of hate speech severely affects $\kappa$ but not Gwet’s $AC_1$. We observe $AC_1 = 0.58$ agreement in binary hate vs.\ non-hate labels. Regarding inter-annotator agreement statistics for rationales supporting labeling decisions, please see our supplementary material \cite{rahman21-neurips21-supplementary}.

{\bf Relative Coverage}. Since we do not use any ``hate words'' to filter tweets, our hate speech dataset includes hateful posts from both C1 and C2 categories (Figure \ref{Figure:landscape_of_hate_speech}). To quantify the coverage of hate in a dataset, we quantify the percentage of annotated hateful posts that do not contain known hate words as {\em Relative Coverage}: $100 \times (N_T - N_{TH})/N_{TH}$, 
where $N_T$ is the total number of hateful posts in the dataset, and $N_{TH}$ is the total number of hateful posts containing known hate words. For the purpose of analysis only, we use \citet{founta2018large}'s hate word list\footnote{The union of~ \url{https://www.hatebase.org} ~and~ \url{https://www.noswearing.com/dictionary}.}. Any hate speech dataset restricted to tweets having these hate words would have $N_T = N_{TH}$, yielding a relative coverage of 0\%, as shown in {\bf Table \ref{Table:data_relative_coverage}}. Because \citet{founta2018large}\footnote{\citet{founta2018large} report 80K tweets but their online dataset contains 100K. They confirm (personal communication) collecting another 20K after publication. The 92K we report reflects removal of 8K duplicates.} use tweets from both the keyword-based search and random sampling, the relative coverage of their dataset is 10.40\%, vs.\ our 14.60\%.    

\begin{table}
\caption{Comparison of Datasets. Pooling is seen to achieve the best balance of prevalence vs.\ relative coverage compared to the keyword-based search (KS) and random sampling (RS) methods.}
\vspace{-1em}
\begin{center}\scalebox{0.9}{
\begin{tabular}{ c r c r r}
\hline
 {\bf Dataset} & {\bf Size} & {\bf Method} & {\bf Prevalence} & {\bf Rel.\ Cov.} \\
 \hline
 G\&K \cite{grimminger2021hate} & 2,999 &  KS & 11.7\% & 0.0\%\\
    \hline
 W\&H \cite{waseem-hovy-2016-hateful} & 16,914 & KS & {\bf 31\%} & 0.0\%\\
 \hline

  G.\ et al.\ \cite{golbeck2017large} & 19,838 &  KS & 15\% & 0.0\%\\
    
\hline
 D.\ et al.\ \cite{davidson2017automated} & 24,783 & KS &  5.77\% & 0.0\% \\
 
 \hline
 F.\ et al.\ \cite{founta2018large} & 91,951 & KS and RS & 4.96\% & 10.40\%\\
 \hline

 {\bf Our approach} & 4,999 &  Pooling &  14.12\% &  {\bf 14.60}\%\\
   
 \hline
 \end{tabular}}
\end{center}
\label{Table:data_relative_coverage}
\vspace{-1em}
\end{table}

{\bf Prevalence}. 
Table \ref{Table:data_relative_coverage} 
shows that the prevalence of our hate speech dataset is 14.12\%, higher than the datasets created by \citet{founta2018large} where the authors cover a broad range of hate speech by combining random sampling with the keyword-based search. In addition to that, the prevalence of our hate speech dataset exceeds that for \citet{davidson2017automated} and is comparable to the datasets created by \citet{grimminger2021hate} and \citet{golbeck2017large}, where the authors employ keyword-based search. Again, prevalence can be inflated by only labeling tweets with known strong hate words. 

\section{Model Benchmarking}
We benchmark three recent hate speech detection models: LSTM-based \citep{badjatiya2017deep}, BiLSTM-based,  \citep{agrawal2018deep}, and a model based on BERT \citep{devlin2018bert}. For both LSTM and BiLSTM models, we do not use the original versions of these models, but rather the the corrected versions reported by \citet{arango2019hate}. 


{\bf Train and Test Sets}. To compare the models, we perform an 80/20 train/test split of our dataset. To maintain class ratios in this split, we apply stratified sampling. For analysis, we also partition test results by presence/absence of known ``hate words'', using \citet{founta2018large}'s hate word list. Train and test set splits contain 3,779 and  946 tweets, respectively. In the test set, the number of normal tweets without hate words and with hate words is 294 and 518 respectively, whereas the number of hateful tweets without hate words and with hate words is 15 and 119, respectively. 

{\bf Results and Discussion}. Performance of the models on the test sets in terms of precision (P), recall (R) and $F_1$ is shown in {\bf Table \ref{Table:Benchamark_results}}. Following \citet{arango2019hate}, we also report these three performance metrics for both hateful and non-hateful class (Table \ref{Table:Benchamark_results}, Column -  Class). We discuss the performance of the models considering the two types of test sets as described above. 

\begin{table}[h]
\caption{Hate classification accuracy of models (bottom 3 rows) provides further support for prior work \cite{arango2019hate,arango2020hate}'s assertion that ``hate speech detection is not as easy as we think.''}

\vspace{-1em}
\begin{center}\scalebox{0.93}{
\begin{tabular}{ l | c | c c c | c c c} 

\hline
       &       & \multicolumn{3}{c}{ \textbf{Without Hate Words}} & \multicolumn{3}{c}{\textbf{With Hate Words}} \\ \hline  
  
  {\bf Method} & {\bf Class} & {\bf P} & {\bf R} & {\bf F1} & {\bf P} & {\bf R} & {\bf F1} \\
  
  \hline
BiLSTM \citep{agrawal2018deep}  & Non-Hate & \cellcolor{blue!45}95.34 & \cellcolor{blue!45}97.61 & \cellcolor{blue!45}96.47 & \cellcolor{blue!35}84.98 & \cellcolor{blue!35}86.29 & \cellcolor{blue!35}85.63 \\
LSTM \citep{badjatiya2017deep} & Non-Hate & \cellcolor{blue!45}95.60 & \cellcolor{blue!45}96.25 & \cellcolor{blue!45}95.93 & \cellcolor{blue!35}84.95 & \cellcolor{blue!35}89.38 & \cellcolor{blue!35}87.11 \\
BERT \citep{devlin2018bert} & Non-Hate & \cellcolor{blue!45}96.25 & \cellcolor{blue!45}96.25 & \cellcolor{blue!45}96.25 & \cellcolor{blue!35}88.84 & \cellcolor{blue!35}83.01 & \cellcolor{blue!35}85.82 \\
\hline
\hline
BiLSTM \citep{agrawal2018deep}  & Hate & \cellcolor{blue!15}12.50 & \cellcolor{blue!15}6.666 & \cellcolor{blue!15}8.695 & \cellcolor{blue!25}36.03 & \cellcolor{blue!25}33.61 & \cellcolor{blue!25}34.78 \\
LSTM \citep{badjatiya2017deep} & Hate & \cellcolor{blue!15}15.38 & \cellcolor{blue!15}13.33 & \cellcolor{blue!15}14.28 & \cellcolor{blue!25}40.21 & \cellcolor{blue!25}31.09 & \cellcolor{blue!25}35.07 \\
BERT \citep{devlin2018bert} & Hate & \cellcolor{blue!15}26.66 & \cellcolor{blue!15}26.66 & \cellcolor{blue!15}26.66 & \cellcolor{blue!25}42.48 & \cellcolor{blue!25}54.62 & \cellcolor{blue!25}47.79 \\
\hline

 \end{tabular}}
\label{Table:Benchamark_results}
\end{center}

\end{table}

{\bf Case  I. Without Hate Words}. Intuition is that correctly classifying a document as hateful when canonical hate words are absent is comparatively difficult for the models. This is because models typically learn training weights on the predictive features, and for hate classification, those hate words are the most vital predictive features which are absent in this setting. By observing $F_1$, we can find that for the hateful class, all these models, including BERT, provide below-average performance ($F_1$  $\leq$ 26.66\% ). In other words, the number of false negatives is very high in this category. On the other hand, when there are no hate words, these same models provide very high performance on the non-hateful class, with $F_1$ $\approx$ 96\%. 

{\bf Case II. With Hate Words}. In this case, hate words exist in documents, and models are more effective than the previous case in their predictive performance (Table \ref{Table:Benchamark_results}, Columns 6, 7, and 8) on the hateful class. For example, $F_1$ of BERT on the hateful class has improved from 26.66\% (Column 3) to 47.79\% (Column 8). This further shows the relative importance of hate words as predictive features for these models. On the other hand, for classifying documents as non-hateful when there are hate words in documents, the performance of models are comparatively low (average $F_1$ across models is $\approx$ 86\%) vs.\ when hate words are absent (average $F_1$ across models is $\approx$ 96\%). 

\textbf{Table \ref{Table:Benchamark_results}} shows clearly that models for hate speech detection struggle to correctly classify documents as hateful when ``hate words'' are absent. While prior work by \citet{arango2019hate} already confirms that there is an issue of overestimation of the performance of current state-of-the-art approaches for hate speech detection, our experimental analysis suggests that models need further help to cover both categories of hate speech (C1 and C2 from Figure \ref{Figure:landscape_of_hate_speech}).

\section{Active Learning vs.\ Pooling}
\label{section:active_learning}

While pooling alone is used to select tweets inclusion and annotation in our dataset, we also report a retrospective evaluation of pooling vs.\ active learning (AL). In this retrospective evaluation, AL is restricted to the set of documents selected by pooling, rather than the original full, random sample of Twitter from which the pooling dataset is derived. Because of this constraint, the retrospective AL results we report here are likely lower than what might be achieved when AL run instead on the full dataset, given the larger set of documents that would be available to choose from for annotation.


\noindent{\bf Experimental Setup}. Each iteration of AL selects one document to be judged next (i.e., batch size $u$ = $1$). The total allotted budget $b$ is set to the size of the hate speech dataset constructed by pooling ($b=4,725$). For the document selection, we report SAL and CAL. As a baseline, we also report a random document selection strategy akin to how traditional supervised learning is performed on a presumably random sample of annotated data. Following prior work's nomenclature, we refer to this as {\em simple passive learning} (SPL) \cite{cormack_autonomy_2015}. As seed documents, we randomly select 5 hateful and 5 non-hateful documents from our hate speech dataset constructed using pooling. 
 
\begin{figure*}[t]
\centerline{\includegraphics[width=1.0\textwidth]{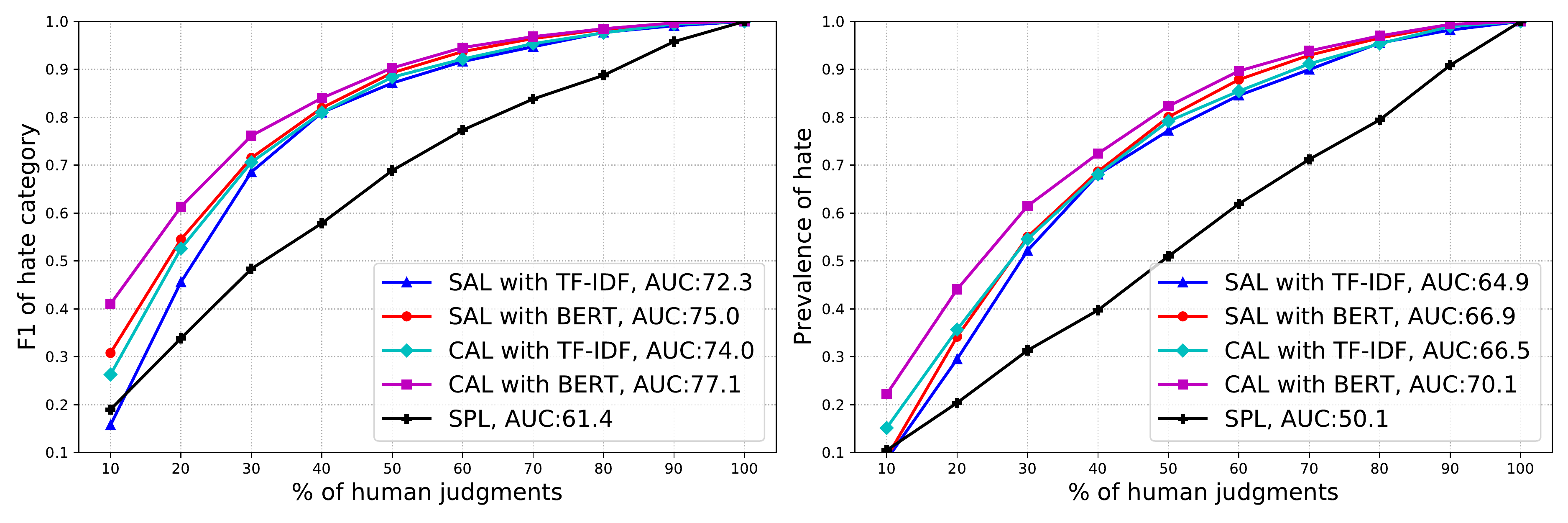}} 
\caption{Left plot shows human judging cost (x-axis) vs. F1 classification accuracy (y-axis) for hybrid human-machine judging. Right plot shows human judging cost vs. prevalence of hate speech for human-only judging of documents. The \% of human judgments on x-axis is wrt.\ the number documents in the hate speech data constructed by the pooling-based approach.}
\vspace{-1em}
\label{Figure:AL_performance}
\end{figure*}

\noindent{\bf Experimental Analysis}. We present the results ({\bf Figure \ref{Figure:AL_performance}}) as plots showing the cost vs.\ effectiveness of each method being evaluated at different cost points  (corresponding to varying evaluation budget sizes). We also report Area Under Curve (AUC) across all cost points, approximated via the Trapezoid rule. To report the effectiveness, Figure \ref{Figure:AL_performance} presents $F_1$ performance (left plot), and prevalence (right plot) results of the three document selection approaches: SPL, SAL, and CAL, along with two feature representations: TF-IDF and contextual embedding from BERT. 

While reporting $F_1$ of our AL classifier (Figure \ref{Figure:AL_performance}, left plot), following prior work \citep{nguyen2015combining}, we consider both human judgments and machine predictions to compute $F_1$. For example, when we collect 20\% human judgments from annotators, the remaining 80\% is predicted by the AL classifier (i.e., machine prediction). Then those two sets (human judgments and machine prediction) are combined to report the final $F_1$ score. However, to report the prevalence of hate (right plot), only human judgments are considered because we need the actual labels of documents, not the machine predictions here. 

{\bf Active vs.\ Passive Learning}. By comparing active learning (SAL and CAL) methods against passive learning (SPL) method in terms of $F_1$, we find that SAL and CAL significantly outperform SPL in terms of AUC. The observation holds for both types of feature representations of documents. The only exception we can see that at 10\% human judgments, $F_1$ of SPL is slightly better than SAL with TF-IDF. Nevertheless, after that point of 10\% human judgments, SAL with TF-IDF consistently outperforms  SPL in terms of $F_1$. Similarly, if we consider the prevalence of hate (right plot), active learning exceeds passive learning by a large margin in AUC.     

{\bf CAL vs. SAL}. From Figure \ref{Figure:AL_performance}, we can see that CAL consistently provides better performance than SAL both in terms of $F_1$ and prevalence. In fact, when the allotted budget is very low (e.g., budget $\le$ 20\% of human judgments), the performance difference in terms of $F_1$ and prevalence between CAL and SAL considering the underlying feature representation is very high. 

{\bf TF-IDF vs. BERT embedding}. It is evident from Figure \ref{Figure:AL_performance} that the  contextual embedding from BERT for document representation provides a significant performance boost over the TF-IDF-based representation. SAL and CAL both achieve better performance when the document is represented using BERT. Furthermore, we can find that for a low-budget situation (e.g., budget $\leq$  20\% of human judgments), CAL with BERT provides the best performance.  

{\bf Judging cost vs. Performance} We also investigate whether the AL-based approach can provide better prevalence and classification accuracy at a lower cost than the pooling-based approach. From Figure \ref{Figure:AL_performance}, we can find that the AL-based approach achieves $F_1 \ge 0.9$ (left plot) and finds 80\% of hateful documents (right plot) by only annotating 50\% of the original documents judged via pooling.

\section{Conclusion \& Future Work}
The success of an automated hate speech detection model largely depends on the quality of the underlying dataset upon which that model is trained on. However, keyword-based filtering approaches \citep{davidson2017automated, waseem-hovy-2016-hateful, grimminger2021hate} do not provide broad coverage of hate, and random sampling-based approaches \citep{kennedy2018gab, de-gibert-etal-2018-hate} suffer from low prevalence of hate. We propose an approach that adapts {\it pooling} from IR \citep{sanderson2010test} by  training multiple classifiers on prior hate speech datasets and combining that with random sampling to improve both the prevalence and the coverage of hate speech in the final constructed dataset. 

Using the pooling technique, we share a new benchmark dataset for hate speech detection on Twitter. Results show that the hate speech dataset developed by applying the proposed pooling-based method achieves better relative coverage (14.60\%) than the hate speech dataset constructed by \citet{founta2018large}  that combines random sampling with the keyword-based search (Table \ref{Table:data_relative_coverage}). Furthermore, the prevalence of hate-related content in our hate speech dataset is comparable to many prior keyword-based approaches \citep{davidson2017automated, grimminger2021hate, golbeck2017large}. We also show a dramatic drop in accuracy of existing detection models \citep{devlin2018bert, badjatiya2017deep, agrawal2018deep} when tested on these broader forms of hate. 

An important limitation of the pooling approach is that it relies on prior hate speech datasets. That said, though the datasets used to train our pooling prediction models may lack diverse tweets, the trained models still learn correlations between hate labels and all vocabulary in the dataset. This enables pooling to identify some hateful tweets (for inclusion in our dataset) that lack known hate words, though this prediction task is clearly challenging (as the bottom left of Table \ref{Table:Benchamark_results} indicates). In general, the purpose of pooling in IR is that diverse models better identify potentially relevant content for human annotation. In the translational use of pooling we present for hate speech, training a cross product of different models across different prior datasets similarly promotes diversity and ``wisdom of crowds'' in identifying potentially hateful content for inclusion and annotation.

To sidestep this reliance on prior hate speech datasets for pooling, we also present an alternative approach that utilizes active learning \citep{settles2012active} to develop the hate speech dataset. Empirical analysis on the hate speech dataset constructed via the pooling-based approach suggests that by only judging 50\% of the originally annotated documents, it is possible to  find 80\% of hateful documents with an $F_1$ accuracy of $\ge 0.9$ (Figure \ref{Figure:AL_performance}) via our AL-based approach. 

Like HateXplain \cite{mathew2020hatexplain}, our collection of rationales as well as labels creates the potential for explainable modeling \cite{strout2019human} and dual-supervision \cite{zhang2016rationale}. However, while HateXplain collects rationales only for the overall labeling decision, our collection of rationales for different annotation sub-tasks creates intriguing possibilities for dual-supervision \citep{xia2017dual} and explanations across different types of evidence contributing to the overall labeling decision \citep{lehman2019inferring}. In particular, annotator rationales identify i) derogatory language, ii) language inciting violence, and iii) the targeted demographic group. As future work, we plan to design dual-supervised \citep{xia2017dual} and/or explainable \citep{deyoung2019eraser} machine learning models that can incorporate the annotators' rationales collected in our hate speech dataset.   

Our definition of hate speech and annotation process assumed that hate speech is composed of two parts: 1) language that is derogatory or inciting violence against 2) a target demographic group. However, an interesting question is whether perceptions of hate differ based on the demographic group in question, e.g., a given derogatory expression toward a political group might be deemed acceptable while the same expression to a racial or ethnic group might be construed as hate speech. HateCheck \cite{rottger-etal-2021-hatecheck}'s differentiation of general templates of hateful language vs. template instantiations for specific demographic targets could provide a nice framework to further investigate this.

Although our annotation  process was highly structured, we ultimately still produced binary labels for hate speech, lacking nuance of finer-grained ordinal scales or categories. It would be interesting to further explore our structured annotation process with such finer-grained scales or categories.

Recent years have brought greater awareness that machine learning datasets (as well as models) can cause harm as well as good \citep{shankar2017no, buolamwini2018gender, noble2018algorithms, eubanksautomating}. For example, harm could come from deploying a machine learning model without considering, mitigating, and/or documenting \cite{gebru2018datasheets, bender2018data, boyd2021datasheets} limitations of its underlying training data, 
such as the risk of racial bias in hate speech annotations \cite{sap2019risk,wich2020impact,davidson2019racial}. 
Both researchers and practitioners of hate speech detection should be well-informed about such potential limitations and risks of any constructed hate speech dataset (including ours) and exercise care and good judgment while deploying a hate speech detection system trained on such a dataset. 

We know of no prior work studying the effect of frequent exposure to hate speech on the well-being of human annotators. However, prior evidence suggests that exposure to online abuse has serious consequences on the mental health of  workers \citep{ybarra2006examining}. Studies by 
\citet{boeckmann2002hate} and \citet{ leets2002experiencing} to understand how people experience hate speech found that low self-esteem, symptoms of trauma exposure, etc., are associated with repeated exposure. The Linguistic Data Consortium (LDC) reported its annotators experiencing nightmares and other overwhelming feelings from labeling news articles \cite{strassel2000quality}. We suggest more attention be directed toward the well-being of the annotators \cite{steiger2021psychological}.

\textbf{Acknowledgements}. We thank the many talented Amazon Mechanical Turk workers who contributed to our study and made it possible. This research was supported in part by Wipro (HELIOS), the Knight Foundation, the Micron Foundation, and Good Systems (\url{https://goodsystems.utexas.edu}), 
a UT Austin Grand Challenge to develop responsible AI technologies. Our opinions are our own.


\bibliographystyle{ACM-Reference-Format}
\bibliography{main_and_supp}

\appendix
\section{Related Work}
\label{section:related_work}
\vspace{-.25em}

A considerable amount of research \citep{waseem-hovy-2016-hateful, waseem2016automatic, davidson2017automated, golbeck2017large, grimminger2021hate, founta2018large, djuric2015hate, de-gibert-etal-2018-hate, kennedy2018gab} has been conducted to construct datasets for hate speech. These approaches can be mainly categorized into three groups: i) keyword-based search \citep{waseem-hovy-2016-hateful, waseem2016automatic, davidson2017automated, golbeck2017large, grimminger2021hate}; ii) random sampling \citep{de-gibert-etal-2018-hate, kennedy2018gab}; and 
iii) random sampling with the keyword-based search  \citep{founta2018large, djuric2015hate}. 


\textbf{Keyword-based search}. 
Because less than 3\% of tweets are hateful \citep{founta2018large}, prior studies for constructing hate speech are  largely based on the keyword-based search. In particular,
a set of manually curated ``hate words'' are defined and documents containing any of these keywords are selected for annotation. 
While defining this list of keywords, prior work often considers hate words that are typically used to spread hatred towards various targeted groups. For example, \citet{waseem-hovy-2016-hateful} identify 17 different keywords as hate words
which covers hate under racism and sexism categories. \citeauthor{waseem-hovy-2016-hateful} find 130K tweets containing those hate words 
and annotate 16,914 of them 
as racist, sexist, or neither.

\citet{golbeck2017large} also define their own set of 10 hate words
which are used to cover racism,
Islamophobia, homophobia, anti-semitism, and
sexism related hate speech. 
Similarly, \citet{warner2012detecting} also construct a hate speech dataset containing 9,000
human-labeled documents from Yahoo News! and {\em American Jewish Congress} where they consider hate words targeting Judaism and Israel. 
Apart from using keywords targeted towards specific groups, 
prior work also utilizes general hate words. For instance, \citet{davidson2017automated}  utilize keywords from the  HateBase \cite{hatebase}, a crowd-sourced list of hate words, whereas \citet{founta2018large} use keywords from both the HateBase and an offensive words dictionary \cite{noswearing}. 

Prior work \citep{kwok2013locate, hosseinmardi2015detection} on hate speech dataset construction also focused on profiles of social media users who are known to generate hateful content. 
For example, \citet{kwok2013locate} search for keywords from Twitter users who claim themselves as racist or are deemed racist based on the news sources they follow. 

\textbf{Random Sampling}.
\citet{kennedy2018gab} argue that the datasets constructed by the keyword-based search are biased towards those keywords and therefore are not representative of the real-world. Thus the authors randomly sample 28,000 Gab (\url {gab.com}) posts for annotation. 
Similarly, \citet{de-gibert-etal-2018-hate} collect documents for annotation from a White Supremacist forum (\url{https://www.stormfront.org}) by selecting documents uniformly at random.

\textbf{Random Sampling with Keyword Search}. 
Prior work has also combined keyword-based search and random sampling to select posts for annotation. For example, \citet{founta2018large} develop a hate speech dataset that contains 91,951 annotated tweets categorized into four categories: abusive, hateful, spam, and normal. While many annotated tweets are randomly sampled from the Twitter API, they also select some tweets for annotation based on the keyword-based search to increase prevalence. 

Similarly, \citet{wulczyn2017ex} develop a dataset of personal attacks from Wikipedia comments that contains 37,000 randomly sampled
comments and 78,000 comments from users who are blocked. The authors mention that since the prevalence of personal attacks on those 37,000 randomly sampled comments is only 0.9\%, they increase the prevalence of personal attacks comments by searching over the blocked users' comments.    

We do not use the keyword-based search method in our work and do not apply only random sampling. Instead, we adapt the well-known pooling method \cite{sanderson2010test} for constructing IR test collections to select documents to be annotated. To our knowledge, this is the first work using pooling method to construct a hate-speech dataset.

\section{Hate Speech Challenges}
\label{section:issues_hate_speech}

There are various challenges associated with developing a dataset for hate speech, and in this section, we will discuss those challenges. However, we should note that most of these challenges are widely debated issues in machine learning research including dataset bias \citep{tommasi2017deeper}, annotator bias \citep{geva-etal-2019-modeling}, documenting datasets \citep{gebru2018datasheets}, task decomposition \citep{JiangEfficient2014}, selection of annotators \citep{aroyo2015truth}  and others. If adequate steps are not taken to mitigate various issues associated with the challenges,  datasets will reflect various forms of biases. Consequently, researchers and practitioners who deprioritize the biases in the dataset would run the risk of inflicting  greater harm to human society by deploying automated systems trained on these biased datasets. 

\subsection{Definition of Hate Speech}
Even experts disagree on what constitutes hate speech \citep{founta2018large, waseem-hovy-2016-hateful, davidson2017automated, sanguinetti2018italian}. It is a complex phenomenon typically associated with relationships between groups and depends on the nuances of languages. Since there is no legal definition of hate speech, various international organizations, social media platforms, and research articles \citep{davidson2017automated, waseem-hovy-2016-hateful} define hate speech  differently. There are two notable similarities between these definitions: 1) hate speech incites violence or is intended to be derogatory, and 2) hate speech is directed towards certain targeted groups. 

However, these definitions are not comprehensive enough to cover the real-world representation of hate speech. For example, \citet{macavaney2019hate} point out these definitions cover whether someone is attacked or humiliated in hate speech. However, praising a particular group (e.g., KKK, Nazi) may also be considered hate speech and this is not covered by the existing definitions.

\subsection{Annotation Schema}
Hate speech is a relatively complex phenomenon because the difference between other related concepts (e.g., cyberbullying \citep{chen2011detecting}, abusive language \citep{nobata2016abusive}, discrimination \citep{thompson2016anti}, etc.) and hate speech is not  obvious \citep{fortuna2018survey}. As a result, different hate speech datasets have different annotation schema for hate speech and other related concepts \citep{davidson2017automated, founta2018large}. 

The binary annotation schema is the basic schema that labels a post as either hate speech or normal speech. However, prior studies mostly annotate hate speech using non-binary schema.  Since offensive language is prevalent in social media and does not necessarily always represent hate speech, \citet{davidson2017automated} annotate their dataset using three categories: i) hate speech, ii) offensive language, and iii) normal speech. \citet{mathew2020hatexplain} also follow these three categories in annotation. Apart from hate speech, offensive language, and normal speech categories,  \citet{founta2018large} annotate their dataset into four (4) other categories, namely: i) abusive language, ii) aggressive behavior, iii) cyberbullying, and iv) spam. 

Non-binary schemes based on the intensity of hate speech are also utilized in prior studies. For example, \citet{del2017hate} implement strong hate,
weak hate, and no hate, \citet{kumar-etal-2018-benchmarking} categorize posts into overtly aggressive, covertly aggressive, not aggressive categories. \citet{poletto2017hate} compare binary annotation scheme against rating scale and best-worst ranking scale for hate speech annotation and find that rating scale is comparatively better than the other two schemes. 


\subsection{Annotation Guidelines}
Often, it is very challenging for the annotators to decide whether a particular post or document is hateful or not \citep{ross2017measuring, sanguinetti2018italian}. Thus a carefully designed annotation guideline is crucial to have a better quality hate speech dataset. However, prior work also significantly differs from each other in terms of designing annotation guidelines. Most of the time, researchers only specify the definition of categories (e.g., hate or offensive) \citep{davidson2017automated, founta2018large} but do not provide any additional clarification about how to interpret each of those categories.   

Furthermore, since annotators are not provided with any contextual information regarding the social media post, different authors provide different types of guidelines to their annotators to resolve this absence of context. For example, \citet{davidson2017automated} instructed the annotators not only to consider the presented tweets but also to think about the context in which tweets might appear before making the judgment. 
However, such practices risk making the task of annotating hate speech more subjective.
 
\subsection{Selection of Annotators} 
Previous studies regarding hate speech also vary in terms of hiring annotators. Given the nuances of language and the degree of difficulty of annotating hate speech,  expert annotators can play an important role in achieving a higher inter-annotator agreement  \citep{gao2017detecting}. Prior work hired experts from different backgrounds including feminists and anti-racism activists \citep{waseem-hovy-2016-hateful}, content moderators \citep{pavlopoulos2017deep}, PhD students in Linguistics \citep{kumar2018aggression}, experts in Natural Language Processing \citep{mathur2018did}. In addition, since expert annotators typically have domain knowledge, it is expected that expert annotators tends to agree more with other experts in annotating hate speech. For example, \citet{waseem2016you} find that crowd-workers have a lower inter-annotator agreement score than experts.

However, hiring expert annotators is expensive, and there is also a scalability issue. Thus for a large-scale annotation task, prior work typically employs crowd-workers  \citep{davidson2017automated, founta2018large}. To make sure that crowd-workers have the necessary expertise to perform the annotation task, prior work sometimes restrict annotation tasks to workers meeting certain qualifications (e.g., Amazon Mechanical Turk). However, crowd-workers lacking proper training are more prone to do the ``keyword-spotting'' while labeling hate speech. As a result,  crowd-workers may be more likely to label a post as hate speech than experts \citep{waseem2016you}.

\subsection{Annotators' Bias}
Prior work also investigates how annotators' demographics (e.g., gender, race, first language) affect the perception of the annotators to hate speech. \citet{gold2018women} find that female annotators who are typically part of the targeted group in hate speech are more likely to annotate a possible gender-related post as sexist than their male counterparts (i.e., {\it gender bias}). A similar type of observation (i.e., {\it racial bias}) is also made by \citet{kwok2013locate} while working on racist hate speech.  Additionally, \citet{sap2019risk} report that annotators who are unfamiliar with the African American English (AAE) dialect  are more likely to label documents containing AAE as racist, although those same documents may be considered non-racist by native AAE speakers. 

The {\it racial bias} problem is even more severe when we consider the crowd-workers; for example, in Amazon Mechanical Turk, non-AAE speakers are overrepresented \citep{hitlin2016research}. Additionally, the annotators' political ideology can also unintentionally manifest in yielding a politically biased dataset \citep{wich2020impact}.  Furthermore, it has been found that both expert and crowd-workers are prone to similar types of bias while annotating hate speech \citep{davidson2019racial}. Given this counter-intuitive observation, prior work \citep{davidson2019racial} argues to develop a better training process for the annotators to mitigate the annotators' bias.     

\subsection{Measurement of Annotator Agreement} 
Previous studies diverge significantly in reporting on the quality of the annotations, especially the inter-annotator agreement score \citep{fortuna2018survey}. Typically, Cohen's $\kappa$, Fleiss $\kappa$, Krippendorf’s $\alpha$, or a plain observed agreement percentage are reported in prior work. On the other hand, there are also many studies in hate speech that do not report any inter-annotator  agreement score \citep{fortuna2018survey}.   

Since many factors are involved in  annotation (e.g., annotation scheme, annotation guidelines, annotators' background), the reported agreement scores among prior studies vary widely. For example, \citet{bohra-etal-2018-dataset} report a Cohen $\kappa$ score of 0.982, whereas \citet{del2017hate} report a Fleiss $\kappa$ score of 0.19. Furthermore, different studies argue for different thresholds for an acceptable inter-annotator agreement score \citep{artstein2008inter,eugenio2004kappa}. Typically more complex annotation schemes \citep{del2017hate, sanguinetti2018italian} produce a lower inter-annotator agreement score than a simple, binary annotation scheme \citep{bohra-etal-2018-dataset, davidson2017automated}. 

We are not familiar with any prior work on hate speech annotation using self-consistency checks \cite{Zhang-sigir14}, which we believe complements traditional use of annotator agreement measures. Conceptually, for objective tasks with a single true answer, we expect reasonable annotator agreement, while on more subjective tasks \cite{Nguyen16-hcomp} (e.g., favorite ice cream flavor) we do not expect annotators to agree.  While an annotator can be expected to be self-consistent for either task type, self-consistency seems particularly valuable for subjective tasks when annotators are expected to disagree with one another. Hate speech annotation lies in the spectrum between objective vs. subjective tasks. Some objectivity is necessary to yield consistent data for training detection models, but low annotator agreement remains common. This is why we believe self-consistency measures can complement traditional practice.


\subsection{Absence of a Benchmark Hate Speech Dataset} 

Although hate speech is a widely discussed topic and there are many publicly available hate speech datasets, there is no commonly accepted benchmark dataset for hate speech detection  \citep{schmidt-wiegand-2017-survey, poletto2021resources, madukwe2020data}. This is largely due to the fact that in hate speech, data degradation is a known issue  \citep{Watanabe, chaudhry20-arxiv}. This is because researchers primarily collect hate speech from social media and release only the IDs of the social media posts in the hate speech domain. For example,  \citet{Watanabe} and \citet{chaudhry20-arxiv} report that a number of tweets released initially by \cite{waseem-hovy-2016-hateful} are not available anymore.  Furthermore, standard benchmark datasets (e.g., SQUAD \citep{rajpurkar2016squad}, MSMARCO \citep{nguyen2016ms}) provide a standard train-test-validation split, whereas most of the hate speech datasets lack in providing this train-test-validation split \citep{madukwe2020data}.  


\subsection{Less Generalizability of Automated Hate Speech Detection Models} 

Although the generalization capability of a model can be largely attributed to the complexity of the model itself, \citet{grondahl2018all} argue that for hate speech, the nature and composition of the datasets are more important than the model itself. This is because researchers differ from each other regarding various related issues of annotating hate speech, including definition, categories, annotation guidelines, types of annotators, aggregation of annotations. Consequently, different hate speech datasets have different natures and compositions. As a result, automated hate speech detection systems trained on one hate speech dataset exhibit poor generalization performance on another hate speech dataset. For example, \citet{arango2019hate} show that the state-of-the-art hate speech detection models \citep{badjatiya2017deep, agrawal2018deep} provide very poor cross-data generalization performance when trained on the dataset created by \citet{waseem-hovy-2016-hateful} but tested on the HateEval dataset \citep{basile-etal-2019-semeval}.


\subsection{Summary of Challenges}
In conclusion, all these issues discussed in this section 
should provide the practitioners a general overview about why they should be vigilant in performing their due diligence while deploying automated hate speech detection systems using any constructed hate speech dataset, including our own. While solving all these issues is beyond the scope of this work,  here, we particularly focus on the data sampling process (i.e., which post to select for annotation) so that the final hate speech dataset has a better coverage of hate speech from all categories (C1 and C2 of Figure 1) 
with a limited budget for annotation. Prior work regarding the data sampling process of hate speech is discussed in the next section.

\section{Tweet Corpus Collection}
\label{section_document_collection_phase}
We construct a collection of documents (e.g., tweets) by collecting a random sample of tweets from the Twitter Public API, 
which usually provides 1\% random sample of the entire Twitter stream in a given time range. In our case, we have collected tweets from May-2017 to Jun-2017. 

Next, we apply regular expressions to get rid of tweets containing retweets, URLs, or short videos. Tweets are also anonymized by removing the @username tag. However, we do not remove any emojis from tweets as those might be useful for annotating hate speech. We filter out any tweets as non-English unless two separate automated language detection tools, Python Langdetect\footnote{https://pypi.org/project/langdetect/} and Python Langid\footnote{https://github.com/saffsd/langid.py}, both classify the Tweet as English. 
Finally, after removing  duplicate tweets, approximately 13.6 million English tweets remain as our tweet corpus  collection.

\section{Document Annotation Process}
\label{section:document_annotation_process}

\subsection{Annotation Guidelines}
Our designed annotation guidelines consist of a clear definition  of what constitutes as hate speech and  examples covering various cases of hate speech. Following \citet{davidson2017automated}, we also instruct the annotators not to annotate any post as hateful if the derogatory language used in the post does not have any target associated with a protected group. Furthermore, annotators are explicitly instructed that they should not label any pornographic content as hateful. 

\subsection{Annotation Interface}
\label{section:interface}
Instructions ask annotators to follow these steps in order:
\begin{enumerate}

\item Highlight any words or phrases in the post ~{\bf INCITING VIOLENCE}.

\item Highlight any {\bf DEROGATORY LANGUAGE} in the post on the basis of group identity.

\item If the post {\bf IMPLICITLY} incites violence or denigrates an individual or group on the basis of group identity, select that option. \textbf{[INCITING VIOLENCE / DEROGATORY LANGUAGE]}

\item If the target is {\bf EXPLICIT}, highlight the INTENDED TARGET in the post. If the target is implicit, {\bf name the target}.

\item Identify the type of group targeted (explicit or implicit). \textbf{[BODY / GENDER /  IDEOLOGY / RACE / RELIGION / SEXUAL ORIENTATION / OTHER]}

\item Based on your answers to the above steps, do you believe the post is hateful? \textbf{[YES / NO]}

\item We welcome any additional explanation of your labeling decisions you would like to provide. \textbf{[TEXTBOX INPUT]} 

\end{enumerate}

This annotation scheme requests the annotators to identify both the violating content (Steps 1-2) and the demographic group targeted (Steps 3-5). When the annotators reach Step 6, they have already completed several sub-tasks. They then decide whether they believe the post is hateful or not. 


{\bf Targeted group identification}. Once  annotators identify the target of hate (implicit or explicit), they select the group identity of the target. There are seven (7) categories of targeted groups in our interface, as listed in the interface (Step 5).

{\bf Categorization of highlighted terms}. While \citet{mathew2020hatexplain} ask the annotators to only highlight terms that are related to hate speech, in our  interface, annotators have to do both highlighting  and categorization of terms that are related to the actions and the targets of hate speech. For example, for this post ``Good morning Kanye. Shut the fuck up", the annotators have to highlight ``Kanye" as the target and they also have to highlight terms ``Shut'' and ``fuck'' and categorize those terms as derogatory terms. In addition to the potential downstream value of the collected rationales, 
it is known from prior work \citep{mcdonnell16-hcomp,Kutlu18-sigir} that requiring annotator rationales improves label quality, even if the rationales are ignored. 


\subsection{Collecting Annotations} 
We hire annotators from Amazon Mechanical Turk. To ensure label quality, only annotators with at least 5,000 approved HITs and a 95\% HIT approval rate are allowed. We pay \$0.16 USD per tweet. 
Since three annotators annotate each tweet, including the platform fees, we have paid  \$4,930.17 USD in total to annotate  9,667 tweets. We apply majority voting to compute the final label. 

Guidelines indicate that if a tweet is hateful, annotators must identify both the targets and the actions related to hate speech; otherwise, their work will be rejected. Furthermore, typically only a few words or phrases are related to targets and actions of hate speech; highlighting all words will yield rejection. 
One might consider this rule as too prohibitive, because highlighting all words might be necessary in some cases. However, 
in our pilot study we observed that annotations in which entire text is highlighted corresponded to low quality work in  almost all cases.

To facilitate quality checks, we collect the annotations in iterative small batches. The quality check typically includes randomly sampling some annotated tweets and checking the annotations. Finally, if we reject any HIT, we notify the worker why we have done so and re-assign the task to others. 

\section{State-of-the-art Models}
\label{appendix-State-of-the-art Models}

{\bf I. LSTM}. The LSTM model implemented by \citet{badjatiya2017deep} achieves 93\% $F_1$ on the hate speech dataset created by \citet{waseem-hovy-2016-hateful} (though it is unclear whether they report macro or micro $F_1$). The deep learning architecture starts with an embedding layer with dimension size of 200.  Then it is followed by  a Long Short-Term
Memory (LSTM) network. Their final layer is a fully connected layer with a soft-max activation function to produce probabilities across three classes, namely sexist, racist, and non-hateful, at the output layer. To train the model, they use the categorical cross-entropy as a loss function and the Adam optimizer. In our case, we modify the output layer with two nodes and use Sigmoid as an activation function as we have a binary classification task. Finally, for the loss function, we use the binary cross-entropy loss function. The model is trained for ten epochs following \citet{badjatiya2017deep}.

{\bf II. BiLSTM}. \citet{agrawal2018deep} design a BiLSTM architecture that achieves $\approx$ 94\% score in terms of both micro and macro averaged $F_1$ on the hate speech dataset constructed by \citet{waseem-hovy-2016-hateful}. Their  architecture consists of the following layers sequentially: 1) Embedding layer, 2) BiLSTM Layer, 3) Fully Connected layer, and 4) output layer with three nodes. They also use the softmax activation function for the final layer and the categorical cross-entropy as loss function with the Adam optimizer. For this BiLSTM model, we also perform the same modification as we do for the LSTM model. In addition, we train the model for 30 epochs. 

{\bf Note.} For both LSTM and BiLSTM models, we use the corrected versions of these models reported by \citet{arango2019hate}. 

{\bf III. BERT}. We also utilize Bidirectional
Encoder Representations from Transformers (BERT) \citep{devlin2018bert} which achieves 67.4\% macro averaged $F_1$ on the hate speech dataset created by \citet{mathew2020hatexplain}. With pooling, we use the BERT-base-uncased model with 12 layers, 768 hidden dimensions, 12 attention heads, and 110M parameters. For fine-tuning BERT, we apply a fully connected layer with the output corresponding to the CLS token. The BERT model is fine-tuned for five epochs. 

{\bf Document Pre-processing}. We pre-process tweets using  {\tt tweet-preprocessor}\footnote{ https://pypi.org/project/tweet-preprocessor/}. Then we tokenize, and normalize those pre-processed tweets.  For the TF-IDF representation, we further stem those tweets  using Porter Stemmer \citep{willett2006porter}. 
 
{\bf Document Representation}. For LSTM and BiLSTM models, documents are represented using a word embedding where the embedding layer is initialized using the Twitter pre-trained GloVe embedding \citep{pennington2014glove} which is pre-trained on 2 billion tweets. For the Logistic Regression and Naive Bayes models, we generate the TF-IDF representation \citep{salton1988term} of documents using bigram, unigram, and trigram features following the work of \citet{davidson2017automated}.

\section{Additional Dataset Properties}
\label{section:additional_analysis}
\begin{figure}[t]
\centerline{\includegraphics[width=0.5\textwidth]{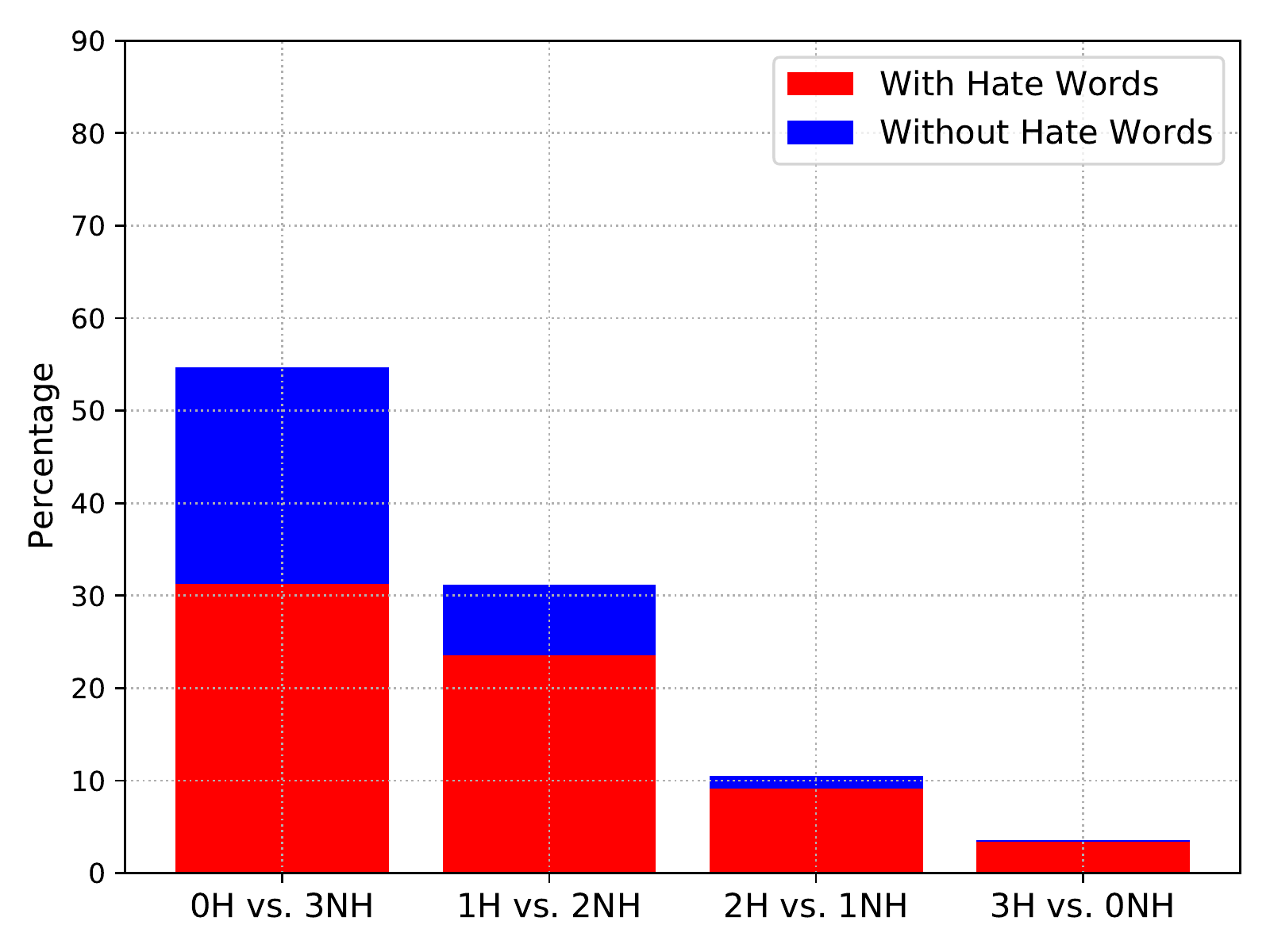}} 
\caption{Distribution of annotators' agreement}
\label{Figure:distribution_annatators_aggrement}
\vspace{-0.5cm}
\end{figure}

{\bf Effect of the presence of hate words on annotation.} We also analyze how the presence of hate words affects the decision-making process of the annotators. To achieve that, we plot the frequency distribution of the number of annotators who agree regarding the label of tweets in {\bf Figure \ref{Figure:distribution_annatators_aggrement}}. For example, the 1H vs. 2NH entry on the x-axis of Figure \ref{Figure:distribution_annatators_aggrement} represents  how many times one annotator labels a tweet as hateful, but two annotators annotate that tweet as non-hateful. The frequency distribution is divided into two sets where one set does not contain hate words, and another set contains hate words. For example, 3.36\% of the time, all three annotators label a tweet as hateful when there are hate words in tweets. In contrast,  only 0.19\% of the time, three annotators label a tweet as hateful when there is no hate word in that tweet (3H vs. 0NH). This observation is also true for the other entries on the x-axis. Annotators agree more given known hate words in tweets.

\begin{figure}[b]
\centerline{\includegraphics[width=0.49\textwidth]{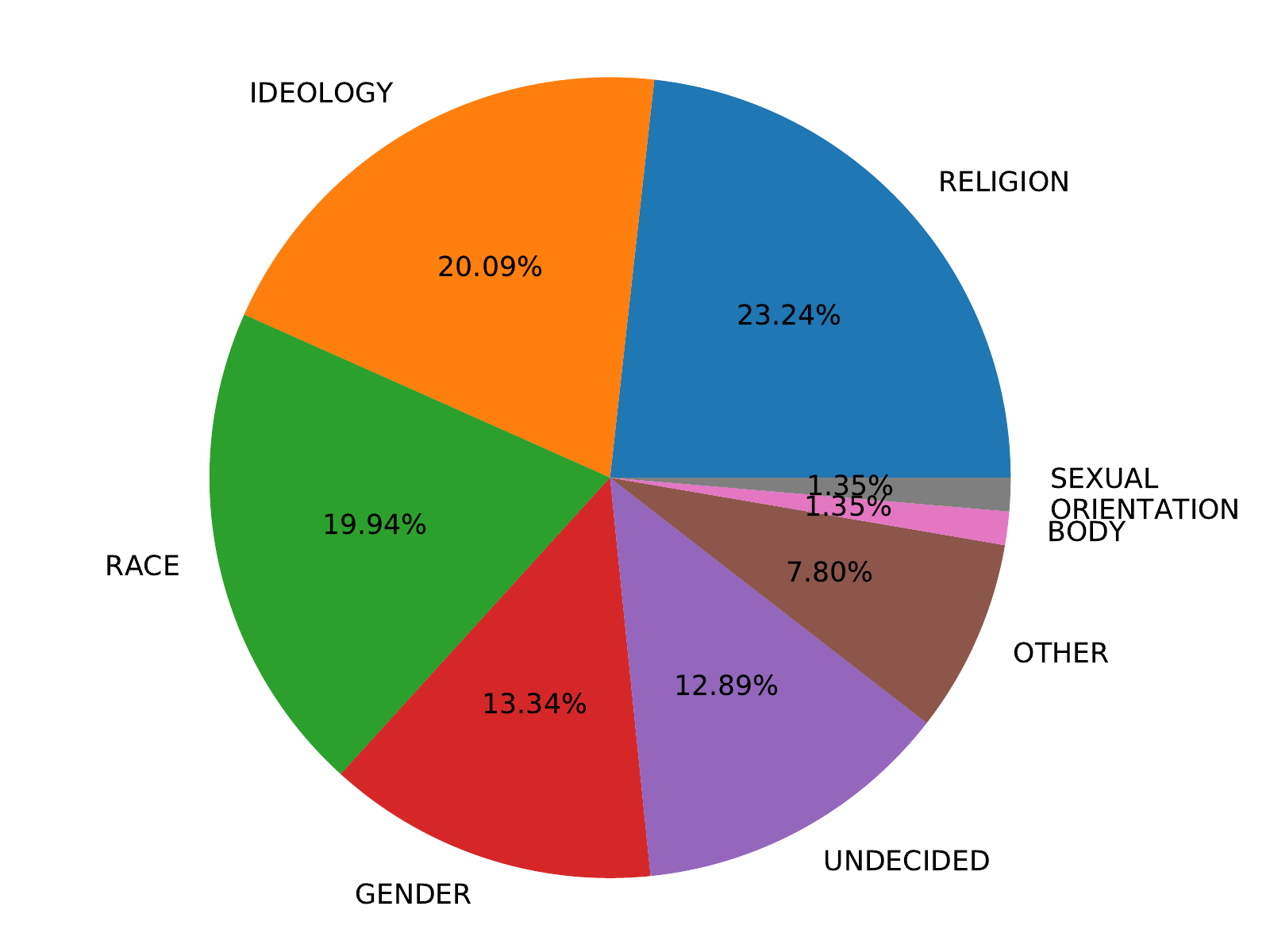}} 
\caption{Hate speech distribution over targeted groups.}
\label{Figure:percentage_hate_group}
\end{figure}

{\bf Targeted group label}. The targeted group label of a hateful tweet is computed using majority voting. However, we find that there are 86 hateful tweets where all three annotators assign different targeted groups. In those cases, the targeted group label is assigned to ``UNDECIDED''. A closer inspection on this ``UNDECIDED'' category reveals that for 83 tweets out of 86 UNDECIDED tweets, one annotator out of three provides ``NONE'' as a targeted group, contrary to our guidelines. Furthermore,  for 35  tweets, one annotator selects ``GENDER'', but another annotator mostly picks ``IDEOLOGY'' or ``RACE''. For example, one annotator selects ``GENDER'' but the other annotator picks ``IDEOLOGY'' for  tweet ``F**k off you old socialist millionaire Clinton b***h''. Here we find ``UNDECIDED'' category because the tweets can be interpreted as hateful to two different target groups.   


{\bf Figure \ref{Figure:percentage_hate_group}} presents the percentage of hate speech under each targeted group. We see that the largest three targeted groups are religion, ideology, and race, accounting for more than 62\% of hate speech in the dataset. Those top three groups have almost equal share ($\approx$ 20\%). To understand further each of these targeted groups, we analyze the top 10 most frequent terms appearing under each of these targeted groups and find that some of these targeted groups cover a broad range of hate speech which is one of our goals. For example, the ``Race'' group consists of hate speech on the basis of ethnicity, race, colour, or descent. Similarly, the ``Religion'' group covers hate towards Muslims and Christians. Furthermore, under the ``Ideology'' group, targeted groups include individual or groups having various ideologies such as liberals, republicans, and feminists. The same observation holds if we consider the ``sexual orientation'' group. However, if we consider the ``Gender'' group, we find that this group mainly covers hate  towards only women.  Finally, 66\% and 20\% of the hateful tweets are explicit and implicit, respectively, whereas the remaining 14\% falls under the ``UNDECIDED'' category of annotator disagreement.

{\bf Inter-annotator agreement for rationales}. Recall that our annotators were asked to distinguish three cases of implicit vs. explicit labeling decisions: types of hate (derogatory language or inciting violence) and demographic group targeted. In explicit cases (only), annotators provided a rationale by highlighting a portion of the tweet supporting their labeling decision. This leads to three complications in how to measure inter-annotator agreement for rationales.  Firstly, there are three different categories of rationales. For simplicity, we ignore this and simply report one statistic over all categories combined.  Secondly, since rationales are only provided in explicit cases, the number of rationales per labeling decision depends on how many annotators identified explicit evidence for their labeling decision. Since inter-annotator agreement measures typically assume the same number of annotations per item, we make another simplifying assumption that all annotators provided rationales, but that for implicit cases, no tokens were part of the rationale.  Thirdly, our interface has annotators highlight rationales at the character level, which we then map to binary token level labels as follows: 1 if all characters in the token are highlighted, and 0 otherwise. After this, we can then calculate annotator agreement as a binary labeling task over all tokens (over all tweets). This yields raw agreement of 95\% (most tokens are not part of rationales), Fleiss $\kappa=0.07$, and Gwet's $AC_1 = 0.95$.

\section{Wellness Risks for Hate Speech Annotators and Moderators}
\label{risk_for_annotators}
To the best of our knowledge, none of the previous work has analyzed the effect of frequent exposure to hate speech on the well-being of human annotators. However, prior evidence suggests that exposure to online abuse has serious consequences on the mental health of the workers \citep{ybarra2006examining}. The studies conducted by 
\citet{boeckmann2002hate} and \citet{ leets2002experiencing} to understand how people experience hate speech have found that low self-esteem, symptoms of trauma exposure, etc., are associated with the constant exposure to hate speech.  

In addition, according to the premier diagnostic manual for
psychological disorders, DSM-5 \citep{american2013diagnostic}, a person can suffer from post-traumatic stress disorder (PTSD) via ``repeated exposure'' to indirect traumatic material, which in our case is online hate speech.  Prior work \citep{ludick2017toward, may2016defining} regarding the psychological effect of indirect trauma also recognizes PTSD as  “secondary traumatic stress”, “compassion fatigue”, and “vicarious traumatization”.  Several studies have been conducted to understand the consequences of constant exposure to trauma, e.g.,   \citet{kleim2011mental} perform research on the first responders, \citet{perez2010secondary}  investigate the police officers, \citet{wagaman2015role} study the social workers, etc. Although there is no prior study regarding the consequences due to the constant exposure to hate speech, recently, some employees have sued Microsoft.  In their lawsuit, they claimed that due to repeated exposure to traumatic contents (e.g., child pornography) as a part of their work, they are being diagnosed with PTSD \citep{microsoft2017}.    

From the above discussion, it is evident that there are some serious consequences for the constant exposure to indirect trauma, and the same is true for annotating hate speech.  As a result, we can see that there are some efforts from organizations to improve the work environment of employees. To be more specific, more than 12 technology companies (e.g., Adobe, Apple, Dropbox, Facebook, GoDaddy, Google, Kik, Microsoft, Oath, PayPal, Snapchat, Twitter) have implemented some guidelines developed by The Technology Coalition  \citep{resilenceGuild2015} to ``support of those employees who have exposure to online child pornography in the course
of their work''. 

Some of the key mitigating steps proposed in the Employee Resilience Guidebook \citep{resilenceGuild2015} are: i) limiting the amount of time an employee can spend on moderating child pornography contents, and ii) acquiring informed consent from employees so that they have a clear understanding of the role as a moderator. The latter strategy has also been emphasized by the University Institutional Review Boards (IRBs).    

Owing to the fact that we work with crowd-workers via a crowd-platform where we do not have any direct control over their work environment, directly implementing the above-mentioned strategies is beyond our control. Consequently, to reduce the risks for the crowd-workers associated with annotating hate speech, we have posted a disclaimer as shown below at the very beginning of the annotation task. 

\begin{quote}
    Our research seeks to reduce the spread of hate speech on social media by training computer programs to automatically detect hate speech. To accomplish this, we ask human annotators to read tweets and label hate speech. We understand that this labeling task requires content that can be disturbing to read. If you prefer to return this task rather than work on it, we understand. In general, if you ever experience mental or emotional distress, please know that help is available online.  Helplines include \url{https://suicidepreventionlifeline.org} in the USA and \url{http://suicide.org/international-suicide-hotlines.html} internationally. For additional reading on this subject, please consult our research article, “The Psychological Well-Being of Content Moderators”( \url{https://www.ischool.utexas.edu/~ml/papers/steiger-chi21.pdf}).

\end{quote}

In the spirit of {\em informed consent}, this disclaimer helps the crowd-workers to make an informed decision about whether or not to accept the task. It also suggests where to seek help regarding any mental or emotional distress.  \\

\section{Discussion and Limitations}
\label{section:limitation}


The primary research goal of this work is to develop a hate speech dataset that covers a broader range of hate speech while maintaining a comparable prevalence of hate. While developing this dataset, we have made various operational decisions regarding various issues discussed in Section  \ref{section:issues_hate_speech}. In this section, we discuss the practical implication of those operational decisions in the constructed dataset. 

I. While defining hate speech, we emphasize that the presence of both targets and actions is necessary to consider a post as hate speech. This definition is consistent with the prior definition of hate speech used in the datasets created by \citet{davidson2017automated, founta2018large, de-gibert-etal-2018-hate} and \citet{nobata2016abusive}. We 
select this hate speech definition because it covers a wide range of hate with a generalized set of targets. However, this also creates room for different interpretations among the annotators, which is reflected in the inter-annotator agreement score of our dataset. On the other hand, \citet{waseem-hovy-2016-hateful} provide an eleven (11) steps approach, including the presence of specific hashtags to consider a post as either sexist or racist. Note that, our annotation interface also has multiple steps to determine whether a post is hateful or not. Another practical limitation of the definition used in this dataset is that it does not cover those hate speech related to praising certain groups (e.g., praising Nazi).

II. Our dataset has been annotated using a binary scheme considering only hate speech and normal speech. However, non-binary schemes are very prevalent in the hate speech domain \citep{davidson2017automated, founta2018large} because it helps us to understand hate and other related concepts (e.g., offensiveness, aggressiveness) using the same annotation effort. One practical limitation of the binary scheme used in this work is that annotators might label an offensive post as hate speech because they have no other categories to specify. For example, pornographic-related posts are typically offensive, and there is a clear instruction regarding this in our guidelines, and yet many annotators label these offensive posts as hate speech in our annotated dataset.  

III. Unlike prior work \citep{waseem-hovy-2016-hateful, davidson2017automated, founta2018large, de-gibert-etal-2018-hate} where a simple annotation interface has been employed, by adapting the suggestion of \citet{sanguinetti2018italian}, we have designed a hierarchical, structured annotation interface to annotate hate speech. The rationale behind this hierarchical interface is to perform the task decomposition, which can help the annotators navigate their decision-making to label a post as hate or normal speech. However, following prior work of \citet{sanguinetti2018italian}, we have also noticed that the use of this structured interface does not necessarily improve the inter-annotator agreement score. Further investigation regarding the annotation interface is needed to understand how different annotation interfaces affect the quality of the annotated data.

IV. Following prior work \citep{founta2018large, davidson2017automated,mathew2020hatexplain}, we have used crowd-workers to annotate hate speech. As mentioned earlier, since training the crowd-workers is practically challenging, we select crowd-workers with specific qualifications (e.g., a minimum HIT approval rate). However, since the crowd-workers are more prone to annotate a post as hate speech based on the keyword-spotting \citep{waseem2016you}, the quality of the annotated data might be affected. To compensate for this issue, we have designed a thorough annotation guideline with various examples considering different boundary cases of hate speech. Furthermore, by noticing the fact that many previous studies do not disclose their guidelines \citep{fortuna2018survey}, we have made our annotation guidelines publicly available with our dataset.

V. We have assumed that annotators can complete the annotation task effectively without any contextual information irrespective of their demographics, expertise, ideologies, etc. Note that this assumption holds for both pooling and active learning methods. However, prior work by \citet{al2020identifying} has shown that if the demographic factors (e.g., first language, gender, etc.,) are not properly handled, potential {\it annotation bias} might arise in the dataset. For example, it has been found that native English speakers are better at detecting toxic comments \citep{al2020identifying} than non-native English speakers when the annotation task is in English.  

VI. Our Twitter-specific dataset is not necessarily representative of how hate is expressed on other social media platforms and forums. For example, tweets have a fixed maximum length (i.e., 280 characters), so our dataset does not cover any hateful expressions longer than this limit.

Other related biases that we should be concerned about regarding the dataset constructed in this work are: i) temporal bias \citep{mehrabi2021survey}, ii) user bias \citep{arango2019hate} and iii) pooling bias \citep{buckley_bias_2006}. Although we have collected tweets from May-2017 to Jun-2017 using a uniform random sample, familiar topics  discussed during that time frame would be over-represented in the constructed corpus and thus introduce the temporal bias in the dataset. Furthermore, \citet{arango2019hate} mention that 65\% of hate speech annotated in the dataset created by \citet{waseem-hovy-2016-hateful} are generated by only two (2) Twitter users. Since we have taken a random sample of tweets from the Twitter API and another random sample from the pooled tweets, user bias should be less prevalent in the constructed hate speech dataset. 

VII. The pooling process introduces two additional biases : i) pool depth bias and ii) system bias and here, we discuss those biases in the context of hate speech. When the pool depth is very shallow, many posts remain unjudged. If several of those unjudged documents are hate speech, that introduces a pool depth bias. Typically, employing a wide pool depth helps reducing this bias. On the other hand, system bias is introduced when the number of machine learning models in pooling is very few and those models are not diverse. This phenomenon reduces the prevalence of hate in the annotated dataset drastically. Generally, system bias can be addressed by increasing the number and diversity of machine learning models \citep{buckley_bias_2006}.

Although handling annotation-related biases (Section \ref{section:issues_hate_speech}) and other biases discussed above is not the key contribution of this  work, readers should be aware of those biases (e.g., racial bias, political bias, gender bias, etc.) while designing an automated hate speech detection system using our hate speech dataset.  Specifically, the presence of these biases might adversely impact the quality of the constructed hate speech dataset. Moreover, when machine learning models are trained on biased datasets, those models typically learn and exasperate those biases. For example, these biased automated hate speech detection systems might flag posts written using American English dialect (AAE) as hate speech or impair political debates on social media platforms because the systems are politically biased. 

Apart from the above-discussed issues of the constructed hate speech dataset, we have made some key assumptions related to pooling and active learning methods that are crucial to achieving our research goal. Here, we discuss those assumptions and their corresponding limitations.   

{\bf Assumption I.} Recall that hate speech is relatively rare in social media (only 3\% of social media posts are hateful \citep{fortuna2018survey}), and annotating everything is not feasible. Thus to maximize the prevalence of hate speech for a given budget, during the pooling, documents (e.g., tweets) that exceed a certain threshold in terms of their likely hatefulness are only considered in the pooled document set. Note that due to this assumption, the process for constructing the pooled document set is a {\it non-random sampling} process, which is prone to {\it sampling bias} because of its nature. In other words, documents having a likely hatefulness score less than the provided threshold are not present in the final dataset, and some of those discarded documents might be hateful. Note that sampling bias is also an issue for active learning \citep{prabhu2019sampling}. In addition, since machine learning models are employed in both pooling and active learning, the selection of posts for annotation is also affected by {\it model bias} \citep{mehrabi2021survey}. This could be most pronounced with our active learning approch because only a single model is used to select tweets, which may reduce the diversity in selection vs.\ the pooling approach across models.

Typically prior work on constructing hate speech datasets is mostly based on searching hate words    \citep{waseem-hovy-2016-hateful, basile-etal-2019-semeval, founta2018large} and/or finding potential hateful social media users \citep{davidson2017automated}. Because of their nature, they are also heavily criticized for having a strong sampling bias. For example, only two users are responsible for generating 70\% of sexist tweets, and only one user generates 99\% of racist tweets \citep{wiegand2019detection} in the dataset constructed by \citet{waseem-hovy-2016-hateful}. Unlike prior work, we do not rely on keyword-based searches or finding hateful users. In addition to that, to mitigate the sampling bias, we have two random sampling steps at two different stages of the pipeline: 1) corpus construction phase and 2) final sampling of documents for annotation. However, potential users of the dataset constructed in this work should be aware of this potential selection bias. They might adopt some de-biasing strategies discussed in prior work \citep{dixon2018measuring, badjatiya2019stereotypical} while designing their automated hate speech detection systems trained on our hate speech dataset.

{\bf Assumption II.} Another key assumption made in the pooling-based approach is that there exist prior hate speech datasets on which prediction models can be trained to kickstart the pooling technique. However, this assumption does not always hold, especially for the less-studied languages (e.g., Amharic, Armenian, etc.). Additionally, since the pooling technique relies on prior hate speech datasets, any known limitations of those datasets will influence the document selection process of the pooling technique. For example, the dataset constructed by \citet{waseem-hovy-2016-hateful} covers sexist posts from the sports domain, and the dataset by \citet{grimminger2021hate} covers political hate speech covering the 2020 US Election topic. This type of {\it topical bias} for the sake of identifying hate speech in the existing hate speech datasets can also be propagated through the pooling technique, and the constructed hate speech dataset can have the same type of topical bias.